\documentclass{article}
\usepackage{jfrStyle}
\usepackage{graphicx}
\usepackage{comment}
\usepackage[natbibapa]{apacite}
\usepackage{setspace}
\usepackage{amsmath}
\usepackage{amssymb}
\usepackage{algorithm} 
\usepackage{algpseudocode}
\usepackage[utf8]{inputenc}
\usepackage{url}
\algrenewcommand\algorithmicrequire{\textbf{Input:}}
\algrenewcommand\algorithmicensure{\textbf{Output:}}

\usepackage{lineno}

\title{A ground mobile robot for autonomous terrestrial laser scanning-based field phenotyping}

\author{
Javier Rodriguez-Sanchez \\
School of Electrical and Computer Engineering\\
The University of Georgia\\
Athens, GA, United States \\
\And
Kyle Johnsen \\
School of Electrical and Computer Engineering\\
The University of Georgia\\
Athens, GA, United States \\
\And
Changying Li\\
Bio-{S}ensing, Automation and Intelligence Laboratory\\
Agricultural \& Biological Engineering\\
The University of Florida\\
Gainesville, FL, United States \\
\texttt{cli2@ufl.edu} \\
}

\begin{document}

\maketitle

\begin{abstract}
Traditional field phenotyping methods are often manual, time-consuming, and destructive, posing a challenge for breeding progress. To address this bottleneck, robotics and automation technologies offer efficient sensing tools to monitor field evolution and crop development throughout the season. This study aimed to develop an autonomous ground robotic system for LiDAR-based field phenotyping in plant breeding trials. A Husky platform was equipped with a high-resolution three-dimensional (3D) laser scanner to collect in-field terrestrial laser scanning (TLS) data without human intervention. To automate the TLS process, a 3D ray casting analysis was implemented for optimal TLS site planning, and a route optimization algorithm was utilized to minimize travel distance during data collection. The platform was deployed in two cotton breeding fields for evaluation, where it autonomously collected TLS data. The system provided accurate pose information through RTK-GNSS positioning and sensor fusion techniques, with average errors of less than 0.6 cm for location and 0.38$^{\circ}$ for heading. The achieved localization accuracy allowed point cloud registration with mean point errors of approximately 2 cm, comparable to traditional TLS methods that rely on artificial targets and manual sensor deployment. This work presents an autonomous phenotyping platform that facilitates the quantitative assessment of plant traits under field conditions of both large agricultural fields and small breeding trials to contribute to the advancement of plant phenomics and breeding programs.
\end{abstract}


\section{INTRODUCTION}
Plant phenotyping has become a pivotal practice in breeding programs, enabling breeders to assess plant traits across diverse environments and over extended periods \citep{Grosskinsky2015}. Field evaluation of these traits plays an important role in developing varieties with enhanced quality, yield, and stress tolerance for targeted locations \citep{Goggin2015}. However, traditional methods for quantifying plant functional traits under field conditions rely on labor-intensive manual measurements and destructive sampling. This not only limits the number of plants that can be assessed but also impedes the scalability of breeding programs \citep{Bowman2004}, presenting a significant bottleneck for breeding progress \citep{Araus2014}. The development of innovative methodologies to automate in-field phenotyping tasks holds the potential to significantly streamline the process and accelerate crop improvement.

Over the past few decades, progress in remote and proximal sensing has significantly contributed to the development of state-of-the-art methods for collecting and processing in-field crop data. These modern techniques, incorporating imaging sensors and three-dimensional (3D) technologies, allow the creation of realistic crop models, enabling the extraction of precise morphological traits to estimate crop growth and productivity \citep{Su2019}. Among the numerous 3D imaging technologies available, light detection and ranging (LiDAR) scanners have emerged as widely utilized tools for field-based phenotyping. LiDAR scanners employ laser beams that can pass through openings in the foliage, surpassing the capabilities of alternative technologies at sensing depth. The measurements obtained from LiDAR scanners exhibit remarkable accuracy and repeatability \citep{Madec2017}. Moreover, LiDARs demonstrate reduced reliance on illumination conditions compared to other sensor types \citep{Hosoi2009}, making them particularly well-suited for field operations.

Terrestrial Laser Scanning (TLS), a technique that utilizes ground-based laser scanners to capture highly accurate and comprehensive 3D data, stands as one of the most prominent LiDAR-based methods for vegetation monitoring \citep{Jin2021Lidar}. To mitigate occlusions and improve laser coverage, LiDAR scans are typically performed from multiple locations across the field, known as multi-scan TLS \citep{Fang2020}. However, the implementation of multi-scan TLS can be both time-consuming and labor-intensive. Traditionally, operators manually carry the LiDAR scanner between different scan locations. Furthermore, to achieve precise co-registration of individual scans, the deployment of artificial targets as common reference points is necessary \citep{Friedli2016}. Depending on the field dimensions and the scanner's specifications (e.g., angular resolution and range), a considerable number of scan locations and artificial targets may be required \citep{Sun2021}. This can significantly hamper the efficiency of the phenotyping process and impose restrictions on the number of plants that can be scanned in each session. Therefore, strategic approaches for survey planning are essential to ensure the execution of efficient field phenotyping tasks, particularly in breeding fields where occlusions can have a substantial impact.

Site planning plays a crucial role in ensuring an effective TLS survey. The planning process for a TLS survey involves identifying the optimal set of scan locations and strategically distributing them throughout the field to acquire comprehensive and accurate crop information. For field phenotyping, recent studies have implemented empirical strategies based on trial and error to determine the most suitable scan locations \citep{Guo2019}. While these methods can be effective in relatively small and straightforward scenarios with minimal anticipated changes in the crop, they prove impractical and inefficient for a typical breeding field with hundreds or thousands of crop plots. Informed decisions based on analytical methods are necessary to optimize the distribution of scan locations, overcoming limitations and ensuring the generalizability of the approach across various field layouts. The integration of robotic technologies with optimized TLS methods offers a promising solution to standardize field phenotyping procedures, reduce dependence on human labor, and improve the throughput of in-field phenotyping tasks, contributing to the advance the field of plant phenomics.

The primary goal of this study was to develop an autonomous robotic system for TLS-based field phenotyping. The study aimed to accomplish the following specific objectives:  (1) integrate a high-resolution 3D LiDAR with a ground robot for autonomous data collection in the field; (2) develop a computer-based approach to optimize the distribution of scan locations for TLS site planning; (3) implement an optimal route planning algorithm for efficient field navigation between scan locations; and (4) conduct field experiments to evaluate the system's performance in multi-plot cotton breeding fields.

The paper is structured as follows: Section 2 provides an extensive review of the various approaches employed in LiDAR-based field phenotyping over the past decade. Section 3 covers the design of the robotic system, addressing both hardware and software aspects. Section 4 delves into the automation of TLS-based field scanning for plant phenotyping. Section 5 presents the experimental plan and details the fields that were utilized for field testing. Section 6 provides a detailed analysis of the results obtained using the developed system. Finally, Section 7 and Section 8 present future research directions and the main conclusions derived from this study.

\section{RELATED WORK}
The emergence of applications based on 3D imaging technologies for plant phenotyping has contributed to increasing the understanding of plant-environment interactions. Canopy architecture is closely related to crop growth rate and yield \citep{Nobel1993} and investigating of this trait is key to identifying those genotypes that ultimately generate a substantial gain in commercially valuable traits \citep{Kumar2015}. However, canopy architecture shows complex spatial and temporal relations that are difficult to estimate \citep{Jin2021Lidar}. 3D technologies allow the creation of realistic crop models that can be used to extract morphological characteristics of plants over the season for the estimation of crop growth and productivity.

\subsection{Use of LiDAR for field phenotyping}
In recent years, there has been a significant increase in the application of 3D imaging approaches to analyze plant architecture. This growth can be attributed to the continuous improvement of optical sensors and advanced data processing pipelines. Techniques like structure from motion and laser scanning, originally used in more mature fields such as architecture and engineering, have now been adapted to agricultural settings, sparking interest in 3D crop analysis within the plant science community \citep{paulus2019}.

LiDAR has become an important technology for plant phenotyping, offering accurate models of crop structures and enabling the extraction of plant traits related to growth and yield. Operating on the principle of lasers emitting pulses, LiDAR scanners acquire spatial information from the environment in the form of point clouds---a discrete set of data points in 3D space \citep{Vosselman2010}). Over time, laser sensors have evolved, becoming more affordable, and the methodologies for processing the point clouds have improved, leading to a progressive increase in the throughput and spatiotemporal resolution of LiDAR-based plant phenotyping systems \citep{Jin2021Lidar}. LiDAR data has been used to estimate canopy-level information for crops like wheat and rice (\citealp{Hosoi2009}; \citealp{Hosoi2012}), determine biomass at the plot level for maize \citep{Jin2020}, and extract plant-level architectural traits under field conditions \citep{Gage2019}. Additionally, high-resolution LiDAR scanners have enabled detailed morphological analysis at the organ level for various plant species, including barley \citep{Paulus2014}, maize \citep{Jin2019Stem}, sorghum \citep{Malambo2019}, and cotton \citep{Sun2021}. However, challenges for LiDAR-based field phenotyping persist, including the need for high-quality data and appropriate data processing methodologies \citep{Jin2021Lidar}.

Different deployment methodologies have been implemented for LiDAR-based field phenotyping. The most widely used approaches include mobile laser scanning (MLS) and terrestrial laser scanning (TLS) techniques. MLS approaches involve carrying the LiDAR sensor through the field to capture laser data. These systems have been implemented using various carrier platforms such as backpacks or wearable devices \citep{Jiang2019,Zhu2021}; gantry systems \citep{Virlet2016,Guo2018,Beauchne2019}, or cable-driven platforms \citep{Bai2019}; as well as mobile vehicles, including self-propelled machines \citep{Rosell2009,Saeys2009,Ehlert2010,Llorens2011,Sun2018,Wang2018}, or custom-made buggies and Phenomobiles \citep{Deery2014,Liu2017,Madec2017,JimenezBerni2018,Walter2019,Yuan2019}. However, MLS faces challenges associated with capturing data while in motion, which can result in limited accuracy and resolution. Robust and precise location information is crucial to ensure proper registration and alignment of the collected point clouds. Additionally, the potential influence of vehicle vibrations can affect the quality of the collected data \citep{Xu2022Areview}, particularly when capturing fine details or objects with small-scale features.

TLS-based approaches, on the other hand, rely on the stationary position of the LiDAR scanner during the scanning process. TLS systems excel in providing detailed and accurate point cloud data, capturing fine-scale details of plant structures with high resolution. In addition, the portable nature of TLS scanners allows for relatively easy movement between scan positions, enabling \textit{in situ} surveys without the need for specialized platforms. Although the throughput of TLS systems can be lower than MLS approaches, their capabilities make TLS a valuable approach for vegetation monitoring \citep{Calders2020}.

\subsection{Current methods for terrestrial laser scanning in field phenotyping} 
While TLS has been widely used in forestry in the last two decades \citep{Calders2020}, its application for crop phenotyping under field conditions is relatively recent. Some studies have successfully applied TLS from only one fixed location (single-scan mode) to estimate height estimation and tracking crop growth \citep{Crommelinck2016,Eitel2016,Malambo2019}. However, the most widely used methodology for TLS-based surveying in agriculture is collecting point clouds from different locations across the field (multi-scan mode) to improve point coverage. Multi-scan mode provides more detailed data and can compensate for occlusion-induced shadowed regions. This mode has been employed to measure growth patterns for various crops \citep{Hoffmeister2013,Eitel2014,Hammerle2014,Koenig2015,Tilly2015,Friedli2016,Kirchgessner2016,Jin2018Deep,Jin2019Stem,Qiu2019,Su2019,Guo2019,Fang2020,Li2020,Jin2020,Elnaggar2021,Sun2021,Lin2022}. However, the lack of efficient methodologies for in-field data collection may be delaying the adoption of these kinds of technologies to its full potential for plant phenotyping \citep{Watt2020}.

The use of robotics for automating TLS data collection in plant phenotyping is an area that has received limited attention in research. Only a couple of studies have investigated this topic. \cite{Qiu2019} conducted a study using a custom-made robotic platform to collect multi-scan TLS data autonomously in a maize field. However, the choice of scan locations was constrained to a paved road running parallel to the field. This limitation led to a restricted field of view for the scanner, resulting in occlusions and subsequent inaccuracies. In one of our previous studies \citep{Rodriguez2022}, we demonstrated an autonomous robot for multi-scan TLS under field conditions. Although we were able to mitigate the occlusion problem, our study highlighted certain limitations on the placement of scan locations to maximize field coverage that can be further improved.

The findings highlight the importance of strategically planning scan point locations in TLS surveys for effective plant phenotyping, particularly in breeding fields characterized by significant occlusion due to adjacent rows of plots. Overcoming these challenges and enhancing survey efficiency require further advancements in optimizing scan locations to maximize field coverage and address occlusion-related issues. The integration of robotic systems for TLS data collection together with strategic site planning algorithms holds promising potential for improving the accuracy and efficiency of plant phenotyping tasks, ultimately contributing to advancements in plant breeding.

\subsection{TLS site planning approaches for in-field phenotyping} \label{TLS_planning}
Before conducting a TLS survey in the field, a series of steps must be undertaken to ensure its success. These steps involve studying the project area needs to identify factors that may cause loss of information, such as obstructions or limited range that can result in shadows or voids in the point clouds. This process can be time-consuming, necessitating prior survey planning to optimize TLS operations. 

The planning of a TLS survey involves identifying the optimal set of scan locations and their distribution within the scene to capture complete information from the target objects. Two methodologies can be found in the literature to determine the best set of scan locations: empirical approaches and computational and data-driven approaches. Empirical approaches involve using real laser scanners to collect TLS data from predefined locations. For example, \cite{OBanion2019} proposed an empirical TLS planning tool for earth surface modeling. The tool provided estimates of the minimum number of scans and optimal resolution needed, based on a database of previously collected point clouds. Similarly, \cite{Guo2019} employed an empirical approach for field phenotyping, comparing different scanning strategies to determine the optimal TLS layout for plant height estimation in a wheat crop. Their results suggested that surveying the field from the four corners was the optimal configuration. However, these findings might not generalize easily to other more complex crops or field layouts.  Empirical methods can be useful in relatively small and simple scenarios with stable target objects, but they become inefficient for large breeding fields with numerous crop plots or when tall and dense crops increase the risk of occlusions.

Computational and data-driven methods utilize digital models of the scene and computational techniques like ray casting \citep{Amanatides1987} to evaluate visibility from every possible scan position.  For instance, \cite{Mozaffar2016} presented an iterative algorithm that employed ray tracing for line-of-sight analysis to minimize the number of TLS survey points needed to cover an outdoor area. \cite{Starek2020} used a similar approach for terrain mapping in an agricultural field, utilizing a simulated annealing algorithm to maximize coverage with the minimum number of scan locations. Computational methods provide efficient computations of visibility but require prior models of the scene, and they may face challenges when significant changes occur during the growing season. 

In recent developments, more advanced techniques based on computer-aided design (CAD) models have emerged, aiming to calculate ray-environment interactions in complex 3D environments. These techniques leverage the concept of 'slabs' to define the bounding volumes of objects within the scene. A 'slab' is a space bounded by two parallel planes \citep{Kay1986}, and the intersection of sets of bounding slabs allows for the efficient definition of any bounding volume. Building upon this idea, \cite{Smits1998} proposed a method to effectively reduce computation resources and improve scene rendering using the slab formulation. Despite the potential advantages, these techniques have seen limited application in TLS layout analysis and, to the best of our knowledge, have not yet been employed in the context of TLS site planning for high-throughput phenotyping.

In this study, we report the first fully autonomous phenotyping platform capable of conducting multi-scan TLS from various locations around and inside the crop for field phenotyping purposes. We employed advanced computational approaches using slab formulation to identify the optimal set of scan locations that ensure an adequate LiDAR coverage of the crop. By integrating this information into the decision‐making pipeline for field data collection, we were able to identify the optimal navigation path for field scanning, thereby increasing the efficiency and autonomy of the TLS data acquisition platform in the field.

\section{ROBOTIC PLATFORM}
\begin{figure} [h]
    \centering
    \includegraphics[width=\textwidth]{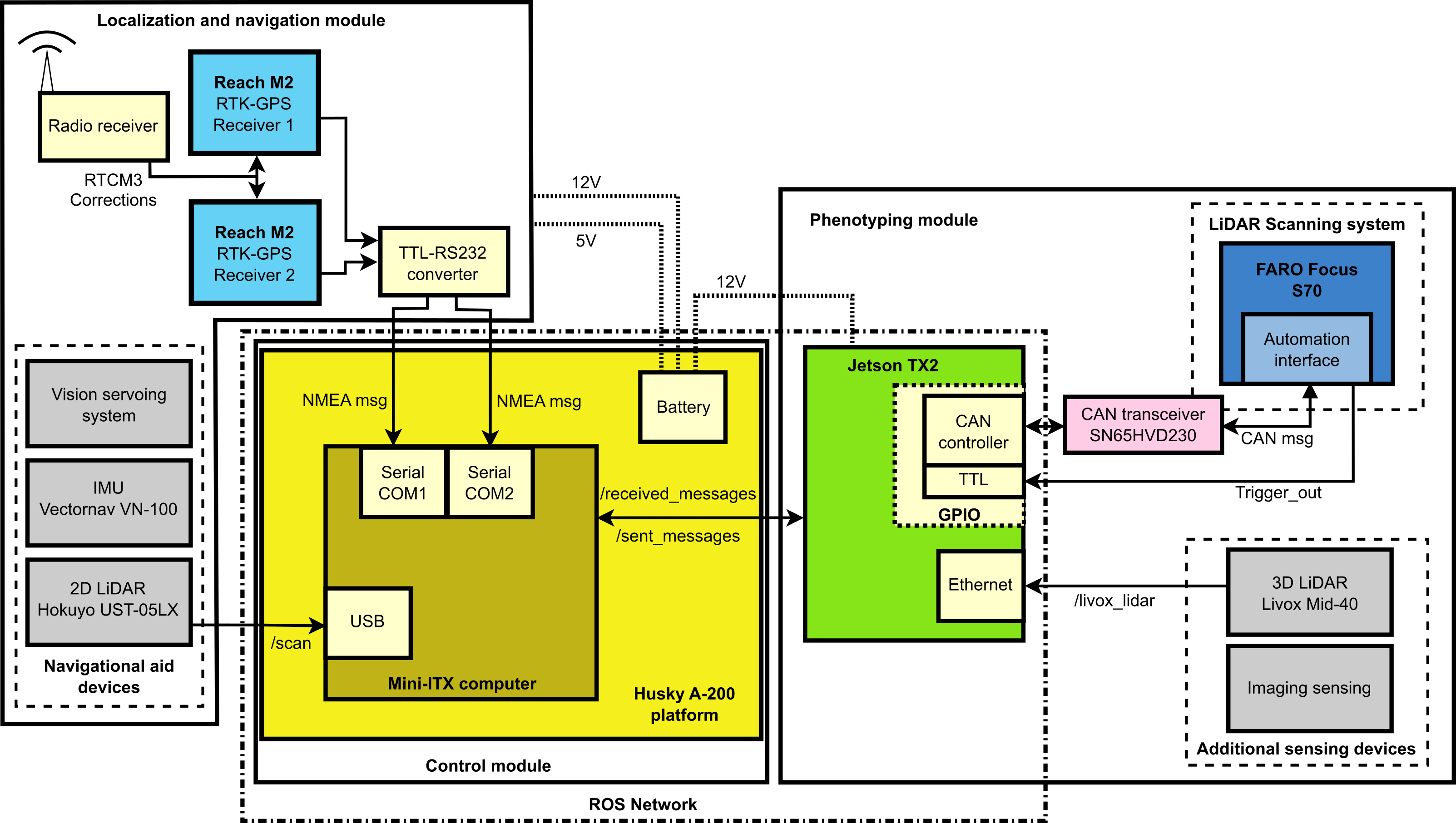}
    \caption{Modular block design for the autonomous phenotyping platform. Colored boxes indicate the main components of each module. Gray boxes indicate additional equipment that could be added to the platform.}
    \label{fig:components}
\end{figure}

In the envisioned autonomous system for LiDAR-based field phenotyping, a robotic platform is in charge of positioning the laser scanner on sufficient scan locations through the field to collect point clouds efficiently. The autonomous phenotyping platform was subdivided into three main constructive blocks: a control module to manage the global operation of the platform, a plant phenotyping module to collect 3D LiDAR data from the crop canopy, and a localization and navigation module to acquire location information and guide the platform to planned scan locations during the mission. Figure \ref{fig:components} shows a block diagram with the main hardware components. The system was integrated under the robotic operating system (ROS) framework to allow for the standardization of methodologies and reduce downtime when different phenotyping tasks are required.

\subsection{Control module}
The control module integrated a series of off-the-shelf hardware devices and software modules to manage the operation of the phenotyping system. This module was composed of a Husky A200 robotic platform (Clearpath Robotics Inc., Ontario, Canada). This robotic platform is a rugged four wheeled skid-steering unmanned ground vehicle (UGV) with dimensions of 990 $\times$ 670 $\times$ 390 mm (L $\times$ W $\times$ H) and a weight of 50 kg. The platform can carry up to 75 kg and has a maximum speed of 1.0 m/s. At the software level, the control module is hosted by a mini-ITX Singleboard Computer System, 2.7 GHz Intel i5-4570TE, with 8 GB RAM and 120 GB SATA2 SSD drive. The control interface, controller manager, and hardware interfaces were integrated under the ROS framework to ensure the interoperability of all its components and the integrity and availability of the data at any given time. The Husky ran ROS Kinetic Kame as the ROS master and included the common ROS packages available at https://github.com/husky/husky.

\subsection{Phenotyping module}
The phenotyping module was composed of a FARO Focus 3D S70 laser scanner (FARO Technologies, Florida, US). This device is a high-resolution 3D LiDAR scanner with a detection range between 0.6 meters and 70 meters and a field of view of 300$^{\circ}$ vertically and 360$^{\circ}$ horizontally. The angular resolution is configurable between step sizes of 0.009$^{\circ}$ and 0.288$^{\circ}$, which is equivalent to a point distance of 1.5 mm and 49.1 mm over 10 m, respectively. This scanner has built-in sensors, such as a dual axis compensator (inclinometer), that can provide additional information for data post-processing. An automation interface adapter (part number: ACCSS8004) was attached to the laser scanner and provided a Controller Area Network (CAN) interface for remote control.

To control the scanner under the CAN protocol, predefined messages were used. The CAN message format followed the CAN 2.0A standard with an 11 bit identifier (CAN ID) and up to 8 bytes of data. The CAN ID field is used as a command key and the data field is used to set the associated parameters for each command key. By complying with the CAN bus specification, the scan data recording can be enabled by the predefined CAN message “Initiate Scan Operation.” The scan data recording can be paused and stopped by the messages “Record/Pause” and “End Scan Operation,” respectively. Under CAN control mode, the scanner returns general acknowledgment messages (ACK) after the reception of each command \citep{Faro2020}. In addition, the automation interface provides two real-time capable transistor-transistor logic (TTL) signals for synchronization, identified as \textit{trigger$_{in}$} and \textit{trigger$_{out}$}. The \textit{trigger$_{in}$} signal supports the integration of time stamps into the CAN messages and the enabling or pausing of the scan data stream recording. The \textit{trigger$_{out}$} signal triggers outgoing CAN messages, emits synchronization pulses during mirror rotation, and can be used to check the current automation status by the signal level.

The FARO LiDAR was interfaced with the ROS network using CAN bus through a Jetson TX2 Developer Kit module (Nvidia Corporation, California, US) that was integrated under the ROS network and was used as the CAN interface controller. The Jetson TX2 module had Ubuntu Bionic Beaver (18.04) installed and ran ROS Melodic Morenia as a slave machine. A CAN board (Waveshare Electronics, Shenzhen, China) based on the SN65HVD230 CAN transceiver was used to interface the single-ended logic used by the CAN controllers at both ends of the bus with the differential signal transmitted over the CAN bus. The components were integrated into ROS using the \textit{socketcan\_brige} ROS package and \textit{CANopen} communication protocol. The Jetson TX2 received actuation orders from the control module and published the CAN messages to actuate the laser scanner. The robotic platform was able to autonomously start the scanning operation of the LiDAR, stop the current scanning, and monitor the status of the scanning process during data collection. This allowed the robot to fully automate the LiDAR data collection process and control the timings for navigating between scan locations. 

\subsection{Localization and navigation module}
For field navigation, the robotic system used real-time kinematic (RTK) positioning based on the global navigation satellite system (GNSS) to obtain robot’s location information. RTK-GNSS provides readings with centimeter-level accuracy in both horizontal and vertical dimensions, with the vertical accuracy typically less precise than the horizontal accuracy. This high precision is achieved through the transmission of positioning corrections from a fixed base station to a mobile rover. To estimate the orientation of the platform during navigation, a GNSS compass technique was implemented using two GNSS receivers (i.e., rovers). This technique uses information from two receivers rigidly mounted at a fixed distance with respect to each other (i.e., compass baseline) to determine the platform heading. The use of two GNSS antennas providing absolute positioning information allows the estimation of the heading angle ($\psi$) that defines the direction of movement of the platform at any given time.

The GNSS receivers provided absolute location information in the form of geographic coordinates (longitude, latitude, and altitude). These global coordinates can be transformed to local coordinates such as the LiDAR scanner or the rovers, by applying the adequate homogeneous transformation matrix (i.e., rotation and translation). In a general form, on a 3D Euclidean space, transformations between a global coordinate system (G) and any arbitrary local coordinate system (L) can be computed using homogeneous transformations to account for the translation and rotation of the coordinate systems as follows:
\begin{gather} \label{eq:fullTransform}
 \begin{bmatrix} X_{G}\\ Y_{G} \\ Z_{G} \end{bmatrix}
 = \ \begin{bmatrix} \Delta_{x}\\ \Delta_{y} \\ \Delta_{z} \end{bmatrix}_{G\rightarrow L}
    +
    \ R_{L}^G \left(\phi,\theta,\psi\right)
   \times
   \begin{bmatrix} X_{L}\\ Y_{L} \\ Z_{L} \\ 1 \end{bmatrix}
\end{gather}

where X$_{G}$, Y$_{G}$, and Z$_{G}$ are the coordinates of the point in the global coordinate system; X$_{L}$, Y$_{L}$, and Z$_{L}$ are the coordinates of the origin of the local coordinate system in the global frame; R represents a 3D rotation matrix expressed by the basic rotations about each one of the axes (x-, y-, z-axis) of the coordinate system; and $\left[ \Delta_{x}, \Delta_{y}, \Delta_{z} \right]$ is a vector that represents the displacement between the origins of both coordinate systems. 

\begin{figure} [H]
\centering
\includegraphics[height=3.0in]{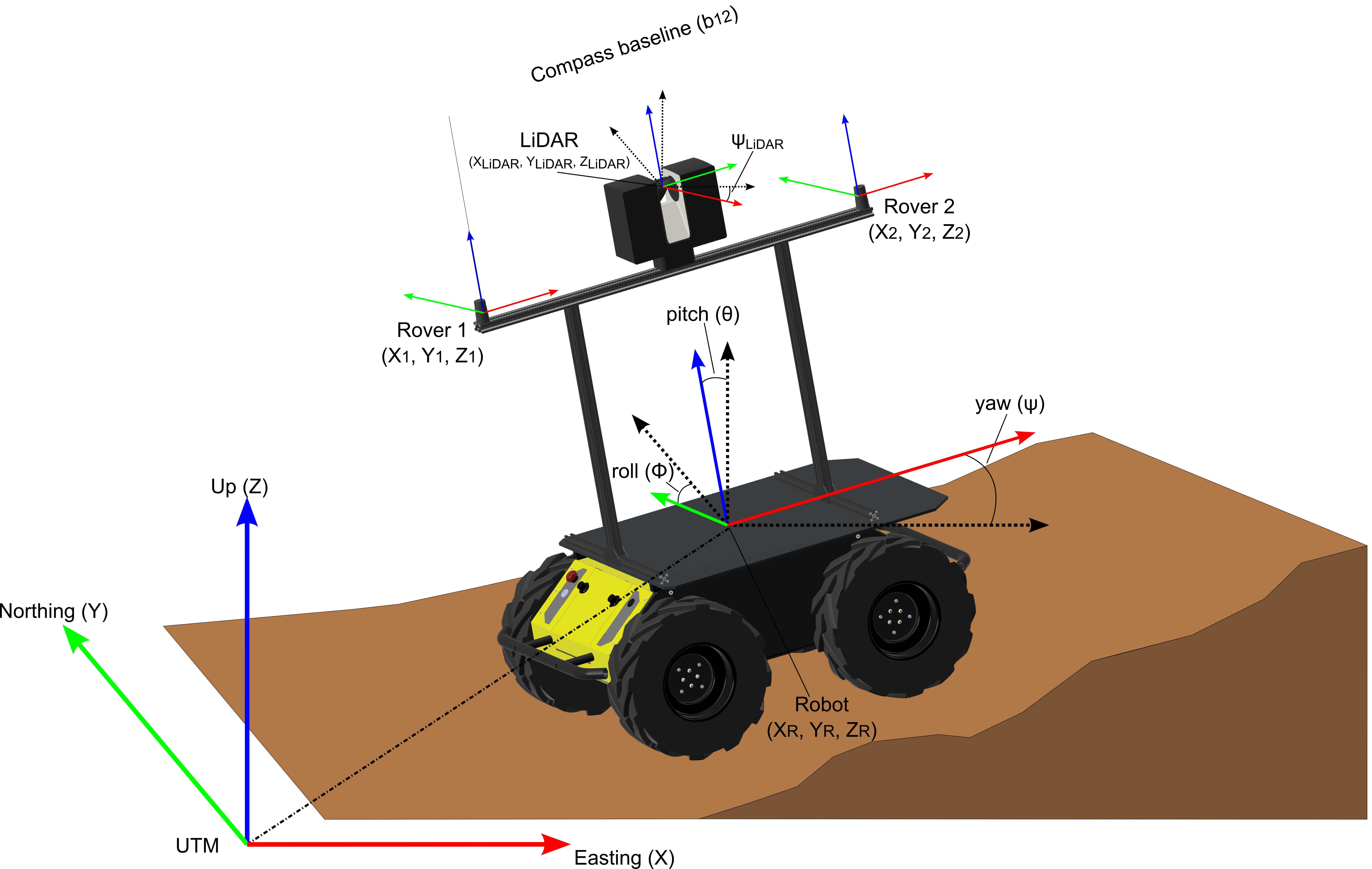}
\caption{Coordinate systems for the phenotyping platform. Global (UTM) and local coordinate systems representation for major components of the robotic platform under an ENU (East-North-Up) coordinate system. Coordinate systems orientations on x-, y-,and z-axis are identified by red, green, and blue arrows, respectively. Dotted vectors indicate orientation of the global coordinate system at robot's origin.}
\label{fig:gnss_compass}
\end{figure}

Figure \ref{fig:gnss_compass} shows the different coordinate systems involved during robot navigation and the relationship of the angles that define the platform's 3D orientation with their respective axes under an ENU (East-North-Up) coordinate system. X$_{R}$, Y$_{R}$, and $Z_{R}$ are the coordinates of the platform in the global coordinate system, which are coincident with the center of rotation of the robot. X$_{1}$, Y$_{1}$, and Z$_{1}$, and X$_{2}$, Y$_{2}$, Z$_{2}$ are the coordinates of the GNSS antennas ($Rover_{1}$ and $Rover_{2}$) in the global frame, respectively. 

Under this coordinate system, the heading angle of the platform will be given by the relationship between rover's X and Y coordinates as follows:

\begin{equation} \label{eq:yaw}
    \psi = \arctan \left(\frac{Y_{2}-Y_{1}}{X_{2}-X_{1}} \right),
\end{equation}
where $\psi\in[-\pi,\pi]$. It is worth noting that the signs of the numerator and the denominator in Equation (\ref{eq:yaw}) must be taken into account to place the heading angle in the right quadrant.

\subsection{Phenotyping robot implementation}
\subsubsection{Hardware components integration}
For our experiments, a GNSS antenna Smart6-L (Novatel Inc., Calgary, AB, Canada) was used as the base station (i.e., base). A FGR2-C radio receiver (FreeWave Technologies, Boulder, CO, USA) was in charge of broadcasting the RTK corrections from the base station. Both the GNSS and the radio antennas for the base station were placed on top of a 6 m post in an open space at approximately 400 meters from the fields under study. The rovers were configured to receive positioning corrections from the base using a shared FGR2-C radio receiver that was connected to a dual-channel MAX3232 RS232 to TTL transceiver breakout board (SparkFun Electronics, Boulder, Colorado) installed on the platform. The positioning system was configured to provide GPS-based pose readings at a rate of 10 Hz.

\begin{figure} [H]
    \centering
    \includegraphics[height=3.0in]{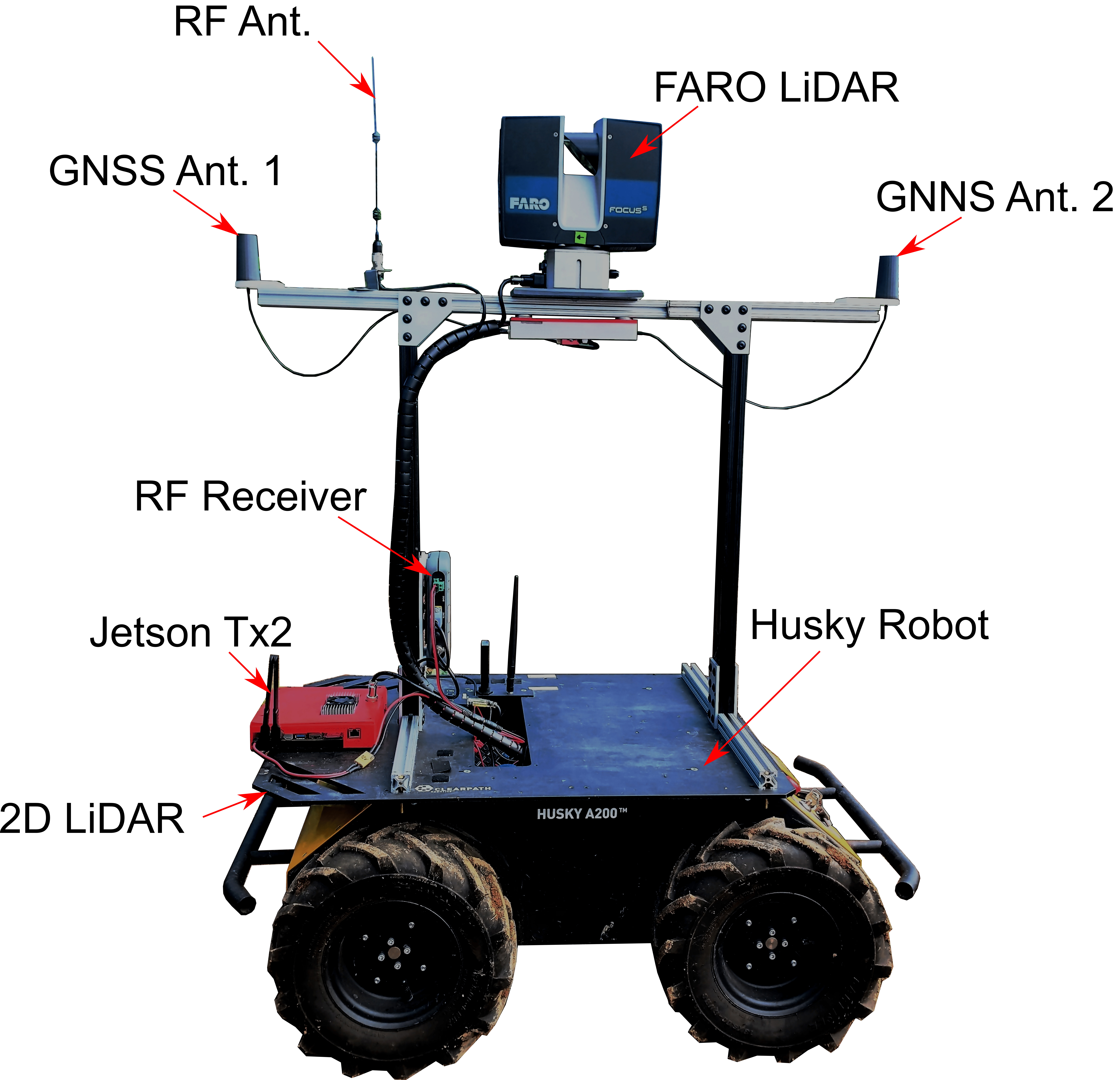}
    \caption{Autonomous phenotyping platform integration.}
    \label{fig:platform integration}
\end{figure}

The robotic system mounted two RTK-GNSS antennas (\textit{GNSS Ant.$_1$} and \textit{GNSS Ant. $_2$}) that were placed longitudinally on the platform in such a way that the baseline vector (i.e., the vector between the two antennas) was oriented towards the moving direction of the robot (Figure \ref{fig:platform integration}). Each GNSS antenna receiver was composed of a Reach M2 multi-band RTK GNSS module (Emlid, Budapest, Hungary) that were attached rigidly to the robotic platform frame on the longitudinal axis. The GNSS receivers were installed at a height of 110 cm above ground, with the same physical orientation relative to each other. The baseline between antennas was fixed to 95 cm. Figure \ref{fig:platform integration} shows the physical integration of all components of the phenotyping platform. 

\subsubsection{Navigation stack}
\begin{figure} [H]
    \centering
    \includegraphics[width=\textwidth]{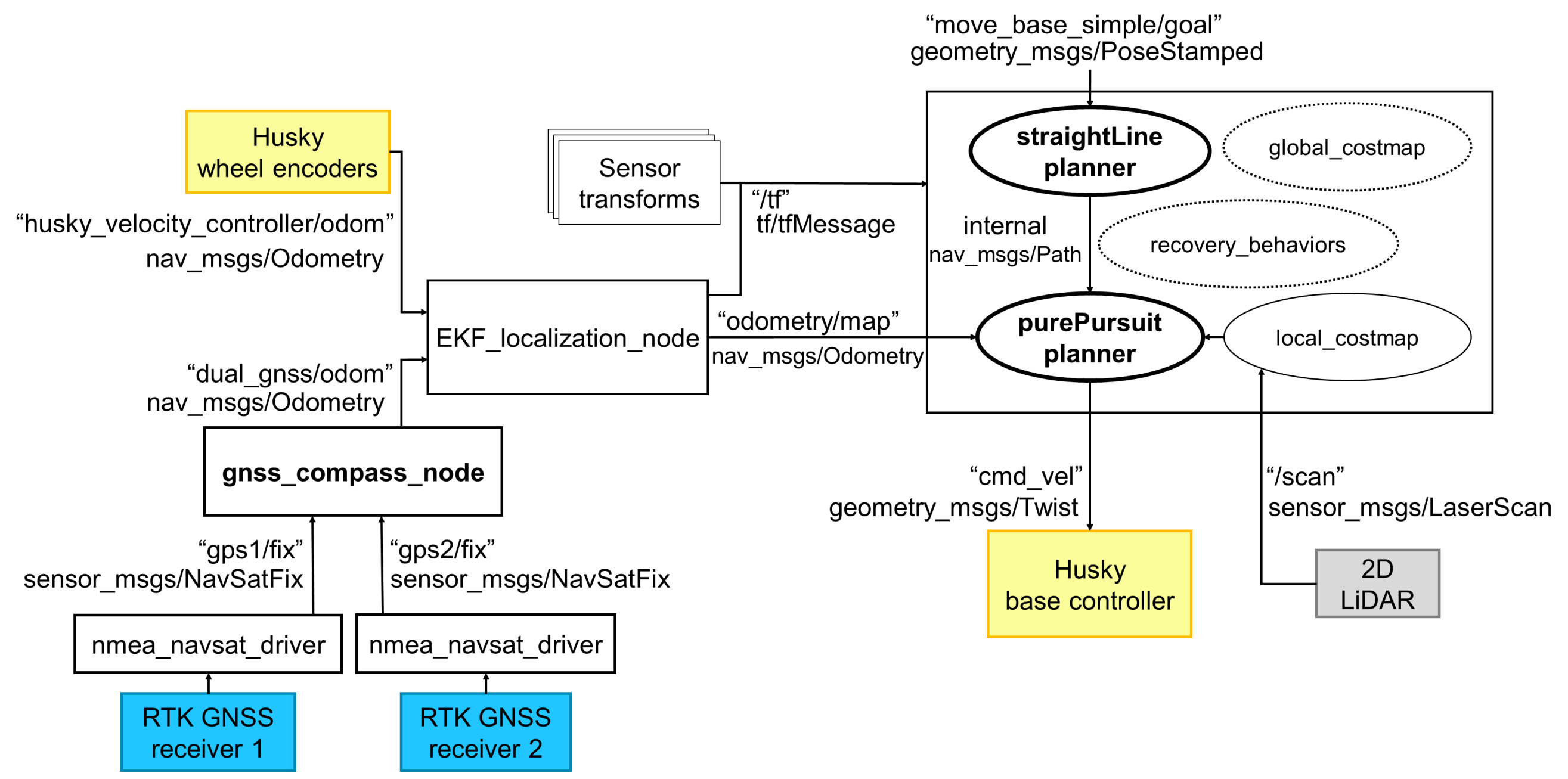}
    \caption{Navigation stack for the autonomous phenotyping platform. }
    \label{fig:nav_stack}
\end{figure}

Path tracking involves the control problem that deals with the constraints of the robot movements and the timing during field navigation. Its solution provides actuation commands to follow the adequate trajectory to reach each waypoint satisfactorily. The control module was in charge of reading the list of waypoints, computing the navigation path between waypoint pairs, and sending the adequate velocity commands to the actuators to follow the path. Our implementation of the ROS \textit{navstack} (Figure \ref{fig:nav_stack}) was based on the \textit{robot\_localization} package \citep{Moore2014}. A single extended Kalman filter (EKF) instance was used to compute state estimates by fusing the odometry information provided by the wheel encoders together with the GNSS compass information provided by the dual RTK-GPS antenna system. In order to provide additional information to the navigation stack and to improve the in-field autonomous navigation, a Hokuyo UST-05LX 2D laser scanner (Hokuyo Automatic Co. Ltd., Osaka, Japan) was installed on the platform for obstacle avoidance. This sensor was integrated into ROS using the \textit{urg\_node} ROS wrapper for the Hokuyo \textit{urg\_c} library. This node supplied an obstacle layer to the navigation stack that was used to avoid colliding with elements placed directly in the robot's trajectory.

\section{AUTOMATION WORKFLOW}
\begin{figure} [H]
    \centering
    \includegraphics[width=\textwidth]{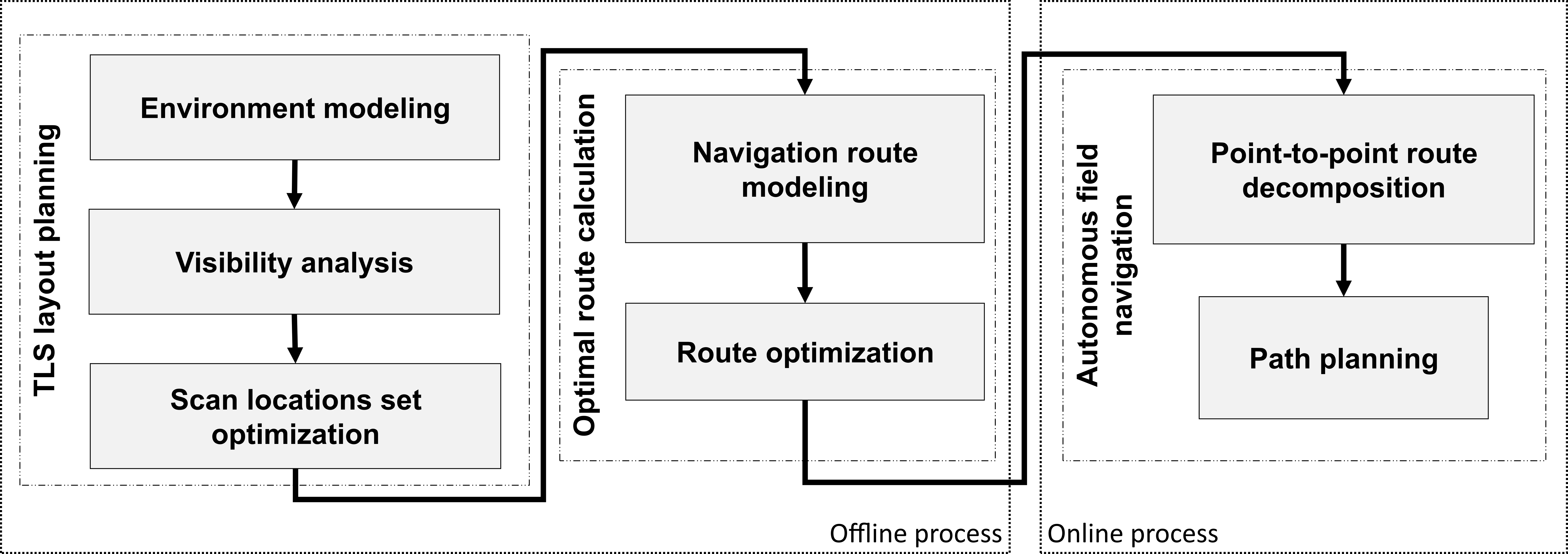}
    \caption{Terrestrial laser scanning automation pipeline. Overview of the proposed methodology for automating field-based plant phenotyping using TLS and robotics}
    \label{fig:approach overview}
\end{figure}

To ensure that the data collected by the LiDAR scanner are suitable for plant breeding purposes, the system must guarantee that the point clouds of the crop are acquired with the adequate data quality, and within time limits. Unfortunately, there are still no standards that help define the minimum quality required for specific phenotyping tasks, and current TLS-based phenotyping studies base their survey planning on personal experience. Robotics and automation technologies can help standardize LiDAR-based field phenotyping tasks by performing efficient TLS surveys without human intervention. Our methodology to automate the TLS process using ground robots is composed of three main steps (Figure \ref{fig:approach overview}). First, the set of scan locations must be optimized to reduce the time needed to perform the TLS survey without losing important information from the crop. Second, the optimized set of scan locations needs to be converted to waypoints for the robot to be able to follow them during field navigation. Finally, the robotic platform needs to navigate the field in a reliable and efficient manner. 

\subsection{TLS layout planning}
Planning a TLS survey can be both challenging and time-consuming, especially when applied to plant phenotyping tasks, which introduces additional limitations. One important aspect to consider is that the accuracy of laser sensors tends to decrease as the distance to the target objects increases, potentially affecting the quality of the point clouds used for plant phenotyping. Furthermore, not all field locations where the scanner can be placed will provide the same level of detailed information about the crop due to occlusions. As a result, site planning becomes a key step for conducting an efficient TLS field survey. Identifying the most suitable set of locations for field scanning will significantly enhance the overall efficiency of the TLS survey, leading to more effective data collection and analysis.

\subsubsection{Environment modeling} \label{fieldDescription}
\begin{figure} [h]
    \centering
    \includegraphics[width=\textwidth]{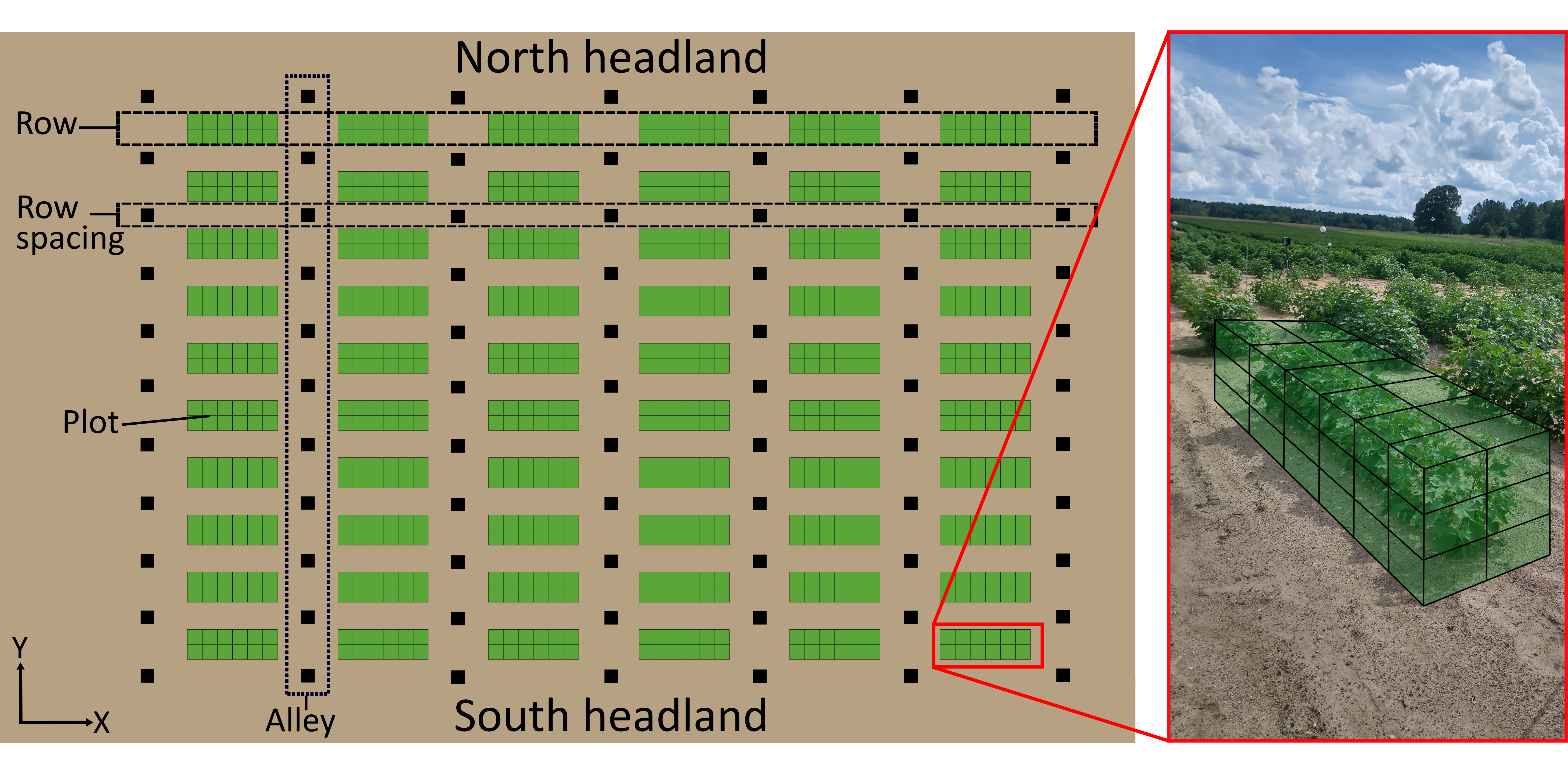}
    \caption{Layout of a typical plot-based breeding field. Green rectangles represent each plot in the field. Black squares represent potential scan locations. The zoomed-in inset on the right shows an example of discretization of one of the plots in the field.}
    \label{fig:field_digitization}
\end{figure}

The first step in identifying the optimal set of scan locations is to determine which positions in the field are suitable for placing the LiDAR scanner. Achieving this requires creating a model of the scene based on basic information about the field layout. Typically, in a breeding field, plots are planted in straight lines using agricultural machinery, forming long rows separated by alleys to divide blocks of different treatments (Figure \ref{fig:field_digitization}). Additionally, the outer edges of the field (headlands) are left unplanted to ensure safe turns for farm equipment during field operations. Both the distance between rows (row spacing) and the alley width are configured according to breeders’ specific needs, ultimately determining the arrangement of the plots in the field.

To model the crop, each plot in the field can be replaced by the imaginary rectangular parallelepiped that encloses its volume (i.e., bounding volume or bounding box) in such a way that they are oriented towards the row direction. This plot representation resembles the oriented bounding boxes (OBB) representation \citep{Gottschalk1996}. An OBB is a rectangular bounding box oriented arbitrarily in space. By choosing the coordinate system for the field in such a way that its axes are parallel to the faces of the bounding boxes, the field can be then modeled using the simplified axis-aligned bounding box (AABB) representation \citep{Schneider2002}. An AABB is a subclass of OBB whose main axes are parallel to the basis vectors of the frame in which they're defined. Using this representation, each plot in the field will be represented by an AABB that can potentially intercept the LiDAR's laser beams (green rectangles in Figure \ref{fig:field_digitization}).

The simple substitution of each plot on the field for its AABB will not allow us to accurately model the interaction of the laser beams with the crop. The footprint of these laser beams is usually small (in the order of mm), so the same plot will be potentially hit multiple times by a great number of different rays. Thus, a finer representation of the plots would be beneficial for modeling ray-crop interactions. Each bounding box can be further discretized using smaller cells or voxels. After some tests, we found that voxels of 0.5 $\times$ 0.5 $\times$ 0.33 meters (length $\times$ width $\times$ height) for discretizing each AABB provided the best tradeoff between resolution and modeling time. For example, for a 3 $\times$ 1 $\times$ 1 plot the mentioned cell division will translate into six cells on the longitudinal axis, two cells on the lateral axis, and three on the vertical axis. In total, each plot will be composed of 36 cells that can intercept rays (Figure \ref{fig:field_digitization} (inset)).

For TLS surveying for field phenotyping, any space not occupied by plants is deemed obstacle-free. While any unoccupied position in the field may appear suitable for scanner placement, certain considerations must be taken into account. Laser scanners have a minimum scan distance (typically ranging from 0.5 to 1 meter) below which accurate distance computation becomes challenging. Thus, the scanner must be positioned at a distance greater than this minimum from any adjacent plots to avoid measurement errors, especially beyond the seedling or first leaf growth stages. Due to this constraint, the space between neighboring rows cannot be considered as potential scan locations. Only alleys and headlands will meet the necessary distance requirement to enable precise LiDAR data collection. Therefore, only these areas should be regarded as obstacle-free spaces for scanner placement to ensure accurate TLS surveys in a breeding field.

Moreover, the use of a ground robot as the carrier for the LiDAR scanner imposes further limitations on the space that can be effectively used for in-field navigation. For a ground robot to navigate the field safely, a minimum clearance with obstacles equivalent to its footprint (the effective area occupied by the robot platform while stationary) is needed. In breeding fields with a grid-like distribution of rows, alleys, and headlands, intersections between row spacing and alleys offer ample space for the robot to maneuver, allowing for smooth directional changes without causing damage to the crop. As a result, only these locations (black squares in Figure \ref{fig:field_digitization}) will be used as potential positions for TLS placement without affecting the crops.

\subsubsection{Visibility analysis}


The second step to optimize the TLS layout is to identify which parts of the crop are visible from each potential scan position. Following a line-of-sight (LOS) problem formulation \citep{Koranne2009} with the LiDAR scanner being considered as the observer, we can identify which parts of the digitized crop are visible from each specific location using ray casting analysis \citep{Appel1968}.

\begin{figure} [h]
    \centering
    \includegraphics[width=\textwidth]{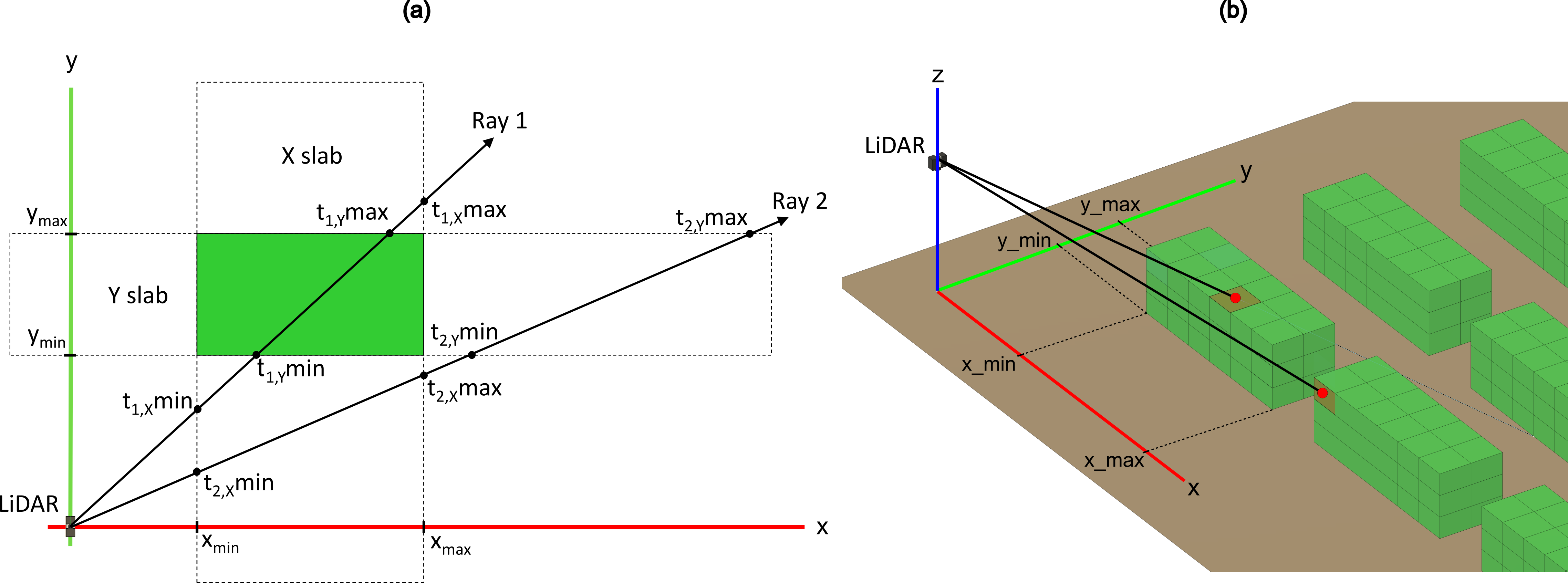}
    \caption{Computation of ray-AABB intersection using the slab method for two rays. (a) Computation example on the XY plane. (b) Cell visibility in the 3D space. Red dots show the impact point of each ray. Darker cells indicate cells visible from the scan location (visibility$=$1) for the specific ray.}
    \label{fig:slab_method}
\end{figure}

The utilization of the slab method, as proposed by \cite{Smits1998}, presents an efficient approach for calculating ray-box intersections within complex 3D environments. Unlike the general case where a ray might intersect all six planes defining the box, the slab method streamlines the computations to just three intersections: X, Y, and Z slabs. By taking advantage of the specific characteristics of the AABB representation, this reduction significantly speeds up the intersection tests. Figure \ref{fig:slab_method} (a) shows the application of the slab method for two different rays on the XY plane.

Following the notation in \cite{Williams2005}, the ray will intersect the bounding volume if both of the following conditions are met: t$_X{min}$ $<$ t$_Y{max}$ and t$_Y{min}$ $<$ t$_X{max}$. Here, t$_{i_{min}}$ and t$_{i_{max}}$ represent the closest and farthest points of intersection with the \textit{i}-slab from the origin of the ray, respectively. Failure to satisfy any of these conditions will result in the ray not intersecting the slab and consequently missing the bounding volume.

If the ray intersects all three slabs, it also successfully intersects the box, leading to two intersection points—--the entry and exit points of the ray. However, for visibility analysis, only the entry point is essential, as it determines which cell of the plot is visible from the current scan location (see Figure \ref{fig:slab_method} (b)). To further optimize computation time, we also considered the hierarchical distribution of bounding boxes and cells. First, we evaluate whether the ray hits the plot level, and only if it does, we proceed to calculate the specific cell intercepted by the ray within the plot.

To conduct the visibility analysis for the entire field, from each scan location a set of laser beams is created according to the parameters of the scanner such as start and end angles, and vertical and horizontal angle steps (i.e., angular resolution). The start angle will determine the first ray direction. At every angle step, the ray casting algorithm will cast new rays and compute the intersections with the bounding boxes in the digitized field. The ray casting algorithm will stop when the end angle is reached. This process is executed in an iterative fashion until all the scan locations in the field are visited.

By iterating over all the rays emitted from a particular scan location, we will obtain the entire field of view for said scan location. This field of view determines the cells' visibility from each specific scan location in the field. The visibility of a cell can be defined as the direct line of sight from a scan location, in such a way that a cell will be considered visible (visibility = 1) if it can be reached by any rays emitted from the scan location, or invisible (visibility = 0) if it is occluded and cannot be reached by any rays 
Repeating this step for all scan locations, a visibility score table for the entire field can be created. This visibility score table will include a row for each scan location and a column for each plot cell on the field. For each iteration of the ray-box intersection algorithm, the table will be updated with the number of cells hit by the current laser beam.

\subsubsection{Scan locations set optimization}
After computing the visibility score table, a greedy algorithm \citep{Chvatal1979} was implemented to iteratively select the optimal or near optimal set of scan locations. The greedy algorithm is a polynomial time algorithm with several variants commonly used to approximate set-covering problems \citep{Slavik1997}. Algorithm \ref{alg:greedy} shows a basic pseudocode to implement the original variant.

Given a set \textit{U} of \textit{m} cells to be covered, a set \textit{V} of \textit{n} candidate scan locations or viewpoints, and a set \textit{C} of \textit{p} cells covered per each scan location, a visibility score list is created by aggregating the number of visible cells per scan location (Algorithm \ref{alg:greedy}, lines 13 to 16). Then, the list is sorted in descending order according to the total number of visible cells per scan location to find the viewpoint with maximum visibility (Algorithm \ref{alg:greedy}, line 17). The first scan location of the list covers the largest number of elements in the scene, and the greedy algorithm chooses this viewpoint as one of the scan locations of the optimal set (Algorithm \ref{alg:greedy}, line 18). Then, this viewpoint is removed from the table together with all the associated cells that are visible from it (Algorithm \ref{alg:greedy}, lines 19 and 20). This process is iterated until all cells are covered.

\begin{algorithm} 
	\caption{Greedy selection of TLS scan locations} \label{alg:greedy}
	\begin{algorithmic}[1]
	\State \textbf{Input:} 
	    \State Cells to be covered $~U=\{ U_{k};~k=1,2,...,m \}$
	    \State Candidate scan locations $~V=\{ V_{j};~j=1,2,...,n \}$
	    \State Cells covered per scan locations $~C=\{ U_{j,l};~l=1,2,...,p; ~p\leq m \}$
	\State \textbf{Output:} 
	    \State Optimal set of scan locations $S=\{ S_{i},~i=1,2,...,q;~ q\leq n \} $
	\State Initialization:
	    \State $S=[~] $
	    \State $V_{temp}=[~] $
        \State $U_{temp}=U$
        \State $i=1$
    \While {$U_{temp} \neq [~]$} 
      \For {$j=1,2,\ldots,n$}
          \State $vis(V(j))=\sum_{l=1}^{m} C(j,l)$\Comment {calculate visibility score per scan location}
          \State $V_{temp}(j) \gets [V(j), vis(V(j))]$ \Comment {populate visibility list}
	    \EndFor
	  \State $V_{temp} \gets sort(V_{temp})|_{Vis(V_{temp}))}$ \Comment{sort list in descending order by visibility score}
	  \State $S(i) \gets V_{temp}(1)$ \Comment{store first scan location}
	  \State $ U_{temp}(C(V_{temp}(1))) \gets [~] $ \Comment{remove cells covered by the scan location}
	  \State $V_{temp}(1) \gets [~]$ \Comment{remove scan location from list}
	  \State $i \gets i+1$
	\EndWhile
	\State return S
	\end{algorithmic} 
\end{algorithm}

\subsection{Optimal route calculation}

\subsubsection{Navigation route modeling} \label{navigationRoute}
\begin{figure} [H]
    \centering
    \includegraphics[height=2in]{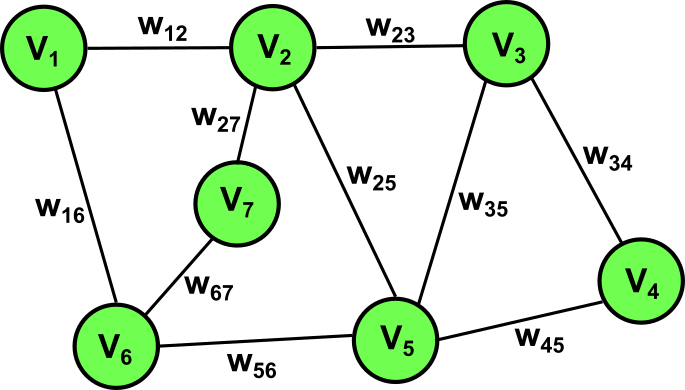}
    \caption{Weighted undirected graph G(V,E). Vertices ($V_{1},...,V_{7}$) represent the set of scan locations to visit. Edges weight ($w_{12},...,w_{67}$) represent the distance between scan location pairs.}
    \label{fig:graphGeneric}
\end{figure}

Once the optimal set of scan locations is determined, the robotic platform must visit each scan location to perform the TLS survey. In order to achieve that, the autonomous system needs to find a valid route to join scan location pairs by using the free space in the field. Using graph theory \citep{Thulasiraman2010}, the navigation problem can be modeled as a weighted connected undirected graph $G=(V,E)$ (Figure \ref{fig:graphGeneric}), where $V$ is the set of vertices or nodes representing the scan locations, and $E$ is the set of edges that connect the nodes, whose weight $w(e)$ is given by the distance between scan locations.

The parameters of the crop such as plot length and width, row spacing, and alley width were used for field digitization. This information can also be used to calculate the weight of the edges of the graph for the routing problem. Assuming a perfectly flat field where the robot platform moves only on a two-dimensional plane, the distance \textit{d(a,b)} between any two points $a$ and $b$ on this plane located at coordinates $a = (a_{x}, a_{y})$ and $b = (b_{x}, b_{y})$ respectively, can be computed in a general form using the Minkowski distance:
\begin{equation} \label{eq:distance}
    d(a,b) = \left( |a_{x}-b_{x}|^p + |a_{y}-b_{y}|^p  \right)^{1/p}
\end{equation}
where \textit{p} represents the order of the norm and determines the type of distance to be calculated.

\begin{figure} [H]
    \centering
    \includegraphics[height=3.0in]{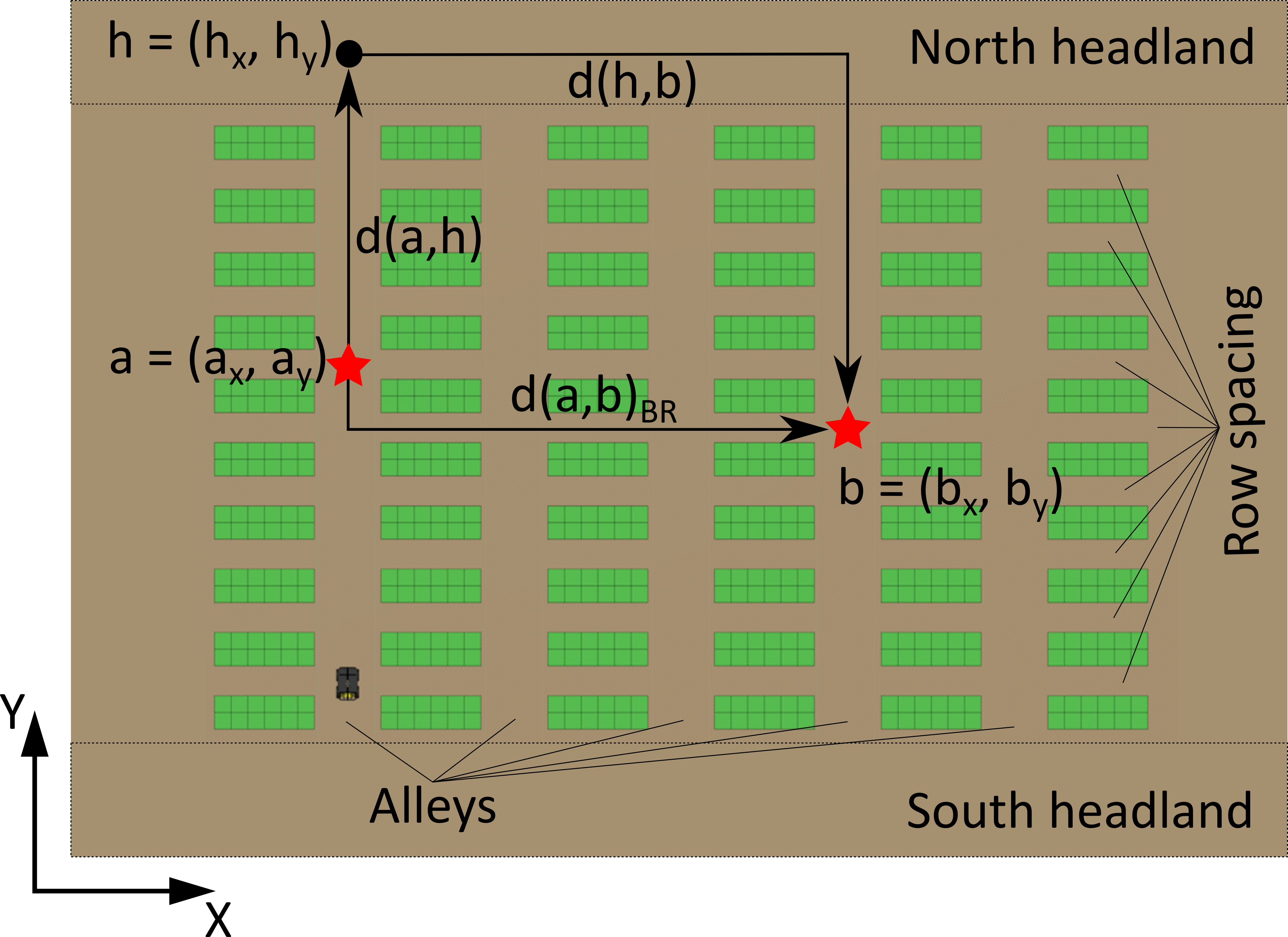}
    \caption{Different approaches for computing distances between scan location pairs. Red stars indicate consecutive scan locations. Black arrows show the direction of the route for each approach.}
    \label{fig:navDistancesExplain}
\end{figure}

Navigation in a crop field will be restricted to the free space (i.e., row spacing, alleys, and headlands). This free space will change through the season as plants grow and the crop evolves. Specifically, the row spacing is limited and may not be completely available during the entire season for robot navigation. That is specially the case when the crop reaches the canopy closure stage, where the space between rows will disappear. According to the growth stage of the crop, two different approaches can be used to calculate the distance between location pairs and give numerical values to each section of the route (Figure \ref{fig:navDistancesExplain}).

In early crop growth stages, row spacing would potentially be accessible, and the robotic platform could safely navigate between rows (i.e., between-rows ($BR_{nav}$) navigation). Under this approach, the distance between scan location pairs can be computed as the L1 norm ($p=1$) or Manhattan distance, sometimes also denoted as taxicab distance \citep{Krause1986}. This distance is equivalent to the distance measured along axes at right angles. Using Equation \ref{eq:distance} with $p=1$, the distance between two scan locations $a$ and $b$ at coordinates $a = (a_{x}, a_{y})$ and $b = (b_{x}, b_{y})$ will be given by:
\begin{equation}
    d(a,b)_{BR_{nav}} = |a_{x}-b_{x}| + |a_{y}-b_{y}|
\end{equation}
When the crop gets close to the canopy closure stage, robot navigability between the rows will be greatly impacted. Passing through the field using the row spacing without damaging the crop will not be possible. In this stage, only alleys and headlands will be available for the robot to navigate the field. Therefore, an alley-headland-alley ($AHA_{nav}$) approach must be used for navigation. Under this approach, the distance between scan location pairs can be computed by combining the L2 distance (Euclidean distance) to the nearest headland, and the Manhattan distance from the headland position to the next scan location. If we define an intermediate waypoint $h = (h_{x}, h_{y})$ located at the headland in such a way that $a_{x} = h_{x}$, the distance between two scan locations $a$ and $b$ at coordinates $a = (a_{x}, a_{y})$ and $b = (b_{x}, b_{y})$ for the $AHA_{nav}$ approach can be computed using the following equation:
\begin{equation}
    d(a,b)_{AHA_{nav}} = d(a,h) + d(h,b) = \sqrt{(a_{y}-h_{y})^2} + |h_{x}-b_{x}| + |h_{y}-b_{y}|
\end{equation}

\subsubsection{Navigation route optimization} \label{route_optimization}
In general, to effectively automate the TLS survey we seek a continuous route that connects all scan locations for the robot platform to follow. A simple nearest neighbor algorithm, where from every scan location the platform always moves to the nearest unvisited location, could be used to obtain a feasible route. However, while the selection of the nearest location at each step may be optimal, it doesn't guarantee that the global solution is the most optimal route for the entire survey. This algorithm may lead to suboptimal paths that do not produce the shortest overall distance. If we further constrain the route to be the route that visits all scan locations using the minimum distance, the problem can be formulated as a travelling salesman problem (TSP): given a list of scan locations and the distance between each pair of scan locations, the mobile platform must visit all of them and return to the starting location using the shortest route. 

The TSP problem can be modeled using a linear function in which the decision variables are constrained to take integer values \citep{Dantzig1954,Miller1960}. Therefore, integer programming can be used to get an approximation to the solution for these kinds of problems. Given a connected undirected graph $G=(V,E)$, with edge weights $w(e)=w_{ij}$ as the distance from location i to location j, and defining $x_{ij}$ as a binary variable with value of 1 if the edge $E_{ij}$ is included in the path, the solution to the TSP problem aims to find the closed cycle that contains all nodes $n$ with minimal length such as:
\begin{equation} \label{eq:TSP}
    min \sum_{i}\sum_{j} w_{ij} x_{ij}; ~i=1,\dots,n,~j=1,\dots,n
\end{equation}
constrained by the requirements that each scan location can be left behind exactly once (i.e., from each node in the graph only one outgoing edge is allowed): 
\begin{equation} \label{eq:const1}
        \sum_{j} x_{ij} = 1;~i=1,\dots,n 
\end{equation}
and that each scan location can be visited exactly once from one other scan location (i.e., at each node only one incoming edge is allowed):
\begin{equation}\label{eq:const2}
        \sum_{i} x_{ij} = 1;~j=1,\dots,n \\
\end{equation}

These two constraints, however, do not prevent the presence of closed routes that do not include all the vertices of the graph (i.e., subtours). Having into account that for any given subtour the sum of edges will be equal to the number of vertices in it, a third constraint can be added to ensure that the solution to the problem is a fully connected tour and not the union of smaller subtours. Using the Dantzig–Fulkerson–Johnson formulation \citep{Dantzig1954}, we can impose the condition that the sum over all edges in a subset of the graph be less than the number of vertices in the subset. In this way, the subset cannot form a proper subtour, and hence the solution to the problem will be a fully connected tour and not the union of smaller subtours.

Being \textit{V} the set of all nodes in the graph, and \textit{S} a subset of \textit{V}, the subtour elimination constraint can be expressed as follows:
\begin{equation}
 \sum_{i\in S}\sum_{j\in S} x_{ij} \leq |S|-1; ~S \subset V, ~|S| \geq 2
\end{equation}

The solution to this problem will provide the optimal sequence of scan locations that minimizes the distance traveled by the robotic platform during the in-field TLS data collection.

\subsection{Autonomous field navigation}
To conduct the TLS survey autonomously, the robot must navigate the field visiting all the scan locations in the optimal set without requiring any input from humans. Robot navigation refers to the problem of moving a mobile platform in the environment following a path free of obstacles. It involves tasks related to mapping, path planning, and path tracking. For ensuring the success of the TLS operation, once the optimal sequence of scan locations has been identified, valid navigation paths connecting consecutive scan locations needs to be created and efficiently followed by the robotic platform.

\subsubsection{Point-to-point route decomposition} \label{route_brakdown}
The navigable space through the field (see Section \ref{fieldDescription}) will provide a grid-like structure of obstacle-free paths that the phenotyping platform can leverage to navigate efficiently. Taking advantage of this grid distribution of navigable paths, the route between two consecutive scan locations can be decomposed into a set of straight paths parallel to both the horizontal and vertical axis in such a way that only straight trajectories and in-place rotations will be needed for field navigation.

\begin{figure} [H]
    \centering
    \includegraphics[width=\textwidth]{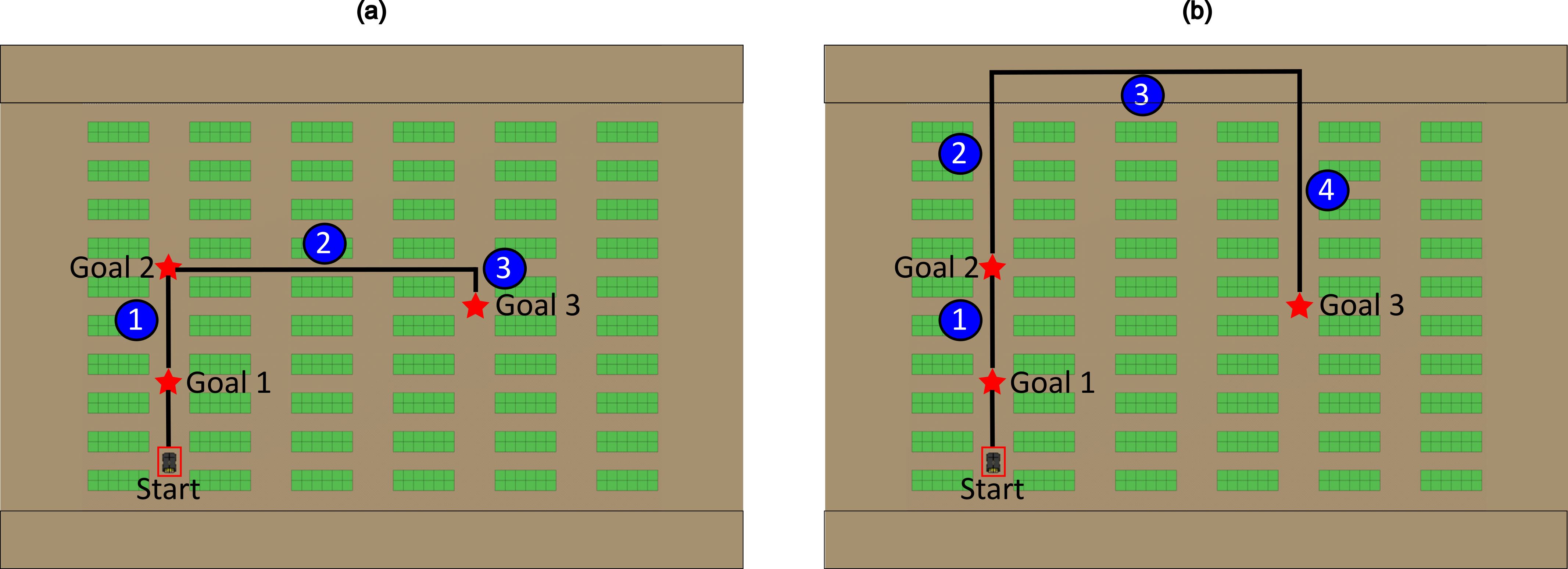}
    \caption{Route decomposition approaches. (a) Between-rows navigation ($BR_{nav}$) steps: 1- goals in same alley; 2, row navigation to change alleys; 3, navigation to goal. (b) Alley-headland-alley navigation ($AHA_{nav}$) steps: 1- goals in same alley; 2- navigation to the nearest headland to change alleys; 3- headland navigation to change alleys; 4- navigation to goal.}
    \label{fig:route_planning}
\end{figure}

In Section \ref{navigationRoute} two different approaches to calculate the distance of the paths between scan locations were introduced. In both approaches, alleys and headlands were considered usable for the robot to navigate. However, the use of row spacing differs between approaches. The $BR_{nav}$ (between-rows) approach allows calculating navigation paths using the row spacing, while the $AHA_{nav}$ (alley-headland-alley) approach will forbid it. These two approaches will lead to two ways of decomposing the navigation route (Figure \ref{fig:route_planning}), but both will be agnostic to the robot planner used for robot navigation.

In general, two different situations can be identified regarding the position of consecutive scan locations for field navigation: scan locations located in the same alley and scan locations in different alleys. When the scan locations are placed in the same alley, the robot will be able to navigate directly from one location to the next using the current alley. This can be observed in step number 1 in both Figure \ref{fig:route_planning} (a) and Figure \ref{fig:route_planning} (b). On the other hand, when the next scan location is located in a different alley, the robotic platform will need to perform a change of alleys. For the $BR_{nav}$ approach, the platform will use the closest space between rows to navigate to the alley where the next scan location is located (step number 2 in Figure \ref{fig:route_planning} (a)). Finally, the robot will reach the goal using the current alley (step number 3 in Figure \ref{fig:route_planning} (a)). For the $AHA_{nav}$ approach, the robot will navigate to the closest headland first (step 2 in Figure \ref{fig:route_planning} (b)). Then, the robot will move to the alley where the next scan location is located using the headland as a corridor (step 3 in Figure \ref{fig:route_planning} (b)). Finally, the platform will move from this position to the goal location using the current alley (step 4 in Figure \ref{fig:route_planning} (b)).

\subsubsection{Path planning} \label{path_planning}
Path planning aims to solving the problem of finding a collision-free path for the robot to navigate from one location in the field to another. Given the robot dimensions and the obstacles in the navigation space, the solution to this problem provides a sequence of positions that joins the initial and goal locations. Our approach for field navigation implemented a point-to-point navigation based on waypoints. Following the approach introduced in Section \ref{route_brakdown}, the route between scan location pairs can be decomposed in more basic straight paths (i.e. subpath). Each subpath of the navigation route will consist of a straight path between a starting waypoint and a goal waypoint that the robot needs to visit.

In general, path planning for robot navigation involves two main tasks: global planning and local planning. The global planner is responsible for finding, at a high level, the optimal path for the navigation stack to follow based on prior knowledge of the environment. The local planner tracks the global path and, if needed, recalculates the robot trajectory locally based on additional information provided by the navigation sensors. To implement autonomous navigation between waypoints, we used the ROS Navigation Stack (ROS \textit{navstack}) depicted in Figure \ref{fig:nav_stack}. The ROS \textit{navstack} uses the \textit{move\_base} package to compute the actions needed to fulfill the mission. This package interconnects the global and local path planners to reach the waypoint goals efficiently.

\subsubsubsection{Global path planning}
The global path planning seeks to construct a valid path between locations based on global information of the environment. For the global planner, we implemented a straight line planner for the \textit{move\_base} package. The straight line planner, which adhered to the textit{nav\_core::BaseGlobalPlanner} interface, computed a straight trajectory to connect the starting and the goal waypoints of each subpath. The final heading at the goal waypoint was calculated as the direction of the vector that links the current goal location and the goal location for the next subpath. That way, when the robot successfully reached the current goal it rotated on the spot to face the next goal.

\begin{figure} [H]
    \centering
    \includegraphics[width=\textwidth]{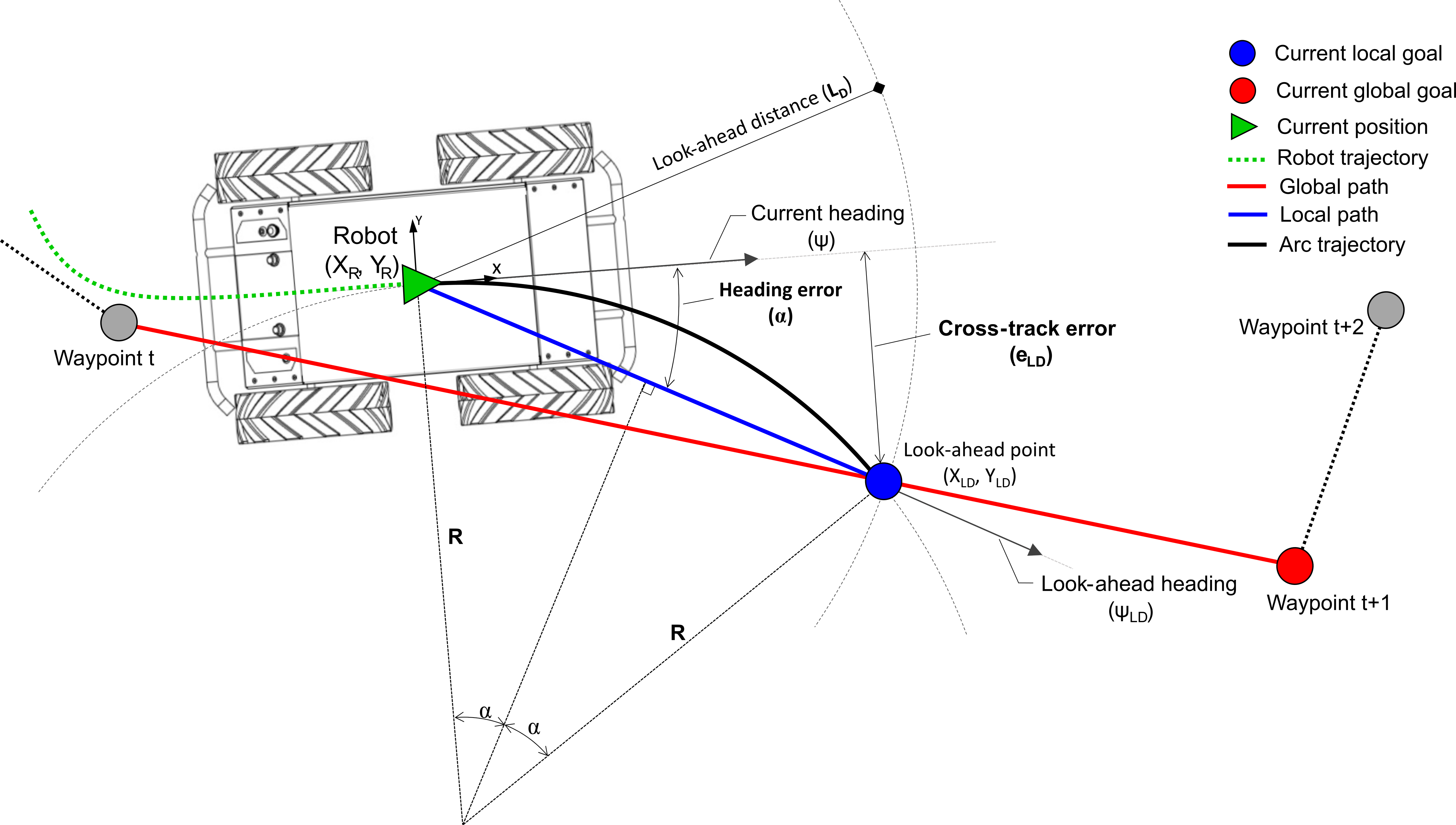}
    \caption{Pure Pursuit algorithm principle.}
    \label{fig:purePursuit_principle}
\end{figure}

\subsubsubsection{Local path planning}
The local path planning aims to generate a trajectory to guide the robot through the global navigation path based on current information about the state of the platform and the environment. The local planner was in charge of enabling the robot to follow the global path between waypoints. The local planning control was achieved using a planner based on the pure pursuit path tracking algorithm \citep{Coulter1992}. The pure pursuit algorithm calculates the arc trajectory needed for the robot to move from the current position to a virtual goal position on the global path located some distance ahead (i.e., look-ahead point). The look-ahead point ($X_{LD}$, $Y_{LD}$) can be computed by finding the intersection point of the circle of radius look-ahead distance (\textit{$L_{D}$}) centered at the robot’s location in the global coordinate system ($X_{R}$,$Y_{R}$), and the global path (Figure \ref{fig:purePursuit_principle}).

The pure pursuit planner computes the radius \textit{R} of the arc trajectory that joins the current robot's location with the look-ahead point located on the global path. Applying the law of sines to Figure \ref{fig:purePursuit_principle} results in:
    \begin{gather*}
    \frac{L_{D}}{\sin({2\alpha})} = \frac{R}{\sin(\frac{\pi}{2} - \alpha)} \\
    \frac{L_{D}}{2\times \sin(\alpha) \times \cos(\alpha)} = \frac{R}{\cos(\alpha)} \\
    \end{gather*}
    
Therefore, the radius of the arc trajectory will be given by:
\begin{equation} \label{eq:purePursuit_rad}
    R = \frac{L_{D}}{2\times \sin(\alpha)}
\end{equation}

The curvature of the arc will be the reciprocal of the radius:
\begin{equation} \label{eq:purePursuit_curv}
    \textit{k} = \frac{2\times \sin(\alpha)}{L_{D}}
\end{equation}

This curvature will determine the control law needed for the vehicle to follow the arc trajectory from the current location. At each control cycle, the pure pursuit planner calculated the curvature and computed the linear and angular velocities to make the robot follow the arc trajectory. The pure pursuit planner was implemented as a local planner for the \textit{move\_base} package as in \cite{Xu2022Amodular} and adhered to the \textit{nav\_core::BaseLocalPlanner} interface. During our field experiments, we set the look-ahead parameter to 1 meter.

\section{PERFORMANCE EVALUATION}

\subsection{Experimental fields} \label{ExperimentalFields}
\begin{figure} [ht]
    \centering
    \includegraphics[width=\textwidth]{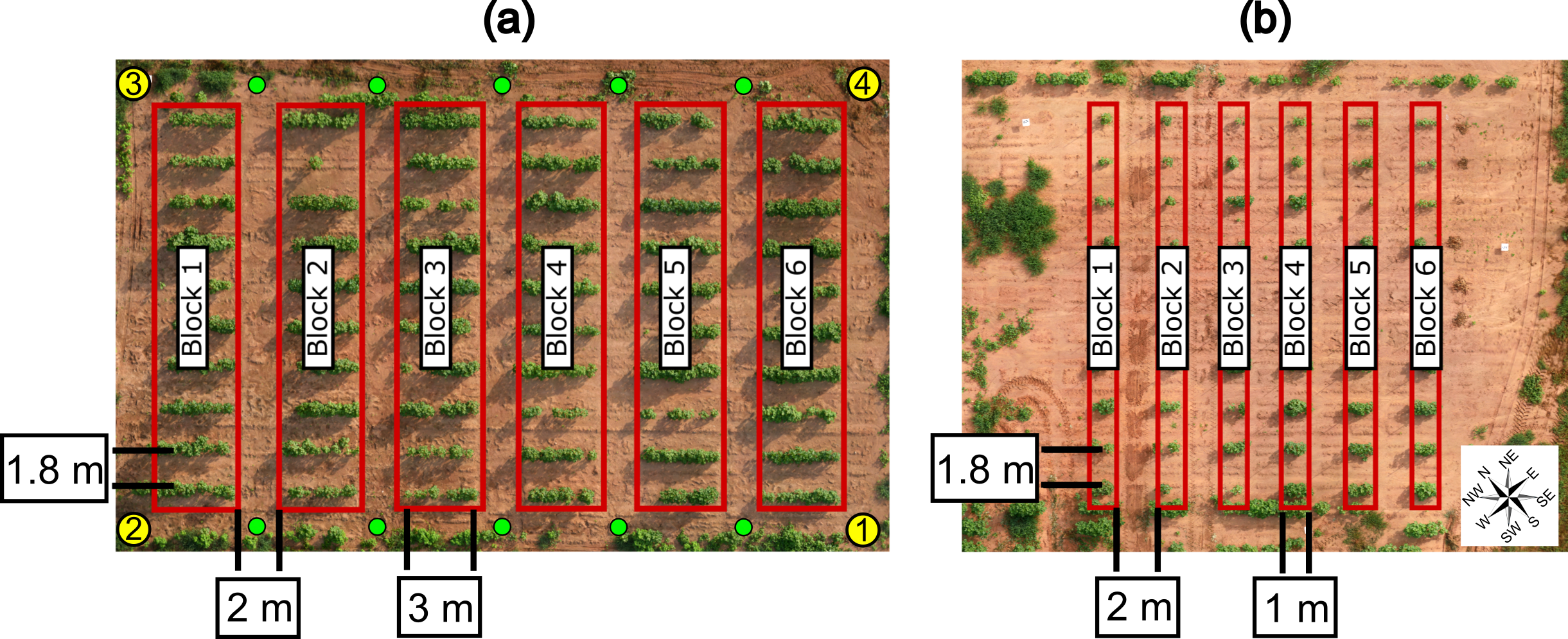}
    \caption{Experimental fields. (a) Aerial view of the engineered (ENGR) breeding field with wide row spacing. Numbers in yellow circles indicate waypoint sequence for border navigation experiment. Green dots indicate waypoints used for serpentine navigation experiment (b) Aerial view of the single plant layout (SPL) field.}
    \label{fig:fields}
\end{figure}

To test and validate our autonomous phenotyping system, two different trial fields were used. The fields were part of a multi-year and multi-location cotton breeding trial and were located at the Iron Horse Farm in Greene County, Georgia, U.S. (33$^{\circ}$43'01.3"N 83$^{\circ}$18'29.1"W). A total of 10 different cotton genotypes 
were planted on the fields for genotype comparison. Two different field layouts were used to test and validate our autonomous phenotyping system (Figure~\ref{fig:fields}).

The first field was an engineered field (ENGR) for breeding purposes. The crop was arranged in plots following a wide row (72-inch rows) planting pattern (Figure~\ref{fig:fields} (a)). The dimensions of this field were 30 m $\times$ 20 m (length $\times$ width). The ENGR field was composed of 60 individual plots following a randomized block design for 10 genotypes and 6 repetitions per genotype. A total of 40 seeds of each genotype were sowed into plots of 3 meter long and thinned to less than 20 plants after crop establishment. Plots were arranged into 6 blocks with 10 plots per block. The distance between plots centers in each block was set to approximately 1.8 meters (i.e., wide-row spacing), while the distance between blocks was set to 2 meters. This field was originally intended for conducting phenotyping experiments based on tractor-mounted proximal sensors, and hence the selection of a wide-row spacing. The second field consisted of a single plant layout (SPL) field with 10 cotton genotypes and 6 repetitions arranged into blocks (Figure~\ref{fig:fields} (b)). The dimensions of this field were 20 m $\times$ 20 m (length $\times$ width). A total of 5 seeds were planted in 1 meter long plots and thinned to 1 plant after germination to ensure an approximate distance of 2 meters between blocks (individual plants in the same row). The distance between plant centers in each block was set to approximately 1.8 meters. This field was intended for architectural traits analysis, where an individual plant layout can help reduce overlapping and occlusion issues between different plants. 

To facilitate posterior comparisons for performance analysis, five 200 mm diameter Koppa registration spheres (Koppa Target Spheres, California, USA) were deployed throughout the field for helping co-register the point clouds collected from multiple locations into a common coordinate system. The targets were installed on top of aluminum rods with a height between 1.4 m and 2 m. At the beginning of each test the spherical targets were placed in the field prior conducting the autonomous TLS survey.

\subsection{Navigation performance evaluation experiments}
To evaluate the navigation performance of our autonomous TLS-based phenotyping platform during the experiments, we made a distinction between non-stationary performance and stationary performance. Non-stationary performance refers to the performance of the platform during autonomous field navigation and accounts for the accuracy of the path-following task. Stationary performance refers to the performance of the platform while the robot is stationary during the LiDAR scanning process at each scan location.

\subsubsection{Non-stationary performance}
To characterize non-stationary performance during autonomous field navigation, two experiments were performed on the ENGR field (Figure~\ref{fig:fields}, (a)). The first experiment used the corners of the field as waypoints to navigate around the field (yellow circles numbered 1 to 4 in Figure~\ref{fig:fields}, (a)). Starting at the corner number 1, the robot moved clockwise visiting corners 2, 3 and 4 sequentially, to finally return back to the starting position. At each corner the robot executed a 90$^\circ$ clockwise in-place rotation to face the next waypoint. The second experiment used the alleys and headlands to navigate through the crop in a serpentine manner. Two full passes per each direction (i.e., right to left and left to right), were carried out. For this experiment, the intersections between alleys and headlands (green circles in Figure~\ref{fig:fields}, (a)) were used as the waypoints for the navigation path. The robot traveled 12 sub-paths (i.e., path joining consecutive waypoints) and performed 12 in-place rotations per direction. For both experiments, the robot was commanded to navigate the path between waypoints following a straight trajectory based on the plan calculated by the global path planner. All ROS topics related to robot navigation, including state estimations for location (X, Y) and heading ($\psi$) from the \textit{robot\_localization} package, were recorded during field navigation using the \textit{ROS bag} file format for further processing. The cross-track error (XTE) and the heading error for each path, as well as the standard deviation (SD) for these two errors were computed to evaluate the navigation accuracy.

\subsubsection{Stationary performance}
To evaluate the stationary performance of the platform during the autonomous TLS survey, we analyzed pose statistics at the LiDAR sensor level, specifically focusing on the location (X, Y) and heading ($\psi$) for each scan location. The primary goal of this analysis was to gain a comprehensive understanding of the accuracy and consistency of our platform's performance throughout the TLS survey.

A total of 11 scan locations were used to collect point clouds from the ENGR field. A total of 7 scan locations were used to collect TLS data from the SPL field. The distribution of scan locations was based on the optimized layout calculated by our methodology, and included both scan locations at the border of the field and inside the field. Location and orientation data were collected during the laser scanning operation while the platform was stationary at each scan location. The robot pose information was stored in a ROS \textit{bag} file while the LiDAR scanner was performing the scan. This information was analyzed offline to extract the location and orientation of the LiDAR scanner (X, Y, Z, and heading) and to calculate positioning statistics.

To quantify the absolute positioning errors, we computed the distance between the planned scan point coordinates and the platform coordinates recorded during the scanning process at the corresponding location. Additionally, to obtain insights into the precision and variability of our system's performance on pose estimation, we examined the residuals for all three components in relation to the estimated sensor location at each scan point. This estimation was achieved by averaging the platform coordinates recorded during the scanning process at each specific location.

\subsection{TLS automation experiments}
\begin{figure} [h]
    \centering
    \includegraphics[width=\textwidth]{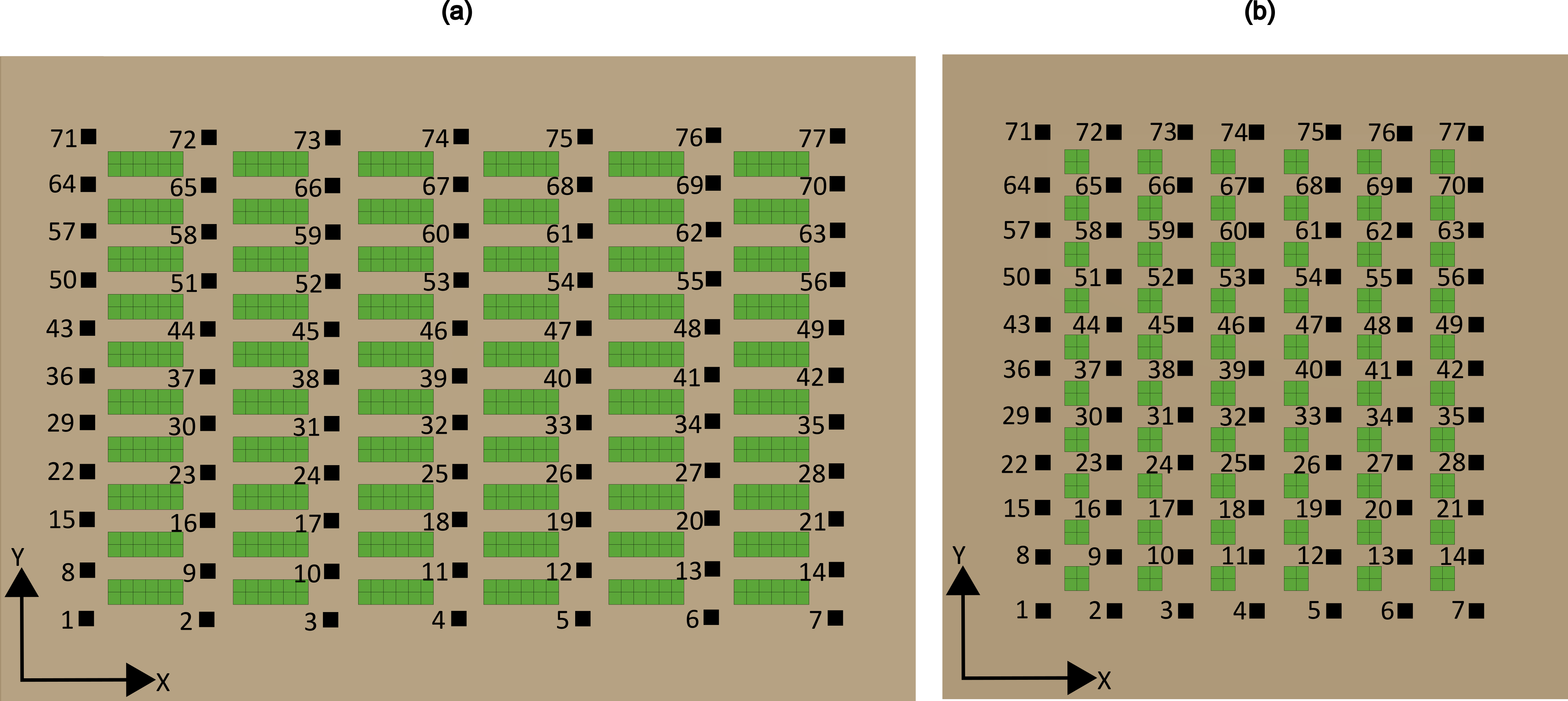}
    \caption{Digital model of the fields for ray-box intersection analysis. (a) ENGR field. (b) SPL field. Plots are represented as green boxes. Potential scan locations are identified as black squares}
    \label{fig:matlab_digitized}
\end{figure}

To evaluate the performance of our methodology at collecting TLS data in a real field, we conducted a TLS survey session using the autonomous phenotyping platform in both the ENGR and SPL fields. We first modeled the field on MATLAB R2021a (The MathWorks Inc., Massachusetts, United States) using the parameters of the field described in Section \ref{ExperimentalFields} (i.e., number of rows, number of plots, row spacing, and alleys width). The dimensions of the plots (length, width, and height) were established based on a mature growth stage (i.e., near canopy closure). The ENGR field (Figure \ref{fig:matlab_digitized}, (a)) was composed of 10 rows with 6 plots per row. Each plot was defined by a box with dimensions 3 $\times$ 1 $\times$ 1.9 (L $\times$ W $\times$ H) given in meters. Using voxels of 0.5 $\times$ 0.5 $\times$ 0.33 meters (length $\times$ width $\times$ height) for discretizing the bounding boxes, each digitized  plot was composed of 72 cells. The SPL field (Figure \ref{fig:matlab_digitized}, (b)) was composed of 10 rows with 6 individual plants per row. Each plant was defined by a box with dimensions 1 $\times$ 1 $\times$ 1.9 (L $\times$ W $\times$ H) in meters. Again, using the same voxel size of 0.5 $\times$ 0.5 $\times$ 0.33 meters, each digitized  plot in the SPL field was composed of 48 cells. Given the field layout and the dimensions of the plots, the effective free space available for robot navigation at this growth stage was 0.8 meters for row spacing and to 1.5 meters for alley width, enough space for the robot to navigate safely.

As indicated in Section \ref{fieldDescription}, the potential scan locations for the TLS survey were limited to the free space were the robot can freely navigate. The plots distribution on the field and the growth stage of the crop reduced the feasible space for placing the scanner to the intersections points between row spacing, alleys and headlands. In addition, to reduce the impact of the minimum range limitation of the LiDAR scanner on the quality of the collected data, only the scan locations whose distance to the crop was greater than 0.8 meters were considered valid for the TLS survey. Therefore, only the intersections between alleys and free space between rows were considered as valid scan locations. For each field, a total of 77 field positions were defined as potential scan locations (black squares in Figure \ref{fig:matlab_digitized}).

In order to compute crop visibility from each potential scan position, the AABB–ray intersection algorithm was implemented on MATLAB using ray casting approaches. LiDAR sensor parameters for ray casting analysis, such as start and end angles, and angular resolution were selected according to the capabilities of the FARO Focus S70 LiDAR scanner. Specifically, vertical angle was configured between -60$^\circ$ and 90$^\circ$, and horizontal angle was configured between 0$^\circ$ and 360$^\circ$. To accelerate code execution and avoid potential memory issues, the angular step was set to 0.36$^\circ$ for both the vertical and horizontal angles, which translates to a laser scanner resolution of $\frac{1}{4}$ (i.e., 6.1 mm over 10 meters). The MATLAB script provided as outputs the number of rays that impacted each cell in the field, the distance from the origin of the ray to the hitting point, the row and plot indices of the plots intercepted per each ray, and the cell index. With this information the visibility score table including the number of cells covered from each scan location was obtained. Finally, the optimal set of scan locations for the TLS survey was identified using Algorithm \ref{alg:greedy}  according to the visibility score.

To optimize the route between scan location pairs from the identified set, the MATLAB Optimization Toolbox \citep{MatlabOTB} was used to solve the TSP problem with subtour elimination constraints (Section \ref{route_optimization}) for both route definition approaches ($BR_{nav}$ and $AHA_{nav}$) described in Section \ref{navigationRoute}.

\subsection{Point clouds registration performance analysis}
In order to evaluate the effectiveness of our system in automating the registration of point clouds, we used the data from the navigation sensors to position and align autonomously collected point clouds from multiple field locations. This involved determining the position ($X_{LiDAR}$, $Y_{LiDAR}$, $Z_{LiDAR}$) and the orientation ($\psi_{LiDAR}$) of the LiDAR sensor relative to its origin. We achieved this by computing these values through the conversion of state estimates obtained from the ROS bag files in the world frame into the LiDAR scanner frame. The homogeneous transformation described in Equation \ref{eq:fullTransform} facilitated this transformation. Additionally, the pitch and roll angles, required to fully characterize the orientation of the LiDAR scanner at each scan location, were directly acquired from the device's internal dual-axis compensator.

To achieve a consistent coordinate system for the collected scans from different locations, the pose information at the LiDAR sensor level was associated with each scan using FARO SCENE software (FARO Technologies, Florida, US). Subsequently, a cloud-to-cloud (C2C) registration was performed using the same software. C2C registration is an automated procedure that enables the alignment of point clouds based solely on overlapping points, without the need for artificial targets. 

Following the C2C registration, the aligned point cloud obtained from this procedure was compared to the point cloud registered using the spherical targets (known as target-based registration), which served as the reference for benchmarking. In order to assess the accuracy of the registration, various metrics were computed. These metrics included target-based statistics, such as the Euclidean distance between corresponding targets, as well as scan point error statistics. To ensure a representative analysis, the scan point error statistics were computed on a subset of points that were uniformly subsampled from each individual scan within the complete point cloud.

In order to evaluate the performance of the automatic registration process, the Hausdorff distance \citep{Rote1991} was adopted as the benchmarking metric. This distance serves as a dissimilarity measure for comparing two sets of points. Specifically, it quantifies the maximum distance between any point in one set (source) and its nearest point in the other set (reference). Given a source set of points $PC_{src}=\{a_1,a_2,...,a_n\}$ and a reference set of points $PC_{ref}=\{b_1,b_2,...,b_n\}$, the Hausdorff distance from the source to the reference will be given by the following equation:
\begin{equation}    
    d_{H}(PC_{src},PC_{ref}) = \max_{a\in PC_{src}}\left(\min_{b\in PC_{ref}}\left(d(a,b)\right)\right),    
\end{equation}

where \textit{a} and \textit{b} represent the 3D points belonging to the point clouds $PC_{src}$ and $PC_{ref}$ respectively, and d(a, b) is the Euclidean distance between these two points.

\section{RESULTS AND DISCUSSION}   
\subsection{Field navigation evaluation}

Overall, the comprehensive analysis of cross-track error (XTE) and heading errors confirm the system's capability to maintain precise trajectories during field navigation, demonstrating the platform's reliability and accuracy in autonomous operation.

\subsubsection{Non-stationary error analysis}
\begin{figure} [h]
    \centering
    \includegraphics[width=\textwidth]{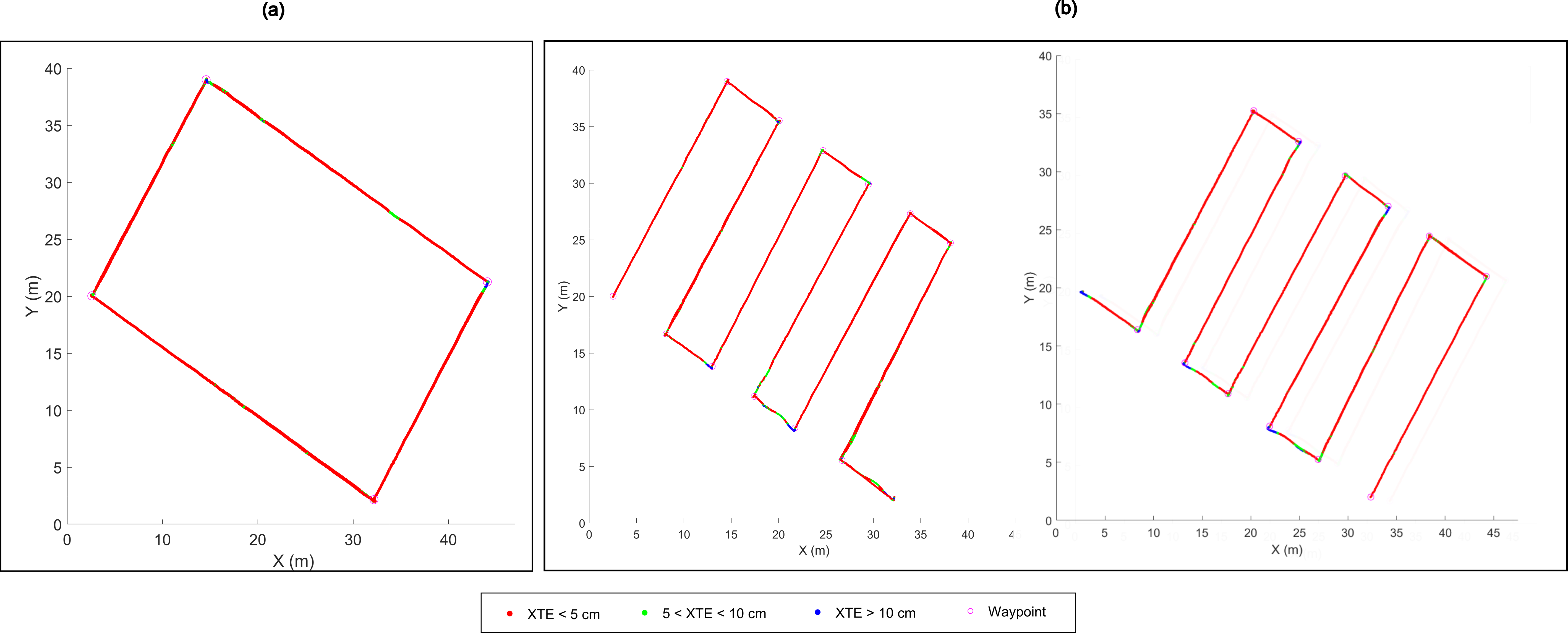}
    \caption{Cross-track error (XTE) observed during the navigation experiments. (a) Border navigation experiment. (b) Serpentine navigation experiment. The color-coded lines indicate the range of error encountered during navigation in relation to the global plan. The waypoints used for autonomous navigation were located at the headlands areas and are marked as circles.}
    \label{fig:navigation errors}
\end{figure}

During the border navigation experiment, the robotic platform traveled a total distance of 970 meters. Throughout the test, the XTE remained below 5 cm for the majority of the path, accounting for over 86\% of the path (Figure \ref{fig:navigation errors} (a)). The average XTE value recorded was 2.370 cm.

In the serpentine navigation experiment, the platform completed 170 meters per pass, accumulating a total of 680 meters after four passes. Similar to the previous navigation experiment, the XTE remained below 5 cm for most of the path. In the right-left direction (Figure \ref{fig:navigation errors} (b), left), the XTE error was below 5 cm for 78\% of the experiment, between 5 cm and 10 cm for 14\% of the experiment, and exceeded 10 cm for 8\% of the experiment. In the left-right direction (Figure \ref{fig:navigation errors} (b), right), the XTE error was below 5 cm for 83\% of the experiment, between 5 cm and 10 cm for a 13\%, and exceeded 10 cm for only 4\%. Across the four passes, the average XTE error measured 3.666 cm.

Overall, the XTE observations depicted in Figure \ref{fig:navigation errors} underscore the system's consistent performance in maintaining a precise trajectory during field navigation. The majority of both the border and serpentine navigation experiments showcased the platform's ability to achieve reliable and accurate autonomous navigation, with XTE values consistently below the 5 cm threshold.

During the entire navigation experiment, the average XTE was approximately 3 cm, slightly higher than expected for a standard RTK‐GPS positioning system. This disparity can be attributed to tracking errors observed during in-place rotations. The robot effectively followed the straight sections of the navigation paths (Figure \ref{fig:navigation errors}), but the in-place rotations to align the heading towards the next waypoint introduced a lateral displacement, contributing to increased XTE. Uneven traction during rotations caused by terrain irregularities and loose soil in certain areas of the field further impacted the effective center of rotation and exacerbated cross-track error, particularly at waypoints requiring 90$^{\circ}$ turns, as can be seen on the corners of Figure \ref{fig:navigation errors}.

Despite these challenges, the robotic platform demonstrated the ability to quickly return to the planned path, even in zones with XTE values exceeding 5 cm (southernmost paths in Figure \ref{fig:navigation errors} (b)). These zones were characterized by loose rocks that affected traction and caused slight slippage. However, the navigation errors in these areas were not excessive, and the platform efficiently regained the intended path within a short distance. It's important to note that our primary objective was accurate pose estimation at scan locations, rather than achieving high accuracy in field navigation (see Section \ref{stationaryErrors}).

\begin{figure} [H]
    \centering
    \includegraphics[width=\textwidth]{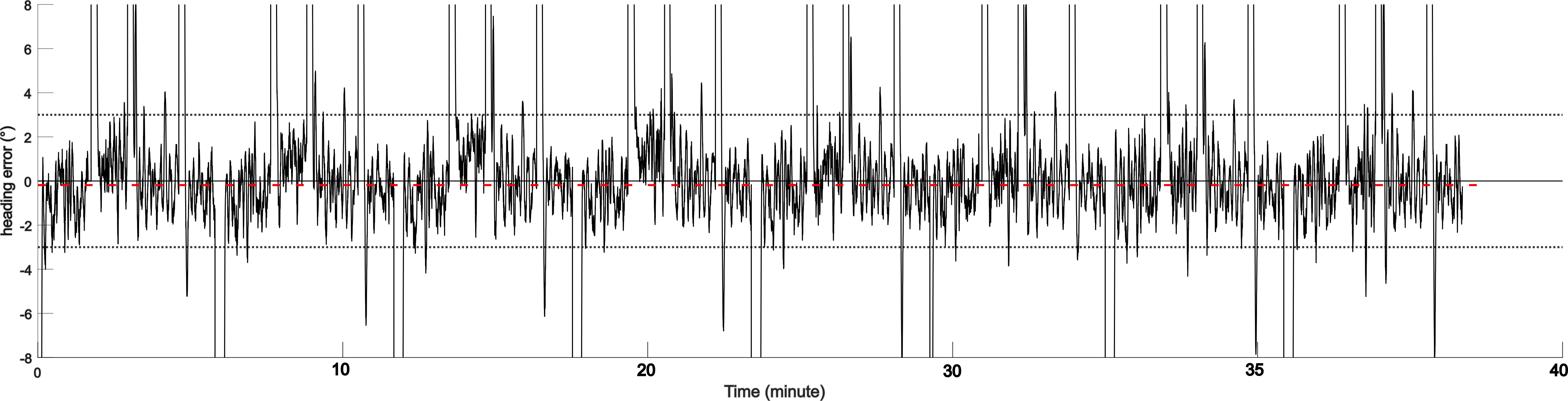}
    \caption{Heading error during the border navigation experiment. The red dash line indicates the average heading error value for the entire experiment (-0.281$^{\circ}$). The black dotted lines indicate the range of error between $\pm$ 3$^{\circ}$. High heading errors occurred when the robot was making turns at the corners of the field.}
    \label{fig:heading_error}
\end{figure}

Heading error analysis during field navigation revealed that the average heading error for the entire border navigation experiment was -0.281$^{\circ}$ (Figure \ref{fig:heading_error}). Once the robot completed the rotation towards the desired heading, it consistently maintained its heading within a range of $\pm$ 2.5$^{\circ}$. However, the serpentine navigation experiment showed an average heading error more than six times  higher than that of border navigation, with an average error of -1.516$^{\circ}$. This discrepancy can be attributed to larger errors accumulated during the in-place rotations to align the platform's heading. The significant difference between the current and objective headings during these rotations led to a temporary increase in heading error. As the platform rotated, the heading error gradually diminished until it reached the path tracking heading tolerance of 2$^{\circ}$, allowing the platform to effectively fulfill its mission.

\subsubsection{Stationary errors analysis} \label{stationaryErrors}
\begin{figure} [H]
    \centering
    \includegraphics[width=\textwidth]{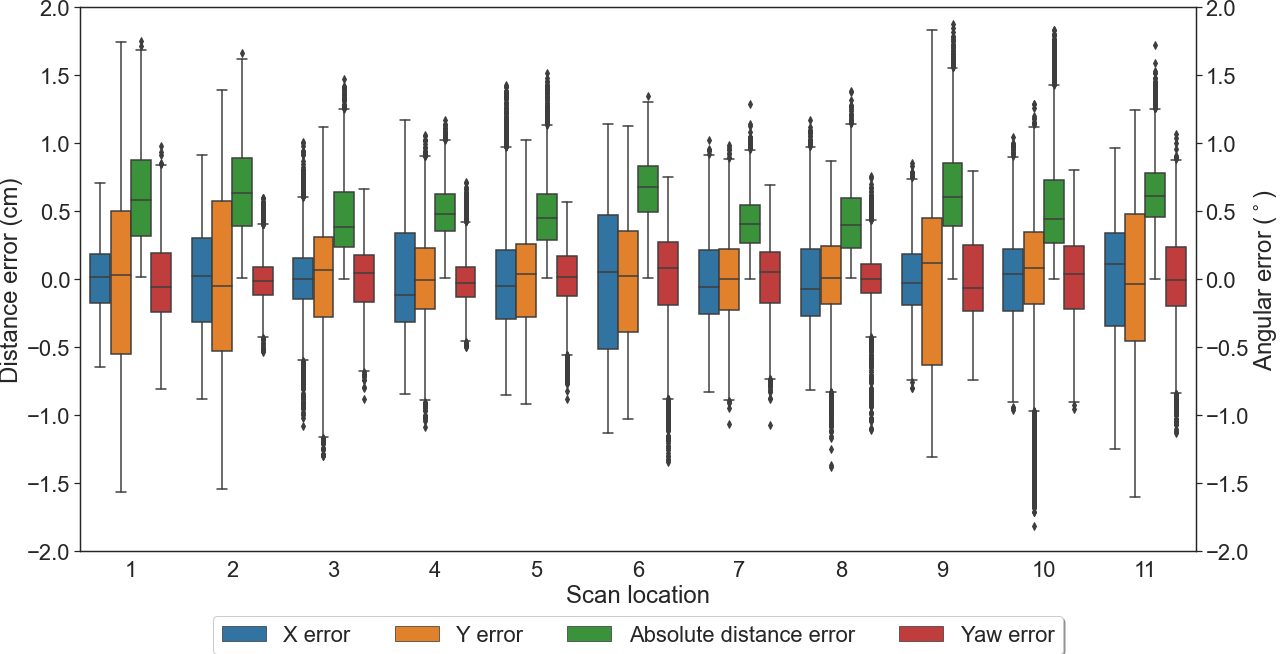}
    \caption{Statistics of location and orientation errors observed in the ENGR field. The  distribution of residuals for the X, Y and heading components is represented by the blue, orange, and red boxes, respectively. The green boxes indicate the absolute distance error between the planned scan location and the actual location.}
    \label{fig:ENGRerrors}
\end{figure}

\begin{figure} [H]
    \centering
    \includegraphics[width=\textwidth]{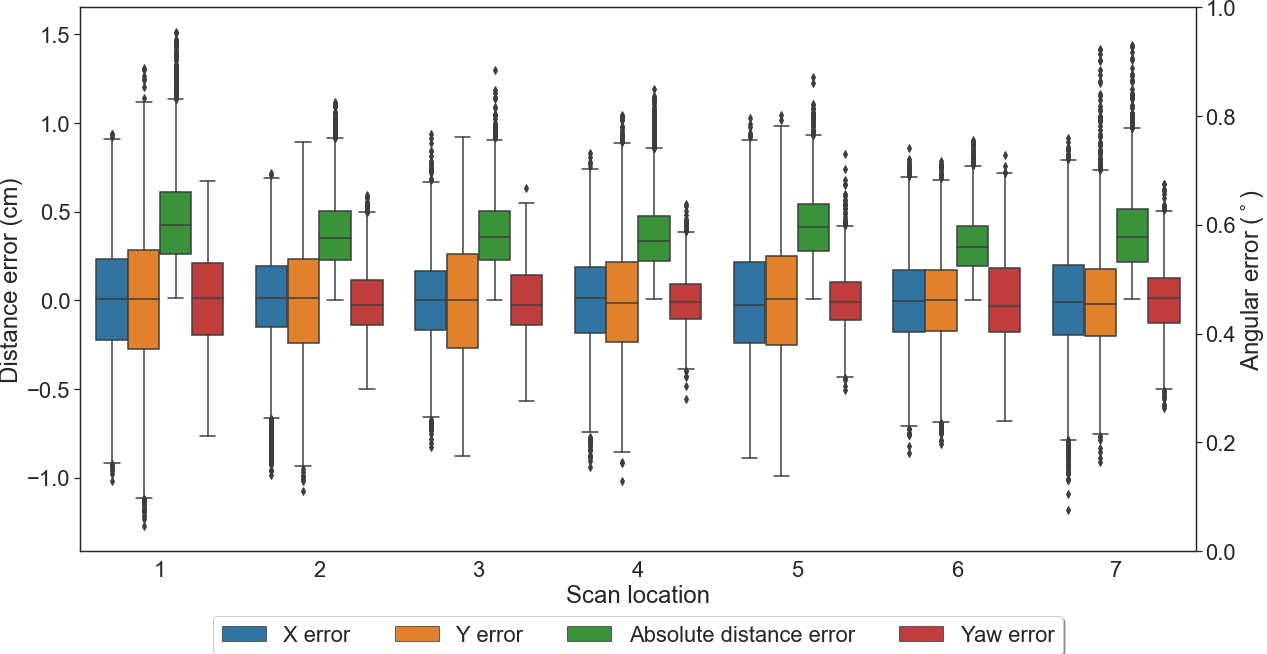}
    \caption{Statistics of location and orientation errors in the SPL field. The distribution of residuals for the X, Y and heading components is represented by the blue, orange, and red boxes, respectively. The green boxes indicate the absolute distance error between the planned scan location and the actual location.}
    \label{fig:SPLerrors}
\end{figure}

In the stationary errors analysis, we investigated the precision and accuracy of our robotic platform during the data collection process at each scan location. The robotic platform remained stationary for 7 to 10 minutes while the LiDAR scanner captured TLS data. To analyze the stationary errors, we gathered data during the laser scanning operation at each scan location and computed pose statistics at the LiDAR sensor level. The observed location errors for each scan location in the ENGR and SPL fields are depicted in Figures \ref{fig:ENGRerrors} and \ref{fig:SPLerrors}, respectively. 

For the ENGR field experiment, our system exhibited a high level of precision in both location and heading estimates. The average location errors was found to be 0.565 cm, and the average heading error was 0.035$^{\circ}$. Analyzing the residuals for X and Y coordinates, we observed a range between approximately $\pm1.25$ cm and $\pm1.75$ cm, respectively. Notably, the average residuals for X and Y coordinates were below 0.5 cm. The standard deviation values for all pose estimates during the scan operation remained consistently low, indicating the system's ability to provide accurate and consistent location estimates. Regarding heading estimates, the yaw error ranged from approximately $\pm$1$^{\circ}$ throughout the entire TLS data collection process. The low standard deviation values obtained in our tests suggest that both location and heading estimates were primarily clustered around the mean value during the scanning operation, further emphasizing the accuracy and reliability of our system.

Moving on to the TLS survey conducted on the SPL field (Figure \ref{fig:SPLerrors}), the system continued to demonstrate precise location and heading estimates. We analyzed a total of 7 scan locations in this field. The residuals for X and Y coordinates fell within the approximate ranges of $\pm1$ cm and $\pm1.25$ cm, respectively. For X coordinates, the average residuals value was 0.25 cm, while for Y coordinates, it was 0.275 cm. Similar to the ENGR field trial, the spread of values for all pose estimates during the scan operation remained below 0.25 cm for location estimates, highlighting the system's consistency in providing accurate location estimates. The average location error for the entire TLS survey session in the SPL field was 0.376 cm. As for heading estimates, the yaw error ranged from approximately $\pm$0.75$^{\circ}$ during the trial, with an average yaw error smaller than 0.005$^{\circ}$. 

Both TLS surveys exhibited low standard deviation values for location and heading estimates, indicating that the estimations were closely clustered around the mean value during the scanning operation. This analysis provides valuable insight into the accuracy and reliability of our system's positioning capabilities in both the ENGR and SPL fields, reaffirming its ability to achieve precise location and orientation during data collection in two different field layouts.

\subsection{Autonomous TLS survey results}
\begin{figure} [H]
    \centering
    \includegraphics[width=\textwidth]{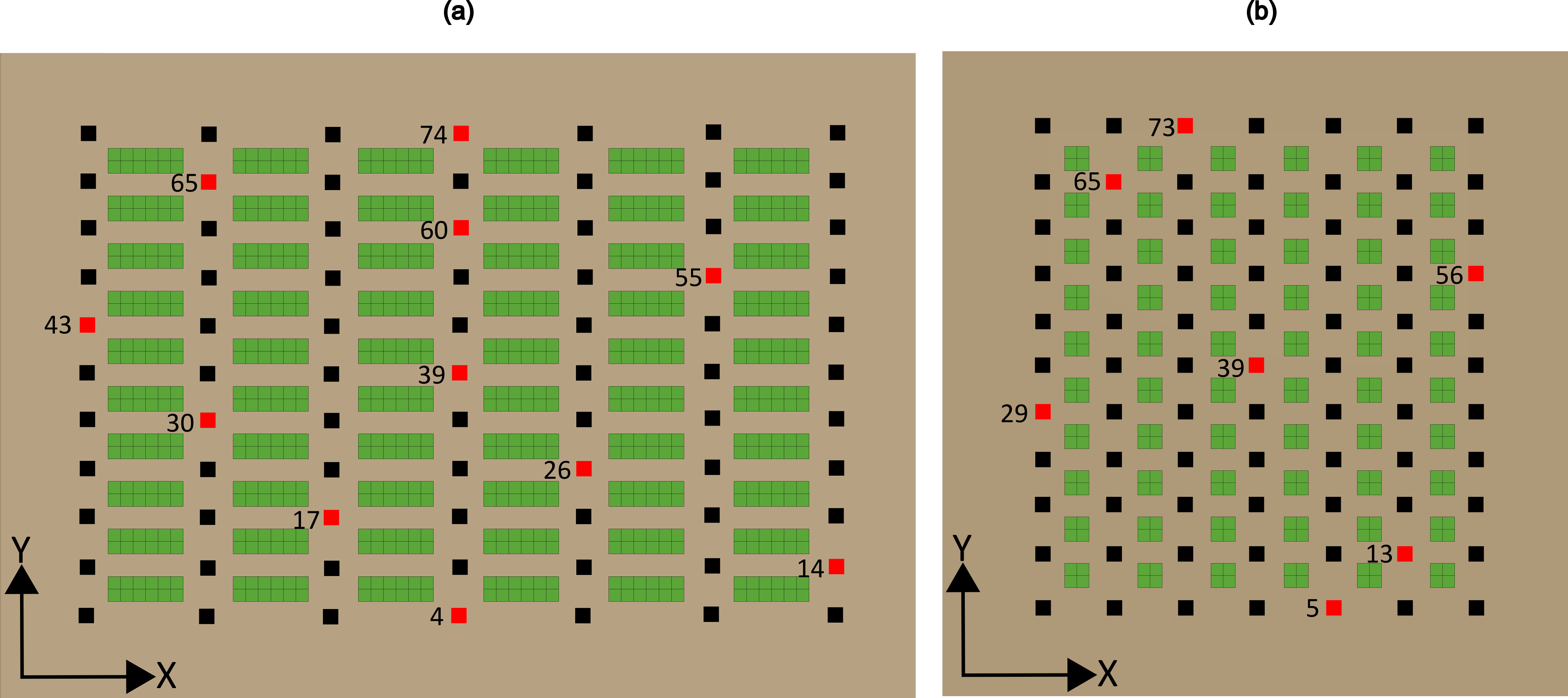}
    \caption{Optimized TLS layout obtained using ray casting approaches for visibility analysis on the ENGR (a) and the SPL (b) digitized fields. Plots are colored as green boxes. Scan locations are identified as black squares. Optimal scan locations are identified by red squares.}
    \label{fig:optimized_layout}
\end{figure}

The optimization process, combining the visibility score and the greedy algorithm (Algorithm \ref{alg:greedy}), resulted in a significant reduction in the required number of scan locations. This reduction ranged from 85\% to 90\%. For instance, in the ENGR field, only 11 scan locations were needed to cover all the cells in the digitized field, compared to the initial 77 potential scan locations that were analyzed. Similarly, for the SPL field, the number of scan locations was reduced to 7. Figure \ref{fig:optimized_layout} shows the optimized layout of scan locations for both fields.

\begin{figure} [H]
    \centering
    \includegraphics[width=\textwidth]{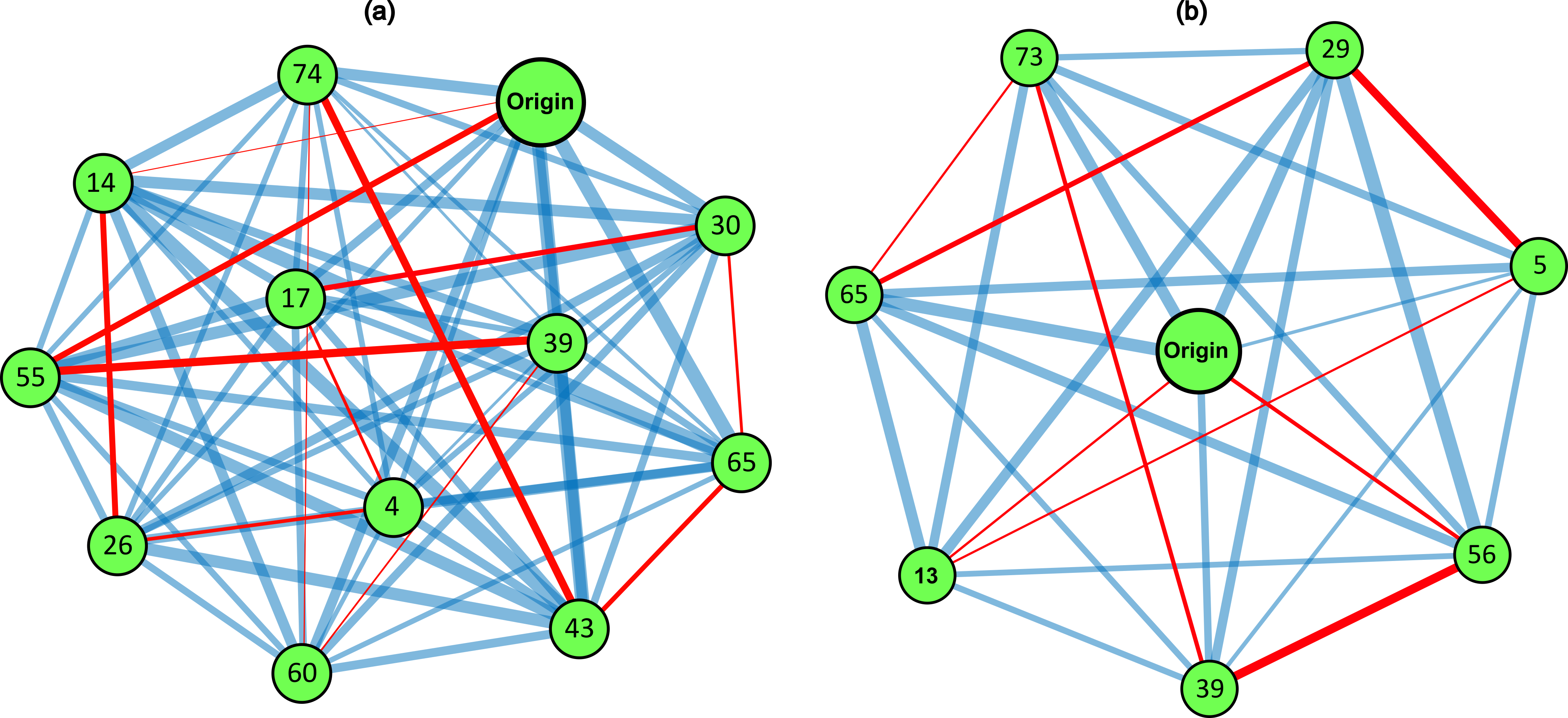}
    \caption{Weighted graph for route optimization for the ENGR field (a) and SPL field (b). The nodes of the graph identifying each scan location are represented as green circles. The edges of the graph are identified by blue lines joining each node pair. The thickness of the lines indicate the weight for each particular edge (i.e., distance between scan locations). The optimal route for both fields, the ENGR (Origin-14-26-4-17-30-65-43-74-60-39-55-Origin) and the SPL (Origin-13-5-29-65-73-39-56-Origin), are identified by red lines.}
    \label{fig:weigthed_graph}
\end{figure}

To determine the optimal route for the TLS survey, we assumed that the robotic platform would start from the bottom right corner of the field (Origin). The pairwise distances between scan locations were represented by the undirected weighted graph depicted in Figure \ref{fig:weigthed_graph}. For simplicity, the weight of the edges have been drawn using lines with variable thickness according to the distance between scan locations.

By solving the graph using the TSP solver, we obtained the optimal route for the ENGR field, as illustrated in red in Figure \ref{fig:weigthed_graph}. The optimal route can be traversed in either direction: Origin-14-26-4-17-30-65-43-74-60-39-55-Origin, or the reverse order. Both routes optimize the collection of TLS data in the field.

\begin{figure} [H]
    \centering
    \includegraphics[width=\textwidth]{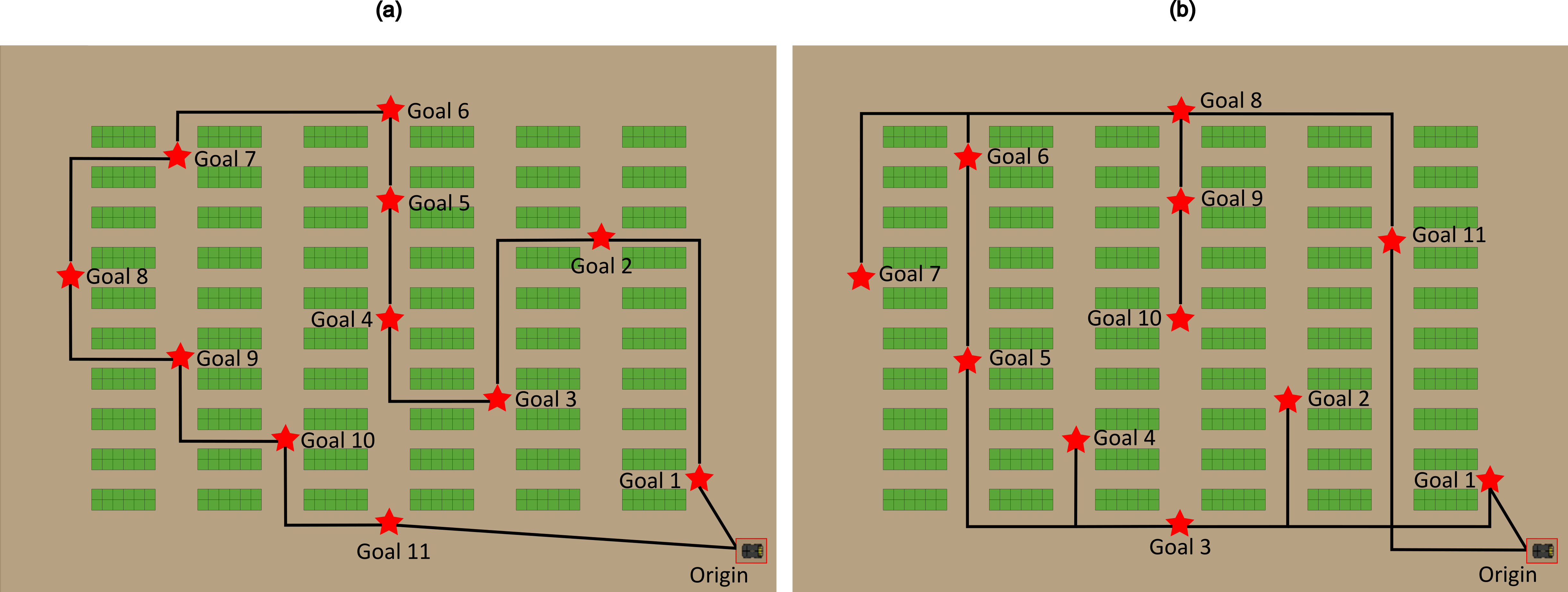}
    \caption{Optimal route for TLS-based field phenotyping. Optimized route for the $BR_{nav}$ approach (a) and the $AHA_{nav}$ approach (b). Plots are colored as green boxes. Optimal scan locations (i.e., goals for robot navigation) are identified by red stars. Black lines identified the navigation paths between scan location pairs.}
    \label{fig:optimized_path}
\end{figure}

The total cost associated with the optimal route, i.e., the total distance that the platform needs to travel during the TLS survey, using the $AHA_{nav}$ approach was determined to be 134.5 meters. For comparison, we also calculated the total navigation cost using the $BR_{nav}$ approach (Section \ref{navigationRoute}), which takes into account the space between rows during the early crop growth stage. After route optimization, the total distance to travel was reduced to 98 meters. A visual comparison of the optimal routes obtained by both approaches is provided in Figure \ref{fig:optimized_path}.

Additionally, we calculated the distance required using a simpler approach based on the nearest neighbor algorithm, where the platform would visit the closest scan location from its current position. In this case, the distance traveled by the platform increased to 146 meters. As expected, optimizing the route resulted in a reduction in the total distance needed to cover the entire field. The use of the TSP formulation for route optimizing led to a 7.8\% decrease in the traveled distance. This improvement increased to 33\% when the platform was able to navigate between the rows of the crop.

The optimization of the route is indeed crucial in minimizing the distance traveled by the robotic platform during TLS-based field phenotyping. However, it is important to recognize that route optimization is not the sole factor impacting the overall productivity and effectiveness of the phenotyping process. Autonomous field navigation can be significantly affected by the presence of obstacles and various factors that introduce variability into the field.

\subsection{Point cloud registration results}
\begin{figure} [H]
    \centering
    \includegraphics[width=\textwidth]{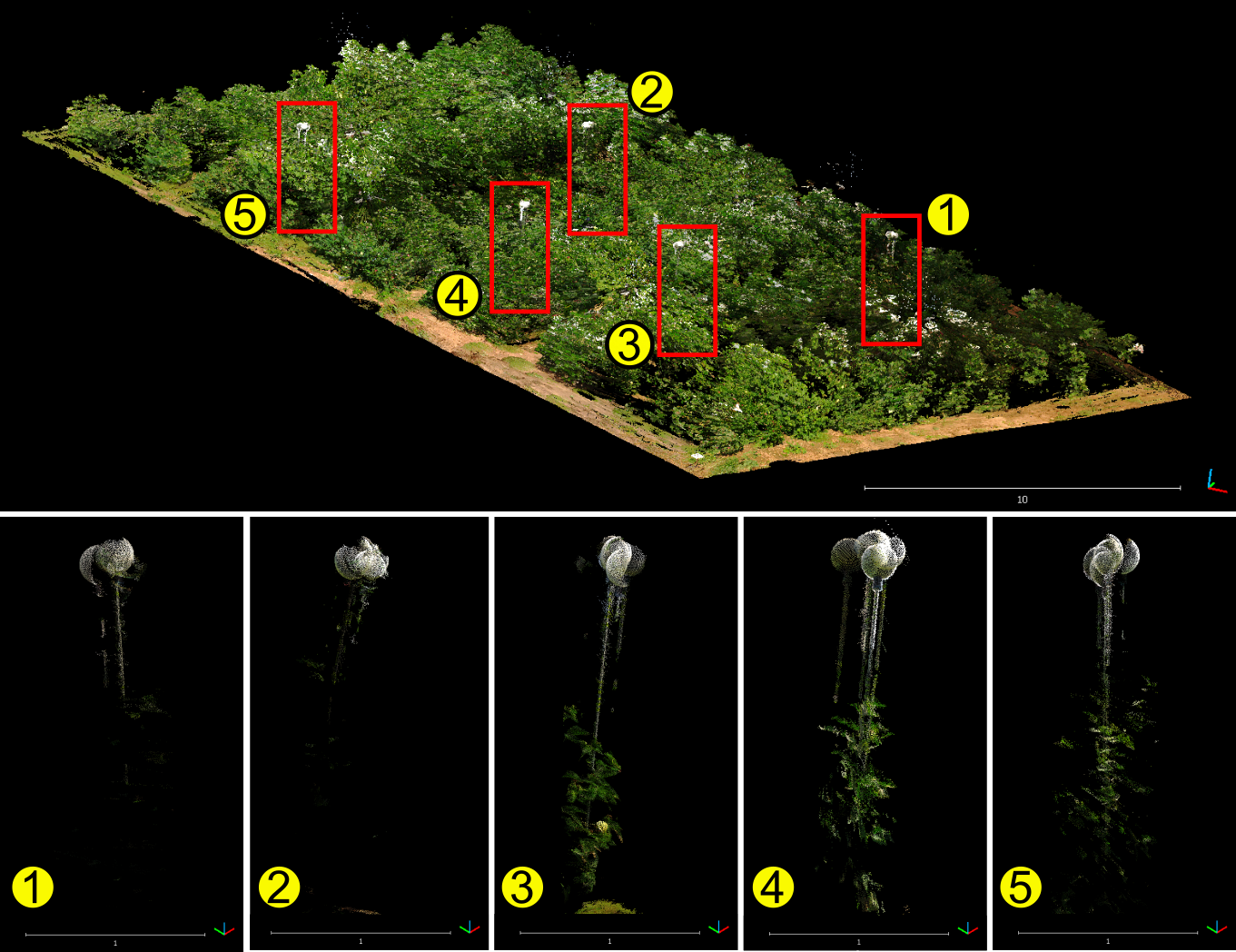}
    \caption{Pre-alignment results using only the information provided by the navigation sensors during the scanning process. The numbered insets show the five spheres deployed in the field for benchmarking}
    \label{fig:ENGR_preAlign}
\end{figure}

Initially, the point clouds were positioned (pre-aligned) based only the information gathered at each scan location during the scanning operation using SCENE software. However, as illustrated in the insets of Figure \ref{fig:ENGR_preAlign}, the point clouds did not align perfectly and displayed noticeable misalignments, with errors on the order of tens of centimeters. The maximum observed mismatch reached 27.3 cm. These discrepancies were primarily due to inaccuracies in pose estimation, specifically in the estimation of pitch and roll during the scan process.

The accurate registration of point clouds heavily relies on the orientation of the LiDAR scanner. To estimate pitch and roll, we utilized the information provided by the internal dual-axis compensator of the LiDAR scanner. Although the sensor's user manual \citep{Faro2020} specified an accuracy of 0.019$^{\circ}$ for inclinations between $\pm$2$^{\circ}$, we encountered challenges in controlling the inclination of the robotic platform during autonomous data collection, leading to higher inclinations in certain areas due to variations in terrain slope, bumps, or other surface irregularities.

\begin{figure} [H]
    \centering
    \includegraphics[width=\textwidth]{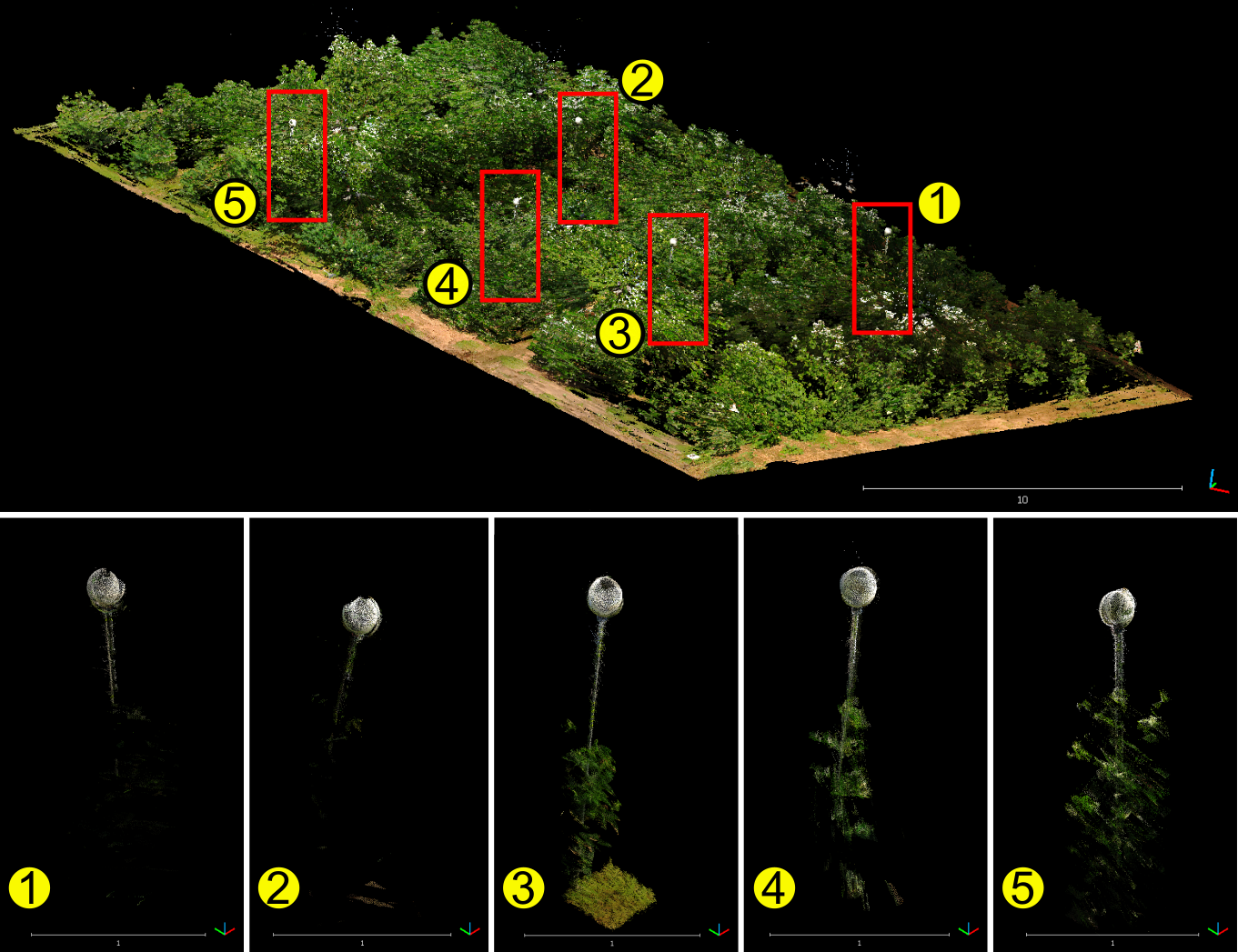}
    \caption{Automatic registration results using the navigation sensors information and automatic cloud-to-cloud registration. The numbered insets show the five spheres deployed in the field for benchmarking}
    \label{fig:ENGR_autoReg}
\end{figure}

Using the pre-aligned point clouds, an automatic C2C registration was performed using SCENE software. To ensure efficient and accurate point clouds matching, we limited the radius to search for common points to 30 cm, given the relatively small alignment errors. The results were highly promising, as demonstrated in Figure \ref{fig:ENGR_autoReg}, with a significant improvement in alignment. At the spheres level, the maximum misalignment observed was reduced to 4.17 cm.

Despite the challenges in controlling the inclination of the robotic platform during data collection, the alignment of the point clouds proved sufficiently accurate to enable automatic registration refinement using SCENE software. Table \ref{tab:registrationResults} provides a summary of the registration results obtained using SCENE software for both the target-based registration and our autonomous system (nav. sensors + C2C). 

\begin{table} [H]
\caption{Registration errors comparison for our method using the navigation sensors and cloud-to-cloud (C2C) registration versus target-based registration.} \label{tab:registrationResults}
\begin{center}

\begin{tabular} {|c||c|c|c|c|c|}
   \hline
   Registration method & \multicolumn{3}{c|}{Target statistics (cm)} & \multicolumn{2}{c|}{Scan point statistics (cm)} \\
   \hline
   & Mean error & Horizontal error & Vertical error & Max. point error & {Mean point error}\\
  \hline\hline
  Target-based & 0.59 & 0.53 & 0.22 & 2.10 & 1.37 \\
  \hline
  Nav. sensors + C2C & 1.72 & 1.34 & 0.82 & 3.02 & 2.07 \\
  \hline
\end{tabular}
\end{center}
\end{table}

Our autonomous point cloud registration method demonstrated high accuracy, with mean errors averaging approximately 2 cm. While target-based registration appears to outperform our autonomous system based on mean target error statistics, this metric can be misleading. It is important to note that target statistics are computed solely based on the points identified by the software as belonging to the spheres, calculating the distance between those points and the best-fitting sphere models. However, this approach may yield biased measurements, particularly when dealing with partial spheres containing only a small number of points. In such cases, the fitted model may not accurately represent the actual shape, leading to potentially misleading results.

A more comprehensive assessment of the registration performance can be obtained by considering scan point statistics. Our system consistently delivered measurements with mean errors close to 2 cm, representing a difference of only 50\% compared to the mean error obtained using spherical targets. This difference is less than 1 cm, indicating that our methodology can achieve results comparable to traditional and more laborious methods for TLS survey.

\begin{figure} [H]
    \centering
    \includegraphics[width=\textwidth]{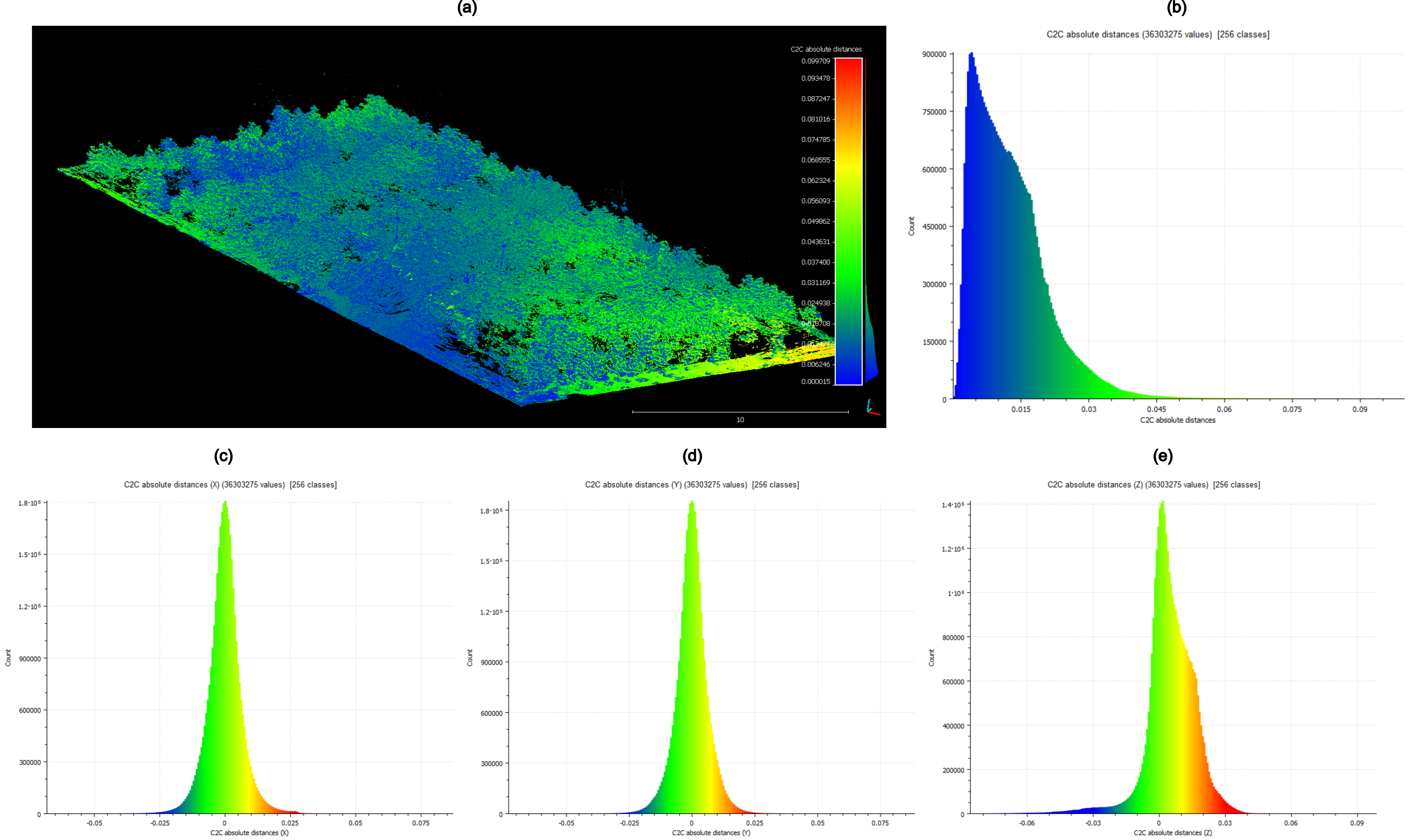}
    \caption{Dissimilarity results for the TLS data collected autonomously versus the target-based registered data. (a) Autonomously reconstructed point cloud colorized according to the Hausdorff distance. (b) Histogram of distances. Decomposition of distance for the X axis $dH_{X}$ (c), Y axis $dH_{Y}$ (d), and Z axis $dH_{Z}$ (e).}
    \label{fig:ENGR_autoReg_distances}
\end{figure}

Finally, we evaluated the dissimilarity between the point cloud registered using our approach and the target-based registered point cloud by computing the Hausdorff distance between them (Figure \ref{fig:ENGR_autoReg_distances}). The calculated distance for the autonomous TLS survey was 1.25 cm with a standard deviation of 0.82 cm. Examining the distribution of distances (Figure \ref{fig:ENGR_autoReg_distances} (b)), we observed that 85\% of the points had errors below 2 cm, and 95\% of the points had errors below 3 cm.

Analyzing the histograms for the X and Y axes (Figure \ref{fig:ENGR_autoReg_distances} (c) and (d)), we found that the distances followed a Gaussian distribution, with means of -0.014 cm and 0.001 cm, and standard deviations of 0.63 cm and 0.62 cm, respectively. However, for the Z axis (Figure \ref{fig:ENGR_autoReg_distances} (e) the distribution of distances exhibited a positive skewness, with a mean of 0.5 cm and a standard deviation of 1.08 cm. It is worth noting that the distances between point clouds along the Z axis were slightly larger compared to those along the X and Y axes. This can be attributed to the known lower accuracy in estimating altitude, typically double the accuracy of horizontal positioning. Nonetheless, the errors remained small and fell within the expected ranges of accuracy for standard RTK-GPS positioning systems.

Our approach provides a robust and efficient alternative to target-based registration, offering accuracy comparable to conventional approaches. While the target-based method may show a smaller mean error on paper, the practical difference is minimal, and our autonomous system demonstrates impressive accuracy in point cloud registration without the need for physical targets. However, considering the advantages of improved accuracy in scanner orientation estimation, it would be beneficial to incorporate dedicated sensors such as inertial measurement units to achieve more reliable results.

\section{LIMITATIONS AND FUTURE WORK}
The application of autonomous robotic systems in agricultural fields is an exciting area of research. However, several challenges arise due to the inherent complexities of real-life agricultural environments. One initial assumption we made was that all plots had the same length and the distance between them remained constant. However, real-life agricultural fields exhibit variability due to factors like unequal germination rates, planting errors, and heterogeneous plant growth. To address these challenges, we aim to investigate the use of digital twins to model the crop and the deployment of physics engines to simulate the interaction between laser beams and the plants more realistically. By doing so, we hope to enhance the system's ability to handle the complexities that arise throughout the agricultural season. This approach will allow for more precise visibility analysis, enabling the system to make informed decisions based on the current state of the field.

Despite our platform successfully navigated the field using the alleys of the field based only on pre-programmed waypoints and a 2D LiDAR to avoid immediate collisions, it is worth noting that certain paths may become obstructed as plants grow or lean, posing challenges for the navigation system as the season progresses. To address these challenges, we will integrate additional sensors like color and depth cameras for visual servoing. By leveraging real-time visual information, the platform can detect and avoid obstacles while traversing the field. Incorporating these sensor capabilities can significantly improve the safety and precision of the autonomous navigation system, enabling seamless adaptation to dynamic field conditions.

The LiDAR scanner on our robotic platform had limitations in capturing comprehensive data from the crop due to its relatively low height compared to the plants. To mitigate this, one solution could be using a taller structure, like a mast or pole, to mount the LiDAR. However, increasing the height must be carefully balanced with system stability, as placing a heavy object far from the robot's center of mass may introduce instability and pose a risk of tipping over. To address this concern, we are exploring novel approaches such as collapsible structures with auto-deployment capabilities, allowing the LiDAR to extend to a sufficient height when the robot is stable. This would optimize data collection while minimizing the risk of instability and system damage. Implementing this improvement will enhance data acquisition accuracy, facilitating better analysis and decision-making in agricultural applications.

Finally, the accurate registration results presented in our study offer the potential for estimating a wide range of morphological traits. To fully leverage the system's capabilities, we aim to explore its application in collecting 3D data over time for comprehensive 4D analysis. This approach would enable us to track trait evolution and deepen our understanding of crop development and its correlation with yield traits. By continuously monitoring and quantifying structural and morphological characteristics, our autonomous system could enable breeders and researchers to gain valuable insights into the dynamic evolution of the crop, improving yield predictions and understanding trait development. The collection and analysis of 4D data have the potential to revolutionize crop research, contributing to more efficient and sustainable agricultural practices.

\section{CONCLUSIONS}
In this study, we successfully introduced an innovative phenotyping platform and autonomous data collection methodology for in-field terrestrial laser scanning (TLS) for plant phenotyping. By integrating readily available equipment within the ROS framework, we achieved efficient and accurate LiDAR data acquisition from multiple field locations without human involvement. The capability to autonomously collect TLS data both inside and outside the field boundaries provides breeders with comprehensive and accurate plant 3D models, surpassing the limitations of conventional surveying methods. This represents a significant milestone in plant phenotyping, streamlining data collection and enhancing accuracy.

Our autonomous phenotyping platform has proven to be a reliable and effective alternative for TLS surveys, offering accuracy on par with traditional approaches while eliminating the need for cumbersome target setups. The precise localization through real-time kinematic positioning and sensor fusion enabled registration of point clouds with mean errors comparable to those using artificial targets. The seamless autonomous navigation further contributed to overall efficiency, showing the potential of our methodology in various LiDAR-based surveying applications for plant phenotyping.

The versatility of our phenotyping platform opens new possibilities for the scientific community, allowing the generalization of workflows and automation technologies to other crops and field layouts. This advancement offers breeders comprehensive and accurate crop models, enabling precise estimation of morphological traits at the plot level and beyond. The widespread adoption of our autonomous phenotyping platform can significantly advance plant phenomics and enhance breeding practices by enabling more efficient and accurate data-driven insights. It has the potential to revolutionize the way we approach agricultural research and crop improvement in the future.

\section{ACKNOWLEDGMENTS}
This work was funded by the National Science Foundation Growing Convergence Research (Award No. 1934481) and the Georgia Cotton Commission. The authors gratefully thank Dr. Rui Xu for his assistance at tuning the robot navigation. The authors also thank Mr. Gary Pierce, and Mr. Zhengkun Li for their help with the preparation of the field for the trials, and Mr. Daniel Petti for helping create the aerial field map.
ChatGPT assisted in sentence editing to improve clarity.


\bibliographystyle{apacite}
\bibliography{main_submission}

\begin{thebibliography}{}

\bibitem [\protect \citeauthoryear {%
Amanatides%
\ \BBA {} Woo%
}{%
Amanatides%
\ \BBA {} Woo%
}{%
{\protect \APACyear {1987}}%
}]{%
Amanatides1987}
\APACinsertmetastar {%
Amanatides1987}%
\begin{APACrefauthors}%
Amanatides, J.%
\BCBT {}\ \BBA {} Woo, A.%
\end{APACrefauthors}%
\unskip\
\newblock
\APACrefYearMonthDay{1987}{}{}.
\newblock
{\BBOQ}\APACrefatitle {A Fast Voxel Traversal Algorithm for Ray Tracing} {A
  fast voxel traversal algorithm for ray tracing}.{\BBCQ}
\newblock
\BIn{} \APACrefbtitle {EG 1987-{T}echnical {P}apers.} {Eg 1987-{T}echnical
  {P}apers.}
\newblock
\APACaddressPublisher{}{Eurographics Association}.
\newblock
\begin{APACrefDOI} \doi{https://doi.org/10.2312/egtp.19871000} \end{APACrefDOI}
\PrintBackRefs{\CurrentBib}

\bibitem [\protect \citeauthoryear {%
Appel%
}{%
Appel%
}{%
{\protect \APACyear {1968}}%
}]{%
Appel1968}
\APACinsertmetastar {%
Appel1968}%
\begin{APACrefauthors}%
Appel, A.%
\end{APACrefauthors}%
\unskip\
\newblock
\APACrefYearMonthDay{1968}{}{}.
\newblock
{\BBOQ}\APACrefatitle {Some Techniques for Shading Machine Renderings of
  Solids} {Some techniques for shading machine renderings of solids}.{\BBCQ}
\newblock
\BIn{} \APACrefbtitle {Proceedings of the April 30--May 2, 1968, Spring Joint
  Computer Conference} {Proceedings of the april 30--may 2, 1968, spring joint
  computer conference}\ (\BPG~37–45).
\newblock
\APACaddressPublisher{New York, NY, USA}{Association for Computing Machinery}.
\newblock
\begin{APACrefDOI} \doi{https://doi.org/10.1145/1468075.1468082}
  \end{APACrefDOI}
\PrintBackRefs{\CurrentBib}

\bibitem [\protect \citeauthoryear {%
Araus%
\ \BBA {} Cairns%
}{%
Araus%
\ \BBA {} Cairns%
}{%
{\protect \APACyear {2014}}%
}]{%
Araus2014}
\APACinsertmetastar {%
Araus2014}%
\begin{APACrefauthors}%
Araus, J\BPBI L.%
\BCBT {}\ \BBA {} Cairns, J\BPBI E.%
\end{APACrefauthors}%
\unskip\
\newblock
\APACrefYearMonthDay{2014}{}{}.
\newblock
{\BBOQ}\APACrefatitle {Field high-throughput phenotyping: The new crop breeding
  frontier} {Field high-throughput phenotyping: The new crop breeding
  frontier}.{\BBCQ}
\newblock
\APACjournalVolNumPages{Trends in Plant Science}{19}{1}{52-61}.
\newblock
\begin{APACrefDOI} \doi{https://doi.org/10.1016/j.tplants.2013.09.008}
  \end{APACrefDOI}
\PrintBackRefs{\CurrentBib}

\bibitem [\protect \citeauthoryear {%
Bai%
\ \protect \BOthers {.}}{%
Bai%
\ \protect \BOthers {.}}{%
{\protect \APACyear {2019}}%
}]{%
Bai2019}
\APACinsertmetastar {%
Bai2019}%
\begin{APACrefauthors}%
Bai, G.%
, Ge, Y.%
, Scoby, D.%
, Leavitt, B.%
, Stoerger, V.%
, Kirchgessner, N.%
\BDBL {}Awada, T.%
\end{APACrefauthors}%
\unskip\
\newblock
\APACrefYearMonthDay{2019}{}{}.
\newblock
{\BBOQ}\APACrefatitle {{NU}-{S}pidercam: A large-scale, cable-driven,
  integrated sensing and robotic system for advanced phenotyping, remote
  sensing, and agronomic research} {{NU}-{S}pidercam: A large-scale,
  cable-driven, integrated sensing and robotic system for advanced phenotyping,
  remote sensing, and agronomic research}.{\BBCQ}
\newblock
\APACjournalVolNumPages{Computers and Electronics in
  Agriculture}{160}{}{71-81}.
\newblock
\begin{APACrefDOI} \doi{https://doi.org/10.1016/j.compag.2019.03.009}
  \end{APACrefDOI}
\PrintBackRefs{\CurrentBib}

\bibitem [\protect \citeauthoryear {%
Beauch{\^e}ne%
\ \protect \BOthers {.}}{%
Beauch{\^e}ne%
\ \protect \BOthers {.}}{%
{\protect \APACyear {2019}}%
}]{%
Beauchne2019}
\APACinsertmetastar {%
Beauchne2019}%
\begin{APACrefauthors}%
Beauch{\^e}ne, K.%
, Leroy, F.%
, Fournier, A.%
, Huet, C.%
, Bonnefoy, M.%
, Lorgeou, J.%
\BDBL {}Cohan, J.%
\end{APACrefauthors}%
\unskip\
\newblock
\APACrefYearMonthDay{2019}{}{}.
\newblock
{\BBOQ}\APACrefatitle {Management and Characterization of Abiotic Stress via
  {P}h{\'e}no{F}ield{\textregistered}, a High-Throughput Field Phenotyping
  Platform} {Management and characterization of abiotic stress via
  {P}h{\'e}no{F}ield{\textregistered}, a high-throughput field phenotyping
  platform}.{\BBCQ}
\newblock
\APACjournalVolNumPages{Frontiers in Plant Science}{10}{}{}.
\newblock
\begin{APACrefDOI} \doi{https://doi.org/10.3389/fpls.2019.00904}
  \end{APACrefDOI}
\PrintBackRefs{\CurrentBib}

\bibitem [\protect \citeauthoryear {%
Bowman%
, Bourland%
, Myers%
, Wallace%
\BCBL {}\ \BBA {} Caldwell%
}{%
Bowman%
\ \protect \BOthers {.}}{%
{\protect \APACyear {2004}}%
}]{%
Bowman2004}
\APACinsertmetastar {%
Bowman2004}%
\begin{APACrefauthors}%
Bowman, D\BPBI T.%
, Bourland, F\BPBI M.%
, Myers, G\BPBI O.%
, Wallace, T\BPBI P.%
\BCBL {}\ \BBA {} Caldwell, D.%
\end{APACrefauthors}%
\unskip\
\newblock
\APACrefYearMonthDay{2004}{}{}.
\newblock
{\BBOQ}\APACrefatitle {Visual selection for yield in cotton breeding programs}
  {Visual selection for yield in cotton breeding programs}.{\BBCQ}
\newblock
\APACjournalVolNumPages{Journal of Cotton Science}{8}{2}{62-68}.
\PrintBackRefs{\CurrentBib}

\bibitem [\protect \citeauthoryear {%
Calders%
\ \protect \BOthers {.}}{%
Calders%
\ \protect \BOthers {.}}{%
{\protect \APACyear {2020}}%
}]{%
Calders2020}
\APACinsertmetastar {%
Calders2020}%
\begin{APACrefauthors}%
Calders, K.%
, Adams, J.%
, Armston, J.%
, Bartholomeus, H.%
, Bauwens, S.%
, Bentley, L\BPBI P.%
\BDBL {}Verbeeck, H.%
\end{APACrefauthors}%
\unskip\
\newblock
\APACrefYearMonthDay{2020}{}{}.
\newblock
{\BBOQ}\APACrefatitle {Terrestrial laser scanning in forest ecology: Expanding
  the horizon} {Terrestrial laser scanning in forest ecology: Expanding the
  horizon}.{\BBCQ}
\newblock
\APACjournalVolNumPages{Remote Sensing of Environment}{251}{}{112102}.
\newblock
\begin{APACrefDOI} \doi{https://doi.org/10.1016/j.rse.2020.112102}
  \end{APACrefDOI}
\PrintBackRefs{\CurrentBib}

\bibitem [\protect \citeauthoryear {%
Chvatal%
}{%
Chvatal%
}{%
{\protect \APACyear {1979}}%
}]{%
Chvatal1979}
\APACinsertmetastar {%
Chvatal1979}%
\begin{APACrefauthors}%
Chvatal, V.%
\end{APACrefauthors}%
\unskip\
\newblock
\APACrefYearMonthDay{1979}{aug}{}.
\newblock
{\BBOQ}\APACrefatitle {A Greedy Heuristic for the Set-Covering Problem} {A
  greedy heuristic for the set-covering problem}.{\BBCQ}
\newblock
\APACjournalVolNumPages{Mathematics of Operations Research}{4}{3}{233–235}.
\newblock
\begin{APACrefDOI} \doi{https://doi.org/10.1287/moor.4.3.233} \end{APACrefDOI}
\PrintBackRefs{\CurrentBib}

\bibitem [\protect \citeauthoryear {%
Coulter%
}{%
Coulter%
}{%
{\protect \APACyear {1992}}%
}]{%
Coulter1992}
\APACinsertmetastar {%
Coulter1992}%
\begin{APACrefauthors}%
Coulter, R\BPBI C.%
\end{APACrefauthors}%
\unskip\
\newblock
\APACrefYearMonthDay{1992}{}{}.
\newblock
\APACrefbtitle {Implementation of the pure pursuit path tracking algorithm}
  {Implementation of the pure pursuit path tracking algorithm}\
  \APACbVolEdTR{}{\BTR{}}.
\newblock
\APACaddressInstitution{Pittsburgh, PA, USA}{The robotic Institute,
  Carnegie-Mellon University}.
\PrintBackRefs{\CurrentBib}

\bibitem [\protect \citeauthoryear {%
Crommelinck%
\ \BBA {} Höfle%
}{%
Crommelinck%
\ \BBA {} Höfle%
}{%
{\protect \APACyear {2016}}%
}]{%
Crommelinck2016}
\APACinsertmetastar {%
Crommelinck2016}%
\begin{APACrefauthors}%
Crommelinck, S.%
\BCBT {}\ \BBA {} Höfle, B.%
\end{APACrefauthors}%
\unskip\
\newblock
\APACrefYearMonthDay{2016}{}{}.
\newblock
{\BBOQ}\APACrefatitle {Simulating an Autonomously Operating Low-Cost Static
  Terrestrial {LiDAR} for Multitemporal Maize Crop Height Measurements}
  {Simulating an autonomously operating low-cost static terrestrial {LiDAR} for
  multitemporal maize crop height measurements}.{\BBCQ}
\newblock
\APACjournalVolNumPages{Remote Sensing}{8}{3}{}.
\newblock
\begin{APACrefDOI} \doi{https://doi.org/10.3390/rs8030205} \end{APACrefDOI}
\PrintBackRefs{\CurrentBib}

\bibitem [\protect \citeauthoryear {%
Dantzig%
, Fulkerson%
\BCBL {}\ \BBA {} Johnson%
}{%
Dantzig%
\ \protect \BOthers {.}}{%
{\protect \APACyear {1954}}%
}]{%
Dantzig1954}
\APACinsertmetastar {%
Dantzig1954}%
\begin{APACrefauthors}%
Dantzig, G.%
, Fulkerson, R.%
\BCBL {}\ \BBA {} Johnson, S.%
\end{APACrefauthors}%
\unskip\
\newblock
\APACrefYearMonthDay{1954}{}{}.
\newblock
{\BBOQ}\APACrefatitle {Solution of a large-scale traveling-salesman problem}
  {Solution of a large-scale traveling-salesman problem}.{\BBCQ}
\newblock
\APACjournalVolNumPages{Journal of the Operations Research Society of
  America}{2}{4}{393--410}.
\newblock
\begin{APACrefDOI} \doi{https://doi.org/10.1287/opre.2.4.393} \end{APACrefDOI}
\PrintBackRefs{\CurrentBib}

\bibitem [\protect \citeauthoryear {%
Deery%
, Jimenez-Berni%
, Jones%
, Sirault%
\BCBL {}\ \BBA {} Furbank%
}{%
Deery%
\ \protect \BOthers {.}}{%
{\protect \APACyear {2014}}%
}]{%
Deery2014}
\APACinsertmetastar {%
Deery2014}%
\begin{APACrefauthors}%
Deery, D.%
, Jimenez-Berni, J.%
, Jones, H.%
, Sirault, X.%
\BCBL {}\ \BBA {} Furbank, R.%
\end{APACrefauthors}%
\unskip\
\newblock
\APACrefYearMonthDay{2014}{}{}.
\newblock
{\BBOQ}\APACrefatitle {Proximal remote sensing buggies and potential
  applications for field-based phenotyping} {Proximal remote sensing buggies
  and potential applications for field-based phenotyping}.{\BBCQ}
\newblock
\APACjournalVolNumPages{Agronomy}{4}{3}{349-379}.
\newblock
\begin{APACrefDOI} \doi{https://doi.org/10.3390/AGRONOMY4030349}
  \end{APACrefDOI}
\PrintBackRefs{\CurrentBib}

\bibitem [\protect \citeauthoryear {%
Ehlert%
, Heisig%
\BCBL {}\ \BBA {} Adamek%
}{%
Ehlert%
\ \protect \BOthers {.}}{%
{\protect \APACyear {2010}}%
}]{%
Ehlert2010}
\APACinsertmetastar {%
Ehlert2010}%
\begin{APACrefauthors}%
Ehlert, D.%
, Heisig, M.%
\BCBL {}\ \BBA {} Adamek, R.%
\end{APACrefauthors}%
\unskip\
\newblock
\APACrefYearMonthDay{2010}{}{}.
\newblock
{\BBOQ}\APACrefatitle {Suitability of a laser rangefinder to characterize
  winter wheat} {Suitability of a laser rangefinder to characterize winter
  wheat}.{\BBCQ}
\newblock
\APACjournalVolNumPages{Precision Agriculture}{11}{}{650-663}.
\newblock
\begin{APACrefDOI} \doi{https://doi.org/10.1007/s11119-010-9191-4}
  \end{APACrefDOI}
\PrintBackRefs{\CurrentBib}

\bibitem [\protect \citeauthoryear {%
Eitel%
, Magney%
, Vierling%
, Brown%
\BCBL {}\ \BBA {} Huggins%
}{%
Eitel%
\ \protect \BOthers {.}}{%
{\protect \APACyear {2014}}%
}]{%
Eitel2014}
\APACinsertmetastar {%
Eitel2014}%
\begin{APACrefauthors}%
Eitel, J\BPBI U.%
, Magney, T\BPBI S.%
, Vierling, L\BPBI A.%
, Brown, T\BPBI T.%
\BCBL {}\ \BBA {} Huggins, D\BPBI R.%
\end{APACrefauthors}%
\unskip\
\newblock
\APACrefYearMonthDay{2014}{}{}.
\newblock
{\BBOQ}\APACrefatitle {{LiDAR} based biomass and crop nitrogen estimates for
  rapid, non-destructive assessment of wheat nitrogen status} {{LiDAR} based
  biomass and crop nitrogen estimates for rapid, non-destructive assessment of
  wheat nitrogen status}.{\BBCQ}
\newblock
\APACjournalVolNumPages{Field Crops Research}{159}{}{21-32}.
\newblock
\begin{APACrefDOI} \doi{https://doi.org/10.1016/j.fcr.2014.01.008}
  \end{APACrefDOI}
\PrintBackRefs{\CurrentBib}

\bibitem [\protect \citeauthoryear {%
Eitel%
, Magney%
, Vierling%
, Greaves%
\BCBL {}\ \BBA {} Zheng%
}{%
Eitel%
\ \protect \BOthers {.}}{%
{\protect \APACyear {2016}}%
}]{%
Eitel2016}
\APACinsertmetastar {%
Eitel2016}%
\begin{APACrefauthors}%
Eitel, J\BPBI U.%
, Magney, T\BPBI S.%
, Vierling, L\BPBI A.%
, Greaves, H\BPBI E.%
\BCBL {}\ \BBA {} Zheng, G.%
\end{APACrefauthors}%
\unskip\
\newblock
\APACrefYearMonthDay{2016}{}{}.
\newblock
{\BBOQ}\APACrefatitle {An automated method to quantify crop height and
  calibrate satellite-derived biomass using hypertemporal lidar} {An automated
  method to quantify crop height and calibrate satellite-derived biomass using
  hypertemporal lidar}.{\BBCQ}
\newblock
\APACjournalVolNumPages{Remote Sensing of Environment}{187}{}{414-422}.
\newblock
\begin{APACrefDOI} \doi{https://doi.org/10.1016/j.rse.2016.10.044}
  \end{APACrefDOI}
\PrintBackRefs{\CurrentBib}

\bibitem [\protect \citeauthoryear {%
El-Naggar%
\ \protect \BOthers {.}}{%
El-Naggar%
\ \protect \BOthers {.}}{%
{\protect \APACyear {2021}}%
}]{%
Elnaggar2021}
\APACinsertmetastar {%
Elnaggar2021}%
\begin{APACrefauthors}%
El-Naggar, A.%
, Jolly, B.%
, Hedley, C.%
, Horne, D.%
, Roudier, P.%
\BCBL {}\ \BBA {} Clothier, B.%
\end{APACrefauthors}%
\unskip\
\newblock
\APACrefYearMonthDay{2021}{}{}.
\newblock
{\BBOQ}\APACrefatitle {The use of terrestrial LiDAR to monitor crop growth and
  account for within-field variability of crop coefficients and water use} {The
  use of terrestrial lidar to monitor crop growth and account for within-field
  variability of crop coefficients and water use}.{\BBCQ}
\newblock
\APACjournalVolNumPages{Computers and Electronics in
  Agriculture}{190}{}{106416}.
\newblock
\begin{APACrefDOI} \doi{https://doi.org/10.1016/j.compag.2021.106416}
  \end{APACrefDOI}
\PrintBackRefs{\CurrentBib}

\bibitem [\protect \citeauthoryear {%
Fang%
\ \protect \BOthers {.}}{%
Fang%
\ \protect \BOthers {.}}{%
{\protect \APACyear {2020}}%
}]{%
Fang2020}
\APACinsertmetastar {%
Fang2020}%
\begin{APACrefauthors}%
Fang, Y.%
, Qiu, X.%
, Guo, T.%
, Wang, Y.%
, Cheng, T.%
, Zhu, Y.%
\BDBL {}Gui, L.%
\end{APACrefauthors}%
\unskip\
\newblock
\APACrefYearMonthDay{2020}{}{}.
\newblock
{\BBOQ}\APACrefatitle {An automatic method for counting wheat tiller number in
  the field with terrestrial {LiDAR}} {An automatic method for counting wheat
  tiller number in the field with terrestrial {LiDAR}}.{\BBCQ}
\newblock
\APACjournalVolNumPages{Plant Methods}{16}{}{}.
\newblock
\begin{APACrefDOI} \doi{https://doi.org/10.1186/s13007-020-00672-8}
  \end{APACrefDOI}
\PrintBackRefs{\CurrentBib}

\bibitem [\protect \citeauthoryear {%
{FARO Technologies}%
}{%
{FARO Technologies}%
}{%
{\protect \APACyear {2020}}%
}]{%
Faro2020}
\APACinsertmetastar {%
Faro2020}%
\begin{APACrefauthors}%
{FARO Technologies}.%
\end{APACrefauthors}%
\unskip\
\newblock
\APACrefYearMonthDay{2020}{1}{}.
\newblock
{\BBOQ}\APACrefatitle {Automation interface user manual} {Automation interface
  user manual}{\BBCQ}\ [\bibcomputersoftwaremanual].
\PrintBackRefs{\CurrentBib}

\bibitem [\protect \citeauthoryear {%
Friedli%
\ \protect \BOthers {.}}{%
Friedli%
\ \protect \BOthers {.}}{%
{\protect \APACyear {2016}}%
}]{%
Friedli2016}
\APACinsertmetastar {%
Friedli2016}%
\begin{APACrefauthors}%
Friedli, M.%
, Kirchgessner, N.%
, Grieder, C.%
, Liebisch, F.%
, Mannale, M.%
\BCBL {}\ \BBA {} Walter, A.%
\end{APACrefauthors}%
\unskip\
\newblock
\APACrefYearMonthDay{2016}{}{}.
\newblock
{\BBOQ}\APACrefatitle {Terrestrial 3D laser scanning to track the increase in
  canopy height of both monocot and dicot crop species under field conditions}
  {Terrestrial 3d laser scanning to track the increase in canopy height of both
  monocot and dicot crop species under field conditions}.{\BBCQ}
\newblock
\APACjournalVolNumPages{Plant Methods}{12}{1}{1--15}.
\newblock
\begin{APACrefDOI} \doi{https://doi.org/10.1186/s13007-016-0109-7}
  \end{APACrefDOI}
\PrintBackRefs{\CurrentBib}

\bibitem [\protect \citeauthoryear {%
Gage%
\ \protect \BOthers {.}}{%
Gage%
\ \protect \BOthers {.}}{%
{\protect \APACyear {2019}}%
}]{%
Gage2019}
\APACinsertmetastar {%
Gage2019}%
\begin{APACrefauthors}%
Gage, J\BPBI L.%
, Richards, E.%
, Lepak, N.%
, Kaczmar, N.%
, Soman, C.%
, Chowdhary, G.%
\BDBL {}Buckler, E\BPBI S.%
\end{APACrefauthors}%
\unskip\
\newblock
\APACrefYearMonthDay{2019}{}{}.
\newblock
{\BBOQ}\APACrefatitle {In‐Field Whole‐Plant Maize Architecture
  Characterized by Subcanopy Rovers and Latent Space Phenotyping} {In‐field
  whole‐plant maize architecture characterized by subcanopy rovers and latent
  space phenotyping}.{\BBCQ}
\newblock
\APACjournalVolNumPages{The Plant Phenome Journal}{2}{}{}.
\newblock
\begin{APACrefDOI} \doi{https://doi.org/10.2135/tppj2019.07.0011}
  \end{APACrefDOI}
\PrintBackRefs{\CurrentBib}

\bibitem [\protect \citeauthoryear {%
Goggin%
, Lorence%
\BCBL {}\ \BBA {} Topp%
}{%
Goggin%
\ \protect \BOthers {.}}{%
{\protect \APACyear {2015}}%
}]{%
Goggin2015}
\APACinsertmetastar {%
Goggin2015}%
\begin{APACrefauthors}%
Goggin, F\BPBI L.%
, Lorence, A.%
\BCBL {}\ \BBA {} Topp, C\BPBI N.%
\end{APACrefauthors}%
\unskip\
\newblock
\APACrefYearMonthDay{2015}{}{}.
\newblock
{\BBOQ}\APACrefatitle {Applying high-throughput phenotyping to plant–insect
  interactions: picturing more resistant crops} {Applying high-throughput
  phenotyping to plant–insect interactions: picturing more resistant
  crops}.{\BBCQ}
\newblock
\APACjournalVolNumPages{Current Opinion in Insect Science}{9}{}{69-76}.
\newblock
\begin{APACrefDOI} \doi{https://doi.org/10.1016/J.COIS.2015.03.002}
  \end{APACrefDOI}
\PrintBackRefs{\CurrentBib}

\bibitem [\protect \citeauthoryear {%
Gottschalk%
, Lin%
\BCBL {}\ \BBA {} Manocha%
}{%
Gottschalk%
\ \protect \BOthers {.}}{%
{\protect \APACyear {1996}}%
}]{%
Gottschalk1996}
\APACinsertmetastar {%
Gottschalk1996}%
\begin{APACrefauthors}%
Gottschalk, S.%
, Lin, M\BPBI C.%
\BCBL {}\ \BBA {} Manocha, D.%
\end{APACrefauthors}%
\unskip\
\newblock
\APACrefYearMonthDay{1996}{}{}.
\newblock
{\BBOQ}\APACrefatitle {{OBBTree}: A Hierarchical Structure for Rapid
  Interference Detection} {{OBBTree}: A hierarchical structure for rapid
  interference detection}.{\BBCQ}
\newblock
\BIn{} \APACrefbtitle {Proceedings of the 23rd Annual Conference on Computer
  Graphics and Interactive Techniques} {Proceedings of the 23rd annual
  conference on computer graphics and interactive techniques}\
  (\BPG~171–180).
\newblock
\APACaddressPublisher{New York, NY, USA}{Association for Computing Machinery}.
\newblock
\begin{APACrefDOI} \doi{https://doi.org/10.1145/237170.237244} \end{APACrefDOI}
\PrintBackRefs{\CurrentBib}

\bibitem [\protect \citeauthoryear {%
Großkinsky%
, Svensgaard%
, Christensen%
\BCBL {}\ \BBA {} Roitsch%
}{%
Großkinsky%
\ \protect \BOthers {.}}{%
{\protect \APACyear {2015}}%
}]{%
Grosskinsky2015}
\APACinsertmetastar {%
Grosskinsky2015}%
\begin{APACrefauthors}%
Großkinsky, D\BPBI K.%
, Svensgaard, J.%
, Christensen, S.%
\BCBL {}\ \BBA {} Roitsch, T.%
\end{APACrefauthors}%
\unskip\
\newblock
\APACrefYearMonthDay{2015}{}{}.
\newblock
{\BBOQ}\APACrefatitle {Plant phenomics and the need for physiological
  phenotyping across scales to narrow the genotype-to-phenotype knowledge gap}
  {Plant phenomics and the need for physiological phenotyping across scales to
  narrow the genotype-to-phenotype knowledge gap}.{\BBCQ}
\newblock
\APACjournalVolNumPages{Journal of experimental botany}{66}{18}{5429-5440}.
\newblock
\begin{APACrefDOI} \doi{https://doi.org/10.1093/jxb/erv345} \end{APACrefDOI}
\PrintBackRefs{\CurrentBib}

\bibitem [\protect \citeauthoryear {%
Q.~Guo%
\ \protect \BOthers {.}}{%
Q.~Guo%
\ \protect \BOthers {.}}{%
{\protect \APACyear {2018}}%
}]{%
Guo2018}
\APACinsertmetastar {%
Guo2018}%
\begin{APACrefauthors}%
Guo, Q.%
, Wu, F.%
, Pang, S.%
, Zhao, X.%
, Chen, L.%
, Liu, J.%
\BDBL {}Chu, C.%
\end{APACrefauthors}%
\unskip\
\newblock
\APACrefYearMonthDay{2018}{March}{}.
\newblock
{\BBOQ}\APACrefatitle {Crop 3{D}-a {LiDAR} based platform for 3{D}
  high-throughput crop phenotyping} {Crop 3{D}-a {LiDAR} based platform for
  3{D} high-throughput crop phenotyping}.{\BBCQ}
\newblock
\APACjournalVolNumPages{Science China. Life sciences}{61}{}{328—339}.
\newblock
\begin{APACrefDOI} \doi{https://doi.org/10.1007/s11427-017-9056-0}
  \end{APACrefDOI}
\PrintBackRefs{\CurrentBib}

\bibitem [\protect \citeauthoryear {%
T.~Guo%
\ \protect \BOthers {.}}{%
T.~Guo%
\ \protect \BOthers {.}}{%
{\protect \APACyear {2019}}%
}]{%
Guo2019}
\APACinsertmetastar {%
Guo2019}%
\begin{APACrefauthors}%
Guo, T.%
, Fang, Y.%
, Cheng, T.%
, Tian, Y.%
, Zhu, Y.%
, Chen, Q.%
\BDBL {}Yao, X.%
\end{APACrefauthors}%
\unskip\
\newblock
\APACrefYearMonthDay{2019}{}{}.
\newblock
{\BBOQ}\APACrefatitle {Detection of wheat height using optimized multi-scan
  mode of {LiDAR} during the entire growth stages} {Detection of wheat height
  using optimized multi-scan mode of {LiDAR} during the entire growth
  stages}.{\BBCQ}
\newblock
\APACjournalVolNumPages{Computers and Electronics in
  Agriculture}{165}{}{104959}.
\newblock
\begin{APACrefDOI} \doi{https://doi.org/10.1016/j.compag.2019.104959}
  \end{APACrefDOI}
\PrintBackRefs{\CurrentBib}

\bibitem [\protect \citeauthoryear {%
Hoffmeister%
\ \protect \BOthers {.}}{%
Hoffmeister%
\ \protect \BOthers {.}}{%
{\protect \APACyear {2013}}%
}]{%
Hoffmeister2013}
\APACinsertmetastar {%
Hoffmeister2013}%
\begin{APACrefauthors}%
Hoffmeister, D.%
, Waldhoff, G.%
, Curdt, C.%
, Tilly, N.%
, Bendig, J.%
\BCBL {}\ \BBA {} Bareth, G.%
\end{APACrefauthors}%
\unskip\
\newblock
\APACrefYearMonthDay{2013}{}{}.
\newblock
{\BBOQ}\APACrefatitle {Spatial variability detection of crop height in a single
  field by terrestrial laser scanning} {Spatial variability detection of crop
  height in a single field by terrestrial laser scanning}.{\BBCQ}
\newblock
\BIn{} J\BPBI V.~Stafford\ (\BED), \APACrefbtitle {Precision agriculture '13}
  {Precision agriculture '13}\ (\BPGS\ 267--274).
\newblock
\APACaddressPublisher{Wageningen}{Wageningen Academic Publishers}.
\newblock
\begin{APACrefDOI} \doi{https://doi.org/10.3920/978-90-8686-778-3}
  \end{APACrefDOI}
\PrintBackRefs{\CurrentBib}

\bibitem [\protect \citeauthoryear {%
Hosoi%
\ \BBA {} Omasa%
}{%
Hosoi%
\ \BBA {} Omasa%
}{%
{\protect \APACyear {2009}}%
}]{%
Hosoi2009}
\APACinsertmetastar {%
Hosoi2009}%
\begin{APACrefauthors}%
Hosoi, F.%
\BCBT {}\ \BBA {} Omasa, K.%
\end{APACrefauthors}%
\unskip\
\newblock
\APACrefYearMonthDay{2009}{}{}.
\newblock
{\BBOQ}\APACrefatitle {Estimating vertical plant area density profile and
  growth parameters of a wheat canopy at different growth stages using
  three-dimensional portable lidar imaging} {Estimating vertical plant area
  density profile and growth parameters of a wheat canopy at different growth
  stages using three-dimensional portable lidar imaging}.{\BBCQ}
\newblock
\APACjournalVolNumPages{ISPRS Journal of Photogrammetry and Remote
  Sensing}{64}{2}{151-158}.
\newblock
\begin{APACrefDOI} \doi{https://doi.org/10.1016/j.isprsjprs.2008.09.003}
  \end{APACrefDOI}
\PrintBackRefs{\CurrentBib}

\bibitem [\protect \citeauthoryear {%
Hosoi%
\ \BBA {} Omasa%
}{%
Hosoi%
\ \BBA {} Omasa%
}{%
{\protect \APACyear {2012}}%
}]{%
Hosoi2012}
\APACinsertmetastar {%
Hosoi2012}%
\begin{APACrefauthors}%
Hosoi, F.%
\BCBT {}\ \BBA {} Omasa, K.%
\end{APACrefauthors}%
\unskip\
\newblock
\APACrefYearMonthDay{2012}{}{}.
\newblock
{\BBOQ}\APACrefatitle {Estimation of vertical plant area density profiles in a
  rice canopy at different growth stages by high-resolution portable scanning
  lidar with a lightweight mirror} {Estimation of vertical plant area density
  profiles in a rice canopy at different growth stages by high-resolution
  portable scanning lidar with a lightweight mirror}.{\BBCQ}
\newblock
\APACjournalVolNumPages{ISPRS Journal of Photogrammetry and Remote
  Sensing}{74}{}{11-19}.
\newblock
\begin{APACrefDOI} \doi{https://doi.org/10.1016/j.isprsjprs.2012.08.001}
  \end{APACrefDOI}
\PrintBackRefs{\CurrentBib}

\bibitem [\protect \citeauthoryear {%
Hämmerle%
\ \BBA {} Höfle%
}{%
Hämmerle%
\ \BBA {} Höfle%
}{%
{\protect \APACyear {2014}}%
}]{%
Hammerle2014}
\APACinsertmetastar {%
Hammerle2014}%
\begin{APACrefauthors}%
Hämmerle, M.%
\BCBT {}\ \BBA {} Höfle, B.%
\end{APACrefauthors}%
\unskip\
\newblock
\APACrefYearMonthDay{2014}{}{}.
\newblock
{\BBOQ}\APACrefatitle {Effects of Reduced Terrestrial {LiDAR} Point Density on
  High-Resolution Grain Crop Surface Models in Precision Agriculture} {Effects
  of reduced terrestrial {LiDAR} point density on high-resolution grain crop
  surface models in precision agriculture}.{\BBCQ}
\newblock
\APACjournalVolNumPages{Sensors}{14}{12}{24212--24230}.
\newblock
\begin{APACrefDOI} \doi{https://doi.org/10.3390/s141224212} \end{APACrefDOI}
\PrintBackRefs{\CurrentBib}

\bibitem [\protect \citeauthoryear {%
Jiang%
\ \protect \BOthers {.}}{%
Jiang%
\ \protect \BOthers {.}}{%
{\protect \APACyear {2019}}%
}]{%
Jiang2019}
\APACinsertmetastar {%
Jiang2019}%
\begin{APACrefauthors}%
Jiang, Y.%
, Li, C.%
, Takeda, F.%
, Kramer, E\BPBI A.%
, Ashrafi, H.%
\BCBL {}\ \BBA {} Hunter, J.%
\end{APACrefauthors}%
\unskip\
\newblock
\APACrefYearMonthDay{2019}{04}{}.
\newblock
{\BBOQ}\APACrefatitle {{3D point cloud data to quantitatively characterize size
  and shape of shrub crops}} {{3D point cloud data to quantitatively
  characterize size and shape of shrub crops}}.{\BBCQ}
\newblock
\APACjournalVolNumPages{Horticulture Research}{6}{}{}.
\newblock
\APACrefnote{43}
\newblock
\begin{APACrefDOI} \doi{https://doi.org/10.1038/s41438-019-0123-9}
  \end{APACrefDOI}
\PrintBackRefs{\CurrentBib}

\bibitem [\protect \citeauthoryear {%
Jim{\'e}nez-Berni%
\ \protect \BOthers {.}}{%
Jim{\'e}nez-Berni%
\ \protect \BOthers {.}}{%
{\protect \APACyear {2018}}%
}]{%
JimenezBerni2018}
\APACinsertmetastar {%
JimenezBerni2018}%
\begin{APACrefauthors}%
Jim{\'e}nez-Berni, J\BPBI A.%
, Deery, D\BPBI M.%
, Rozas-Larraondo, P.%
, Condon, A\BPBI G.%
, Rebetzke, G\BPBI J.%
, James, R.%
\BDBL {}Sirault, X.%
\end{APACrefauthors}%
\unskip\
\newblock
\APACrefYearMonthDay{2018}{}{}.
\newblock
{\BBOQ}\APACrefatitle {High Throughput Determination of Plant Height, Ground
  Cover, and Above-Ground Biomass in Wheat with {LiDAR}} {High throughput
  determination of plant height, ground cover, and above-ground biomass in
  wheat with {LiDAR}}.{\BBCQ}
\newblock
\APACjournalVolNumPages{Frontiers in Plant Science}{9}{}{}.
\newblock
\begin{APACrefDOI} \doi{https://doi.org/10.3389/fpls.2018.00237}
  \end{APACrefDOI}
\PrintBackRefs{\CurrentBib}

\bibitem [\protect \citeauthoryear {%
Jin%
\ \protect \BOthers {.}}{%
Jin%
\ \protect \BOthers {.}}{%
{\protect \APACyear {2018}}%
}]{%
Jin2018Deep}
\APACinsertmetastar {%
Jin2018Deep}%
\begin{APACrefauthors}%
Jin, S.%
, Su, Y.%
, Gao, S.%
, Wu, F.%
, Hu, T.%
, Liu, J.%
\BDBL {}Guo, Q.%
\end{APACrefauthors}%
\unskip\
\newblock
\APACrefYearMonthDay{2018}{}{}.
\newblock
{\BBOQ}\APACrefatitle {Deep Learning: Individual Maize Segmentation From
  Terrestrial Lidar Data Using Faster {R-CNN} and Regional Growth Algorithms}
  {Deep learning: Individual maize segmentation from terrestrial lidar data
  using faster {R-CNN} and regional growth algorithms}.{\BBCQ}
\newblock
\APACjournalVolNumPages{Frontiers in Plant Science}{9}{}{}.
\newblock
\begin{APACrefDOI} \doi{https://doi.org/10.3389/fpls.2018.00866}
  \end{APACrefDOI}
\PrintBackRefs{\CurrentBib}

\bibitem [\protect \citeauthoryear {%
Jin%
\ \protect \BOthers {.}}{%
Jin%
\ \protect \BOthers {.}}{%
{\protect \APACyear {2020}}%
}]{%
Jin2020}
\APACinsertmetastar {%
Jin2020}%
\begin{APACrefauthors}%
Jin, S.%
, Su, Y.%
, Song, S.%
, Xu, K.%
, Hu, T.%
, Yang, Q.%
\BDBL {}others%
\end{APACrefauthors}%
\unskip\
\newblock
\APACrefYearMonthDay{2020}{}{}.
\newblock
{\BBOQ}\APACrefatitle {Non-destructive estimation of field maize biomass using
  terrestrial lidar: {A}n evaluation from plot level to individual leaf level}
  {Non-destructive estimation of field maize biomass using terrestrial lidar:
  {A}n evaluation from plot level to individual leaf level}.{\BBCQ}
\newblock
\APACjournalVolNumPages{Plant Methods}{16}{1}{1--19}.
\newblock
\begin{APACrefDOI} \doi{https://doi.org/10.1186/s13007-020-00613-5}
  \end{APACrefDOI}
\PrintBackRefs{\CurrentBib}

\bibitem [\protect \citeauthoryear {%
Jin%
\ \protect \BOthers {.}}{%
Jin%
\ \protect \BOthers {.}}{%
{\protect \APACyear {2019}}%
}]{%
Jin2019Stem}
\APACinsertmetastar {%
Jin2019Stem}%
\begin{APACrefauthors}%
Jin, S.%
, Su, Y.%
, Wu, F.%
, Pang, S.%
, Gao, S.%
, Hu, T.%
\BDBL {}Guo, Q.%
\end{APACrefauthors}%
\unskip\
\newblock
\APACrefYearMonthDay{2019}{}{}.
\newblock
{\BBOQ}\APACrefatitle {Stem–Leaf Segmentation and Phenotypic Trait Extraction
  of Individual Maize Using Terrestrial {LiDAR} Data} {Stem–leaf segmentation
  and phenotypic trait extraction of individual maize using terrestrial {LiDAR}
  data}.{\BBCQ}
\newblock
\APACjournalVolNumPages{IEEE Transactions on Geoscience and Remote
  Sensing}{57}{3}{1336-1346}.
\newblock
\begin{APACrefDOI} \doi{https://doi.org/10.1109/TGRS.2018.2866056}
  \end{APACrefDOI}
\PrintBackRefs{\CurrentBib}

\bibitem [\protect \citeauthoryear {%
Jin%
\ \protect \BOthers {.}}{%
Jin%
\ \protect \BOthers {.}}{%
{\protect \APACyear {2021}}%
}]{%
Jin2021Lidar}
\APACinsertmetastar {%
Jin2021Lidar}%
\begin{APACrefauthors}%
Jin, S.%
, Sun, X.%
, Wu, F.%
, Su, Y.%
, Li, Y.%
, Song, S.%
\BDBL {}Guo, Q.%
\end{APACrefauthors}%
\unskip\
\newblock
\APACrefYearMonthDay{2021}{}{}.
\newblock
{\BBOQ}\APACrefatitle {Lidar sheds new light on plant phenomics for plant
  breeding and management: {R}ecent advances and future prospects} {Lidar sheds
  new light on plant phenomics for plant breeding and management: {R}ecent
  advances and future prospects}.{\BBCQ}
\newblock
\APACjournalVolNumPages{ISPRS Journal of Photogrammetry and Remote
  Sensing}{171}{}{202-223}.
\newblock
\begin{APACrefDOI} \doi{https://doi.org/10.1016/j.isprsjprs.2020.11.006}
  \end{APACrefDOI}
\PrintBackRefs{\CurrentBib}

\bibitem [\protect \citeauthoryear {%
Kay%
\ \BBA {} Kajiya%
}{%
Kay%
\ \BBA {} Kajiya%
}{%
{\protect \APACyear {1986}}%
}]{%
Kay1986}
\APACinsertmetastar {%
Kay1986}%
\begin{APACrefauthors}%
Kay, T\BPBI L.%
\BCBT {}\ \BBA {} Kajiya, J\BPBI T.%
\end{APACrefauthors}%
\unskip\
\newblock
\APACrefYearMonthDay{1986}{}{}.
\newblock
{\BBOQ}\APACrefatitle {Ray Tracing Complex Scenes} {Ray tracing complex
  scenes}.{\BBCQ}
\newblock
\BIn{} \APACrefbtitle {Proceedings of the 13th Annual Conference on Computer
  Graphics and Interactive Techniques} {Proceedings of the 13th annual
  conference on computer graphics and interactive techniques}\
  (\BPG~269–278).
\newblock
\APACaddressPublisher{New York, NY, USA}{Association for Computing Machinery}.
\newblock
\begin{APACrefDOI} \doi{https://doi.org/10.1145/15922.15916} \end{APACrefDOI}
\PrintBackRefs{\CurrentBib}

\bibitem [\protect \citeauthoryear {%
Kirchgessner%
\ \protect \BOthers {.}}{%
Kirchgessner%
\ \protect \BOthers {.}}{%
{\protect \APACyear {2016}}%
}]{%
Kirchgessner2016}
\APACinsertmetastar {%
Kirchgessner2016}%
\begin{APACrefauthors}%
Kirchgessner, N.%
, Liebisch, F.%
, Yu, K.%
, Pfeifer, J.%
, Friedli, M.%
, Hund, A.%
\BCBL {}\ \BBA {} Walter, A.%
\end{APACrefauthors}%
\unskip\
\newblock
\APACrefYearMonthDay{2016}{}{}.
\newblock
{\BBOQ}\APACrefatitle {The {ETH} field phenotyping platform {FIP}: {A}
  cable-suspended multi-sensor system.} {The {ETH} field phenotyping platform
  {FIP}: {A} cable-suspended multi-sensor system.}{\BBCQ}
\newblock
\APACjournalVolNumPages{Functional Plant Biology}{44}{}{154-168}.
\newblock
\begin{APACrefDOI} \doi{https://doi.org/10.1071/FP16165} \end{APACrefDOI}
\PrintBackRefs{\CurrentBib}

\bibitem [\protect \citeauthoryear {%
Koenig%
\ \protect \BOthers {.}}{%
Koenig%
\ \protect \BOthers {.}}{%
{\protect \APACyear {2015}}%
}]{%
Koenig2015}
\APACinsertmetastar {%
Koenig2015}%
\begin{APACrefauthors}%
Koenig, K.%
, Höfle, B.%
, Hämmerle, M.%
, Jarmer, T.%
, Siegmann, B.%
\BCBL {}\ \BBA {} Lilienthal, H.%
\end{APACrefauthors}%
\unskip\
\newblock
\APACrefYearMonthDay{2015}{}{}.
\newblock
{\BBOQ}\APACrefatitle {Comparative classification analysis of post-harvest
  growth detection from terrestrial {LiDAR} point clouds in precision
  agriculture} {Comparative classification analysis of post-harvest growth
  detection from terrestrial {LiDAR} point clouds in precision
  agriculture}.{\BBCQ}
\newblock
\APACjournalVolNumPages{ISPRS Journal of Photogrammetry and Remote
  Sensing}{104}{}{112-125}.
\newblock
\begin{APACrefDOI} \doi{https://doi.org/10.1016/j.isprsjprs.2015.03.003}
  \end{APACrefDOI}
\PrintBackRefs{\CurrentBib}

\bibitem [\protect \citeauthoryear {%
Koranne%
}{%
Koranne%
}{%
{\protect \APACyear {2009}}%
}]{%
Koranne2009}
\APACinsertmetastar {%
Koranne2009}%
\begin{APACrefauthors}%
Koranne, S.%
\end{APACrefauthors}%
\unskip\
\newblock
\APACrefYear{2009}.
\newblock
\APACrefbtitle {Practical Computing on the Cell Broadband Engine} {Practical
  computing on the cell broadband engine}\ (\PrintOrdinal{1st}\ \BEd).
\newblock
\APACaddressPublisher{}{Springer Publishing Company, Incorporated}.
\newblock
\begin{APACrefDOI} \doi{https://doi.org/10.1007/978-1-4419-0308-2}
  \end{APACrefDOI}
\PrintBackRefs{\CurrentBib}

\bibitem [\protect \citeauthoryear {%
Krause%
}{%
Krause%
}{%
{\protect \APACyear {1986}}%
}]{%
Krause1986}
\APACinsertmetastar {%
Krause1986}%
\begin{APACrefauthors}%
Krause, E\BPBI F.%
\end{APACrefauthors}%
\unskip\
\newblock
\APACrefYear{1986}.
\newblock
\APACrefbtitle {Taxicab geometry---an adventure in non-{E}uclidean geometry}
  {Taxicab geometry---an adventure in non-{E}uclidean geometry}.
\newblock
\APACaddressPublisher{}{Courier Corporation}.
\newblock
\begin{APACrefDOI} \doi{https://doi.org/10.2307/3618288} \end{APACrefDOI}
\PrintBackRefs{\CurrentBib}

\bibitem [\protect \citeauthoryear {%
Kumar%
, Pratap%
\BCBL {}\ \BBA {} Kumar%
}{%
Kumar%
\ \protect \BOthers {.}}{%
{\protect \APACyear {2015}}%
}]{%
Kumar2015}
\APACinsertmetastar {%
Kumar2015}%
\begin{APACrefauthors}%
Kumar, J.%
, Pratap, A.%
\BCBL {}\ \BBA {} Kumar, S.%
\end{APACrefauthors}%
\unskip\
\newblock
\APACrefYearMonthDay{2015}{}{}.
\newblock
{\BBOQ}\APACrefatitle {Plant Phenomics: An Overview} {Plant phenomics: An
  overview}.{\BBCQ}
\newblock
\BIn{} J.~Kumar, A.~Pratap\BCBL {}\ \BBA {} S.~Kumar\ (\BEDS), \APACrefbtitle
  {Phenomics in Crop Plants: Trends, Options and Limitations} {Phenomics in
  crop plants: Trends, options and limitations}\ (\BPGS\ 1--10).
\newblock
\APACaddressPublisher{New Delhi}{Springer India}.
\newblock
\begin{APACrefDOI} \doi{https://doi.org/10.1007/978-81-322-2226-2_1}
  \end{APACrefDOI}
\PrintBackRefs{\CurrentBib}

\bibitem [\protect \citeauthoryear {%
Li%
\ \protect \BOthers {.}}{%
Li%
\ \protect \BOthers {.}}{%
{\protect \APACyear {2020}}%
}]{%
Li2020}
\APACinsertmetastar {%
Li2020}%
\begin{APACrefauthors}%
Li, P.%
, Zhang, X.%
, Wang, W.%
, Zheng, H.%
, Yao, X.%
, Tian, Y.%
\BDBL {}Cheng, T.%
\end{APACrefauthors}%
\unskip\
\newblock
\APACrefYearMonthDay{2020}{}{}.
\newblock
{\BBOQ}\APACrefatitle {Estimating aboveground and organ biomass of plant
  canopies across the entire season of rice growth with terrestrial laser
  scanning} {Estimating aboveground and organ biomass of plant canopies across
  the entire season of rice growth with terrestrial laser scanning}.{\BBCQ}
\newblock
\APACjournalVolNumPages{International Journal of Applied Earth Observation and
  Geoinformation}{91}{}{102132}.
\newblock
\begin{APACrefDOI} \doi{https://doi.org/10.1016/j.jag.2020.102132}
  \end{APACrefDOI}
\PrintBackRefs{\CurrentBib}

\bibitem [\protect \citeauthoryear {%
Lin%
, Hu%
, Peng%
, Wang%
\BCBL {}\ \BBA {} Zhai%
}{%
Lin%
\ \protect \BOthers {.}}{%
{\protect \APACyear {2022}}%
}]{%
Lin2022}
\APACinsertmetastar {%
Lin2022}%
\begin{APACrefauthors}%
Lin, C.%
, Hu, F.%
, Peng, J.%
, Wang, J.%
\BCBL {}\ \BBA {} Zhai, R.%
\end{APACrefauthors}%
\unskip\
\newblock
\APACrefYearMonthDay{2022}{}{}.
\newblock
{\BBOQ}\APACrefatitle {Segmentation and Stratification Methods of Field Maize
  Terrestrial {LiDAR} Point Cloud} {Segmentation and stratification methods of
  field maize terrestrial {LiDAR} point cloud}.{\BBCQ}
\newblock
\APACjournalVolNumPages{Agriculture}{}{}{}.
\newblock
\begin{APACrefDOI} \doi{https://doi.org/10.3390/agriculture12091450}
  \end{APACrefDOI}
\PrintBackRefs{\CurrentBib}

\bibitem [\protect \citeauthoryear {%
Liu%
\ \protect \BOthers {.}}{%
Liu%
\ \protect \BOthers {.}}{%
{\protect \APACyear {2017}}%
}]{%
Liu2017}
\APACinsertmetastar {%
Liu2017}%
\begin{APACrefauthors}%
Liu, S.%
, Baret, F.%
, Abichou, M.%
, Boudon, F.%
, Thomas, S.%
, Zhao, K.%
\BDBL {}de Solan, B.%
\end{APACrefauthors}%
\unskip\
\newblock
\APACrefYearMonthDay{2017}{}{}.
\newblock
{\BBOQ}\APACrefatitle {Estimating wheat green area index from ground-based
  {LiDAR} measurement using a {3D} canopy structure model} {Estimating wheat
  green area index from ground-based {LiDAR} measurement using a {3D} canopy
  structure model}.{\BBCQ}
\newblock
\APACjournalVolNumPages{Agricultural and Forest Meteorology}{247}{}{12-20}.
\newblock
\begin{APACrefDOI} \doi{https://doi.org/10.1016/j.agrformet.2017.07.007}
  \end{APACrefDOI}
\PrintBackRefs{\CurrentBib}

\bibitem [\protect \citeauthoryear {%
Llorens%
, Gil%
, Llop%
\BCBL {}\ \BBA {} Escolà%
}{%
Llorens%
\ \protect \BOthers {.}}{%
{\protect \APACyear {2011}}%
}]{%
Llorens2011}
\APACinsertmetastar {%
Llorens2011}%
\begin{APACrefauthors}%
Llorens, J.%
, Gil, E.%
, Llop, J.%
\BCBL {}\ \BBA {} Escolà, A.%
\end{APACrefauthors}%
\unskip\
\newblock
\APACrefYearMonthDay{2011}{}{}.
\newblock
{\BBOQ}\APACrefatitle {Ultrasonic and {LiDAR} Sensors for Electronic Canopy
  Characterization in Vineyards: Advances to Improve Pesticide Application
  Methods} {Ultrasonic and {LiDAR} sensors for electronic canopy
  characterization in vineyards: Advances to improve pesticide application
  methods}.{\BBCQ}
\newblock
\APACjournalVolNumPages{Sensors}{11}{2}{2177--2194}.
\newblock
\begin{APACrefDOI} \doi{https://doi.org/10.3390/s110202177} \end{APACrefDOI}
\PrintBackRefs{\CurrentBib}

\bibitem [\protect \citeauthoryear {%
Madec%
\ \protect \BOthers {.}}{%
Madec%
\ \protect \BOthers {.}}{%
{\protect \APACyear {2017}}%
}]{%
Madec2017}
\APACinsertmetastar {%
Madec2017}%
\begin{APACrefauthors}%
Madec, S.%
, Baret, F.%
, de Solan, B.%
, Thomas, S.%
, Dutartre, D.%
, J{\'e}z{\'e}quel, S.%
\BDBL {}Comar, A.%
\end{APACrefauthors}%
\unskip\
\newblock
\APACrefYearMonthDay{2017}{}{}.
\newblock
{\BBOQ}\APACrefatitle {High-Throughput Phenotyping of Plant Height: Comparing
  Unmanned Aerial Vehicles and Ground {LiDAR} Estimates} {High-throughput
  phenotyping of plant height: Comparing unmanned aerial vehicles and ground
  {LiDAR} estimates}.{\BBCQ}
\newblock
\APACjournalVolNumPages{Frontiers in Plant Science}{8}{}{}.
\newblock
\begin{APACrefDOI} \doi{https://doi.org/10.3389/fpls.2017.02002}
  \end{APACrefDOI}
\PrintBackRefs{\CurrentBib}

\bibitem [\protect \citeauthoryear {%
Malambo%
, Popescu%
, Horne%
, Pugh%
\BCBL {}\ \BBA {} Rooney%
}{%
Malambo%
\ \protect \BOthers {.}}{%
{\protect \APACyear {2019}}%
}]{%
Malambo2019}
\APACinsertmetastar {%
Malambo2019}%
\begin{APACrefauthors}%
Malambo, L.%
, Popescu, S.%
, Horne, D.%
, Pugh, N.%
\BCBL {}\ \BBA {} Rooney, W.%
\end{APACrefauthors}%
\unskip\
\newblock
\APACrefYearMonthDay{2019}{}{}.
\newblock
{\BBOQ}\APACrefatitle {Automated detection and measurement of individual
  sorghum panicles using density-based clustering of terrestrial lidar data}
  {Automated detection and measurement of individual sorghum panicles using
  density-based clustering of terrestrial lidar data}.{\BBCQ}
\newblock
\APACjournalVolNumPages{ISPRS Journal of Photogrammetry and Remote
  Sensing}{149}{}{1-13}.
\newblock
\begin{APACrefDOI} \doi{https://doi.org/10.1016/j.isprsjprs.2018.12.015}
  \end{APACrefDOI}
\PrintBackRefs{\CurrentBib}

\bibitem [\protect \citeauthoryear {%
Miller%
, Tucker%
\BCBL {}\ \BBA {} Zemlin%
}{%
Miller%
\ \protect \BOthers {.}}{%
{\protect \APACyear {1960}}%
}]{%
Miller1960}
\APACinsertmetastar {%
Miller1960}%
\begin{APACrefauthors}%
Miller, C\BPBI E.%
, Tucker, A\BPBI W.%
\BCBL {}\ \BBA {} Zemlin, R\BPBI A.%
\end{APACrefauthors}%
\unskip\
\newblock
\APACrefYearMonthDay{1960}{oct}{}.
\newblock
{\BBOQ}\APACrefatitle {Integer Programming Formulation of Traveling Salesman
  Problems} {Integer programming formulation of traveling salesman
  problems}.{\BBCQ}
\newblock
\APACjournalVolNumPages{J. ACM}{7}{4}{326–329}.
\newblock
\begin{APACrefDOI} \doi{https://doi.org/10.1145/321043.321046} \end{APACrefDOI}
\PrintBackRefs{\CurrentBib}

\bibitem [\protect \citeauthoryear {%
Moore%
\ \BBA {} Stouch%
}{%
Moore%
\ \BBA {} Stouch%
}{%
{\protect \APACyear {2014}}%
}]{%
Moore2014}
\APACinsertmetastar {%
Moore2014}%
\begin{APACrefauthors}%
Moore, T.%
\BCBT {}\ \BBA {} Stouch, D.%
\end{APACrefauthors}%
\unskip\
\newblock
\APACrefYearMonthDay{2014}{July}{}.
\newblock
{\BBOQ}\APACrefatitle {A Generalized Extended Kalman Filter Implementation for
  the Robot Operating System} {A generalized extended kalman filter
  implementation for the robot operating system}.{\BBCQ}
\newblock
\BIn{} \APACrefbtitle {Proceedings of the 13th International Conference on
  Intelligent Autonomous Systems (IAS-13).} {Proceedings of the 13th
  international conference on intelligent autonomous systems (ias-13).}
\newblock
\APACaddressPublisher{}{Springer}.
\newblock
\begin{APACrefDOI} \doi{https://doi.org/10.1007/978-3-319-08338-4_25}
  \end{APACrefDOI}
\PrintBackRefs{\CurrentBib}

\bibitem [\protect \citeauthoryear {%
Mozaffar%
\ \BBA {} Varshosaz%
}{%
Mozaffar%
\ \BBA {} Varshosaz%
}{%
{\protect \APACyear {2016}}%
}]{%
Mozaffar2016}
\APACinsertmetastar {%
Mozaffar2016}%
\begin{APACrefauthors}%
Mozaffar, M\BPBI H.%
\BCBT {}\ \BBA {} Varshosaz, M.%
\end{APACrefauthors}%
\unskip\
\newblock
\APACrefYearMonthDay{2016}{}{}.
\newblock
{\BBOQ}\APACrefatitle {Optimal Placement of a Terrestrial Laser Scanner with an
  Emphasis on Reducing Occlusions} {Optimal placement of a terrestrial laser
  scanner with an emphasis on reducing occlusions}.{\BBCQ}
\newblock
\APACjournalVolNumPages{The Photogrammetric Record}{31}{}{}.
\newblock
\begin{APACrefDOI} \doi{https://doi.org/10.1111/phor.12162} \end{APACrefDOI}
\PrintBackRefs{\CurrentBib}

\bibitem [\protect \citeauthoryear {%
Nobel%
, Forseth%
\BCBL {}\ \BBA {} Long%
}{%
Nobel%
\ \protect \BOthers {.}}{%
{\protect \APACyear {1993}}%
}]{%
Nobel1993}
\APACinsertmetastar {%
Nobel1993}%
\begin{APACrefauthors}%
Nobel, P\BPBI S.%
, Forseth, I\BPBI N.%
\BCBL {}\ \BBA {} Long, S\BPBI P.%
\end{APACrefauthors}%
\unskip\
\newblock
\APACrefYearMonthDay{1993}{}{}.
\newblock
{\BBOQ}\APACrefatitle {Canopy structure and light interception} {Canopy
  structure and light interception}.{\BBCQ}
\newblock
\BIn{} D\BPBI O.~Hall, J\BPBI M\BPBI O.~Scurlock, H\BPBI
  R.~Bolh{\`a}r-Nordenkampf, R\BPBI C.~Leegood\BCBL {}\ \BBA {} S\BPBI P.~Long\
  (\BEDS), \APACrefbtitle {Photosynthesis and Production in a Changing
  Environment: A field and laboratory manual} {Photosynthesis and production in
  a changing environment: A field and laboratory manual}\ (\BPGS\ 79--90).
\newblock
\APACaddressPublisher{Dordrecht}{Springer Netherlands}.
\newblock
\begin{APACrefDOI} \doi{10.1007/978-94-011-1566-7_6} \end{APACrefDOI}
\PrintBackRefs{\CurrentBib}

\bibitem [\protect \citeauthoryear {%
O'Banion%
\ \BBA {} Olsen%
}{%
O'Banion%
\ \BBA {} Olsen%
}{%
{\protect \APACyear {2019}}%
}]{%
OBanion2019}
\APACinsertmetastar {%
OBanion2019}%
\begin{APACrefauthors}%
O'Banion, M\BPBI S.%
\BCBT {}\ \BBA {} Olsen, M\BPBI J.%
\end{APACrefauthors}%
\unskip\
\newblock
\APACrefYearMonthDay{2019}{}{}.
\newblock
{\BBOQ}\APACrefatitle {Efficient Planning and Acquisition of Terrestrial Laser
  Scanning–Derived Digital Elevation Models: Proof of Concept Study}
  {Efficient planning and acquisition of terrestrial laser scanning–derived
  digital elevation models: Proof of concept study}.{\BBCQ}
\newblock
\APACjournalVolNumPages{Journal of Surveying Engineering}{}{}{}.
\newblock
\begin{APACrefDOI} \doi{https://doi.org/10.1061/(ASCE)SU.1943-5428.0000265}
  \end{APACrefDOI}
\PrintBackRefs{\CurrentBib}

\bibitem [\protect \citeauthoryear {%
Paulus%
}{%
Paulus%
}{%
{\protect \APACyear {2019}}%
}]{%
paulus2019}
\APACinsertmetastar {%
paulus2019}%
\begin{APACrefauthors}%
Paulus, S.%
\end{APACrefauthors}%
\unskip\
\newblock
\APACrefYearMonthDay{2019}{}{}.
\newblock
{\BBOQ}\APACrefatitle {Measuring crops in {3D}: using geometry for plant
  phenotyping} {Measuring crops in {3D}: using geometry for plant
  phenotyping}.{\BBCQ}
\newblock
\APACjournalVolNumPages{Plant methods}{15}{1}{1--13}.
\newblock
\begin{APACrefDOI} \doi{https://doi.org/10.1186/s13007-019-0490-0}
  \end{APACrefDOI}
\PrintBackRefs{\CurrentBib}

\bibitem [\protect \citeauthoryear {%
Paulus%
, Schumann%
, Kuhlmann%
\BCBL {}\ \BBA {} Léon%
}{%
Paulus%
\ \protect \BOthers {.}}{%
{\protect \APACyear {2014}}%
}]{%
Paulus2014}
\APACinsertmetastar {%
Paulus2014}%
\begin{APACrefauthors}%
Paulus, S.%
, Schumann, H.%
, Kuhlmann, H.%
\BCBL {}\ \BBA {} Léon, J.%
\end{APACrefauthors}%
\unskip\
\newblock
\APACrefYearMonthDay{2014}{}{}.
\newblock
{\BBOQ}\APACrefatitle {High-precision laser scanning system for capturing {3D}
  plant architecture and analysing growth of cereal plants} {High-precision
  laser scanning system for capturing {3D} plant architecture and analysing
  growth of cereal plants}.{\BBCQ}
\newblock
\APACjournalVolNumPages{Biosystems Engineering}{121}{}{1-11}.
\newblock
\begin{APACrefDOI} \doi{https://doi.org/10.1016/j.biosystemseng.2014.01.010}
  \end{APACrefDOI}
\PrintBackRefs{\CurrentBib}

\bibitem [\protect \citeauthoryear {%
Qiu%
\ \protect \BOthers {.}}{%
Qiu%
\ \protect \BOthers {.}}{%
{\protect \APACyear {2019}}%
}]{%
Qiu2019}
\APACinsertmetastar {%
Qiu2019}%
\begin{APACrefauthors}%
Qiu, Q.%
, Sun, N.%
, Bai, H.%
, Wang, N.%
, Fan, Z.%
, Wang, Y.%
\BDBL {}Cong, Y.%
\end{APACrefauthors}%
\unskip\
\newblock
\APACrefYearMonthDay{2019}{}{}.
\newblock
{\BBOQ}\APACrefatitle {Field-based high-throughput phenotyping for maize plant
  using {3D} {LiDAR} point cloud generated with a “{P}henomobile”}
  {Field-based high-throughput phenotyping for maize plant using {3D} {LiDAR}
  point cloud generated with a “{P}henomobile”}.{\BBCQ}
\newblock
\APACjournalVolNumPages{Frontiers in Plant Science}{10}{}{554}.
\newblock
\begin{APACrefDOI} \doi{https://doi.org/10.3389/fpls.2019.00554}
  \end{APACrefDOI}
\PrintBackRefs{\CurrentBib}

\bibitem [\protect \citeauthoryear {%
Rodriguez-Sanchez%
\ \BBA {} Li%
}{%
Rodriguez-Sanchez%
\ \BBA {} Li%
}{%
{\protect \APACyear {2022}}%
}]{%
Rodriguez2022}
\APACinsertmetastar {%
Rodriguez2022}%
\begin{APACrefauthors}%
Rodriguez-Sanchez, J.%
\BCBT {}\ \BBA {} Li, C.%
\end{APACrefauthors}%
\unskip\
\newblock
\APACrefYearMonthDay{2022}{}{}.
\newblock
{\BBOQ}\APACrefatitle {An Autonomous Ground System for {3D} {LiDAR}-based Crop
  Scouting} {An autonomous ground system for {3D} {LiDAR}-based crop
  scouting}.{\BBCQ}
\newblock
\BIn{} \APACrefbtitle {2022 {ASABE} Annual International Meeting} {2022 {ASABE}
  annual international meeting}\ (\BPG~1).
\newblock
\begin{APACrefDOI} \doi{https://doi.org/10.13031/aim.202200142}
  \end{APACrefDOI}
\PrintBackRefs{\CurrentBib}

\bibitem [\protect \citeauthoryear {%
{Rosell Polo}%
\ \protect \BOthers {.}}{%
{Rosell Polo}%
\ \protect \BOthers {.}}{%
{\protect \APACyear {2009}}%
}]{%
Rosell2009}
\APACinsertmetastar {%
Rosell2009}%
\begin{APACrefauthors}%
{Rosell Polo}, J\BPBI R.%
, Sanz, R.%
, Llorens, J.%
, Arnó, J.%
, Escolà, A.%
, Ribes-Dasi, M.%
\BDBL {}Palacín, J.%
\end{APACrefauthors}%
\unskip\
\newblock
\APACrefYearMonthDay{2009}{}{}.
\newblock
{\BBOQ}\APACrefatitle {A tractor-mounted scanning {LIDAR} for the
  non-destructive measurement of vegetative volume and surface area of tree-row
  plantations: {A} comparison with conventional destructive measurements} {A
  tractor-mounted scanning {LIDAR} for the non-destructive measurement of
  vegetative volume and surface area of tree-row plantations: {A} comparison
  with conventional destructive measurements}.{\BBCQ}
\newblock
\APACjournalVolNumPages{Biosystems Engineering}{102}{2}{128-134}.
\newblock
\begin{APACrefDOI} \doi{https://doi.org/10.1016/j.biosystemseng.2008.10.009}
  \end{APACrefDOI}
\PrintBackRefs{\CurrentBib}

\bibitem [\protect \citeauthoryear {%
Rote%
}{%
Rote%
}{%
{\protect \APACyear {1991}}%
}]{%
Rote1991}
\APACinsertmetastar {%
Rote1991}%
\begin{APACrefauthors}%
Rote, G.%
\end{APACrefauthors}%
\unskip\
\newblock
\APACrefYearMonthDay{1991}{}{}.
\newblock
{\BBOQ}\APACrefatitle {Computing the minimum {H}ausdorff distance between two
  point sets on a line under translation} {Computing the minimum {H}ausdorff
  distance between two point sets on a line under translation}.{\BBCQ}
\newblock
\APACjournalVolNumPages{Information Processing Letters}{38}{3}{123--127}.
\newblock
\begin{APACrefDOI} \doi{https://doi.org/10.1016/0020-0190(91)90233-8}
  \end{APACrefDOI}
\PrintBackRefs{\CurrentBib}

\bibitem [\protect \citeauthoryear {%
Saeys%
, Lenaerts%
, Craessaerts%
\BCBL {}\ \BBA {} {De Baerdemaeker}%
}{%
Saeys%
\ \protect \BOthers {.}}{%
{\protect \APACyear {2009}}%
}]{%
Saeys2009}
\APACinsertmetastar {%
Saeys2009}%
\begin{APACrefauthors}%
Saeys, W.%
, Lenaerts, B.%
, Craessaerts, G.%
\BCBL {}\ \BBA {} {De Baerdemaeker}, J.%
\end{APACrefauthors}%
\unskip\
\newblock
\APACrefYearMonthDay{2009}{}{}.
\newblock
{\BBOQ}\APACrefatitle {Estimation of the crop density of small grains using
  {LiDAR} sensors} {Estimation of the crop density of small grains using
  {LiDAR} sensors}.{\BBCQ}
\newblock
\APACjournalVolNumPages{Biosystems Engineering}{102}{1}{22-30}.
\newblock
\begin{APACrefDOI} \doi{https://doi.org/10.1016/j.biosystemseng.2008.10.003}
  \end{APACrefDOI}
\PrintBackRefs{\CurrentBib}

\bibitem [\protect \citeauthoryear {%
Schneider%
\ \BBA {} Eberly%
}{%
Schneider%
\ \BBA {} Eberly%
}{%
{\protect \APACyear {2002}}%
}]{%
Schneider2002}
\APACinsertmetastar {%
Schneider2002}%
\begin{APACrefauthors}%
Schneider, P\BPBI J.%
\BCBT {}\ \BBA {} Eberly, D.%
\end{APACrefauthors}%
\unskip\
\newblock
\APACrefYear{2002}.
\newblock
\APACrefbtitle {Geometric Tools for Computer Graphics} {Geometric tools for
  computer graphics}.
\newblock
\APACaddressPublisher{USA}{Elsevier Science Inc.}
\newblock
\begin{APACrefDOI} \doi{https://doi.org/10.1016/b978-1-55860-594-7.x5000-0}
  \end{APACrefDOI}
\PrintBackRefs{\CurrentBib}

\bibitem [\protect \citeauthoryear {%
Slavík%
}{%
Slavík%
}{%
{\protect \APACyear {1997}}%
}]{%
Slavik1997}
\APACinsertmetastar {%
Slavik1997}%
\begin{APACrefauthors}%
Slavík, P.%
\end{APACrefauthors}%
\unskip\
\newblock
\APACrefYearMonthDay{1997}{}{}.
\newblock
{\BBOQ}\APACrefatitle {A Tight Analysis of the Greedy Algorithm for Set Cover}
  {A tight analysis of the greedy algorithm for set cover}.{\BBCQ}
\newblock
\APACjournalVolNumPages{Journal of Algorithms}{25}{2}{237-254}.
\newblock
\begin{APACrefDOI} \doi{https://doi.org/10.1006/jagm.1997.0887}
  \end{APACrefDOI}
\PrintBackRefs{\CurrentBib}

\bibitem [\protect \citeauthoryear {%
Smits%
}{%
Smits%
}{%
{\protect \APACyear {1998}}%
}]{%
Smits1998}
\APACinsertmetastar {%
Smits1998}%
\begin{APACrefauthors}%
Smits, B.%
\end{APACrefauthors}%
\unskip\
\newblock
\APACrefYearMonthDay{1998}{feb}{}.
\newblock
{\BBOQ}\APACrefatitle {Efficiency Issues for Ray Tracing} {Efficiency issues
  for ray tracing}.{\BBCQ}
\newblock
\APACjournalVolNumPages{Journal of Graphical Tools}{3}{2}{1–14}.
\newblock
\begin{APACrefDOI} \doi{https://doi.org/10.1080/10867651.1998.10487488}
  \end{APACrefDOI}
\PrintBackRefs{\CurrentBib}

\bibitem [\protect \citeauthoryear {%
Starek%
, Chu%
, Mitasova%
\BCBL {}\ \BBA {} Harmon%
}{%
Starek%
\ \protect \BOthers {.}}{%
{\protect \APACyear {2020}}%
}]{%
Starek2020}
\APACinsertmetastar {%
Starek2020}%
\begin{APACrefauthors}%
Starek, M\BPBI J.%
, Chu, T.%
, Mitasova, H.%
\BCBL {}\ \BBA {} Harmon, R\BPBI S.%
\end{APACrefauthors}%
\unskip\
\newblock
\APACrefYearMonthDay{2020}{}{}.
\newblock
{\BBOQ}\APACrefatitle {Viewshed simulation and optimization for digital terrain
  modelling with terrestrial laser scanning} {Viewshed simulation and
  optimization for digital terrain modelling with terrestrial laser
  scanning}.{\BBCQ}
\newblock
\APACjournalVolNumPages{International Journal of Remote
  Sensing}{41}{16}{6409-6426}.
\newblock
\begin{APACrefDOI} \doi{https://doi.org/10.1080/01431161.2020.1752952}
  \end{APACrefDOI}
\PrintBackRefs{\CurrentBib}

\bibitem [\protect \citeauthoryear {%
Su%
\ \protect \BOthers {.}}{%
Su%
\ \protect \BOthers {.}}{%
{\protect \APACyear {2019}}%
}]{%
Su2019}
\APACinsertmetastar {%
Su2019}%
\begin{APACrefauthors}%
Su, Y.%
, Wu, F.%
, Ao, Z.%
, Jin, S.%
, Qin, F.%
, Liu, B.%
\BDBL {}Guo, Q.%
\end{APACrefauthors}%
\unskip\
\newblock
\APACrefYearMonthDay{2019}{}{}.
\newblock
{\BBOQ}\APACrefatitle {Evaluating maize phenotype dynamics under drought stress
  using terrestrial lidar} {Evaluating maize phenotype dynamics under drought
  stress using terrestrial lidar}.{\BBCQ}
\newblock
\APACjournalVolNumPages{Plant Methods}{15}{}{}.
\newblock
\begin{APACrefDOI} \doi{https://doi.org/10.1186/s13007-019-0396-x}
  \end{APACrefDOI}
\PrintBackRefs{\CurrentBib}

\bibitem [\protect \citeauthoryear {%
Sun%
\ \protect \BOthers {.}}{%
Sun%
\ \protect \BOthers {.}}{%
{\protect \APACyear {2021}}%
}]{%
Sun2021}
\APACinsertmetastar {%
Sun2021}%
\begin{APACrefauthors}%
Sun, S.%
, Li, C.%
, Chee, P\BPBI W.%
, Paterson, A\BPBI H.%
, Meng, C.%
, Zhang, J.%
\BDBL {}Adhikari, J.%
\end{APACrefauthors}%
\unskip\
\newblock
\APACrefYearMonthDay{2021}{}{}.
\newblock
{\BBOQ}\APACrefatitle {High resolution {3D} terrestrial {LiDAR} for cotton
  plant main stalk and node detection} {High resolution {3D} terrestrial
  {LiDAR} for cotton plant main stalk and node detection}.{\BBCQ}
\newblock
\APACjournalVolNumPages{Computers and Electronics in
  Agriculture}{187}{}{106276}.
\newblock
\begin{APACrefDOI} \doi{https://doi.org/10.1016/j.compag.2021.106276}
  \end{APACrefDOI}
\PrintBackRefs{\CurrentBib}

\bibitem [\protect \citeauthoryear {%
Sun%
\ \protect \BOthers {.}}{%
Sun%
\ \protect \BOthers {.}}{%
{\protect \APACyear {2018}}%
}]{%
Sun2018}
\APACinsertmetastar {%
Sun2018}%
\begin{APACrefauthors}%
Sun, S.%
, Li, C.%
, Paterson, A\BPBI H.%
, Jiang, Y.%
, Xu, R.%
, Robertson, J\BPBI S.%
\BDBL {}Chee, P\BPBI W.%
\end{APACrefauthors}%
\unskip\
\newblock
\APACrefYearMonthDay{2018}{}{}.
\newblock
{\BBOQ}\APACrefatitle {In-field High Throughput Phenotyping and Cotton Plant
  Growth Analysis Using {LiDAR}} {In-field high throughput phenotyping and
  cotton plant growth analysis using {LiDAR}}.{\BBCQ}
\newblock
\APACjournalVolNumPages{Frontiers in Plant Science}{9}{}{}.
\newblock
\begin{APACrefDOI} \doi{https://doi.org/10.3389/fpls.2018.00016}
  \end{APACrefDOI}
\PrintBackRefs{\CurrentBib}

\bibitem [\protect \citeauthoryear {%
{The MathWorks Inc.}%
}{%
{The MathWorks Inc.}%
}{%
{\protect \APACyear {2021}}%
}]{%
MatlabOTB}
\APACinsertmetastar {%
MatlabOTB}%
\begin{APACrefauthors}%
{The MathWorks Inc.}%
\end{APACrefauthors}%
\unskip\
\newblock
\APACrefYearMonthDay{2021}{}{}.
\newblock
\APACrefbtitle {MATLAB Optimization Toolbox.} {Matlab optimization toolbox.}
\newblock
\APACaddressPublisher{Natick, Massachusetts, United States}{The MathWorks Inc.}
\newblock
\begin{APACrefURL} \url{https://www.mathworks.com/help/stats/index.html}
  \end{APACrefURL}
\PrintBackRefs{\CurrentBib}

\bibitem [\protect \citeauthoryear {%
Thulasiraman%
, Arumugam%
, Brandst{\"a}dt%
\BCBL {}\ \BBA {} Nishizeki%
}{%
Thulasiraman%
\ \protect \BOthers {.}}{%
{\protect \APACyear {2016}}%
}]{%
Thulasiraman2010}
\APACinsertmetastar {%
Thulasiraman2010}%
\begin{APACrefauthors}%
Thulasiraman, K.%
, Arumugam, S.%
, Brandst{\"a}dt, A.%
\BCBL {}\ \BBA {} Nishizeki, T.%
\end{APACrefauthors}%
\ (\BEDS).
\unskip\
\newblock
\APACrefYear{2016}.
\newblock
\APACrefbtitle {Handbook of graph theory, combinatorial optimization, and
  algorithms} {Handbook of graph theory, combinatorial optimization, and
  algorithms}.
\newblock
\APACaddressPublisher{}{CRC Press}.
\newblock
\begin{APACrefDOI} \doi{https://doi.org/10.1201/b19163} \end{APACrefDOI}
\PrintBackRefs{\CurrentBib}

\bibitem [\protect \citeauthoryear {%
Tilly%
\ \protect \BOthers {.}}{%
Tilly%
\ \protect \BOthers {.}}{%
{\protect \APACyear {2015}}%
}]{%
Tilly2015}
\APACinsertmetastar {%
Tilly2015}%
\begin{APACrefauthors}%
Tilly, N.%
, Hoffmeister, D.%
, Cao, Q.%
, Lenz-Wiedemann, V.%
, Miao, Y.%
\BCBL {}\ \BBA {} Bareth, G.%
\end{APACrefauthors}%
\unskip\
\newblock
\APACrefYearMonthDay{2015}{}{}.
\newblock
{\BBOQ}\APACrefatitle {Transferability of Models for Estimating Paddy Rice
  Biomass from Spatial Plant Height Data} {Transferability of models for
  estimating paddy rice biomass from spatial plant height data}.{\BBCQ}
\newblock
\APACjournalVolNumPages{Agriculture}{5}{3}{538--560}.
\newblock
\begin{APACrefDOI} \doi{https://doi.org/10.3390/agriculture5030538}
  \end{APACrefDOI}
\PrintBackRefs{\CurrentBib}

\bibitem [\protect \citeauthoryear {%
Virlet%
, Sabermanesh%
, Sadeghi-Tehran%
\BCBL {}\ \BBA {} Hawkesford%
}{%
Virlet%
\ \protect \BOthers {.}}{%
{\protect \APACyear {2016}}%
}]{%
Virlet2016}
\APACinsertmetastar {%
Virlet2016}%
\begin{APACrefauthors}%
Virlet, N.%
, Sabermanesh, K.%
, Sadeghi-Tehran, P.%
\BCBL {}\ \BBA {} Hawkesford, M\BPBI J.%
\end{APACrefauthors}%
\unskip\
\newblock
\APACrefYearMonthDay{2016}{}{}.
\newblock
{\BBOQ}\APACrefatitle {Field Scanalyzer: An automated robotic field phenotyping
  platform for detailed crop monitoring} {Field scanalyzer: An automated
  robotic field phenotyping platform for detailed crop monitoring}.{\BBCQ}
\newblock
\APACjournalVolNumPages{Functional Plant Biology}{44}{}{143-153}.
\newblock
\begin{APACrefDOI} \doi{https://doi.org/10.1071/FP16163} \end{APACrefDOI}
\PrintBackRefs{\CurrentBib}

\bibitem [\protect \citeauthoryear {%
Vosselman%
\ \BBA {} Maas%
}{%
Vosselman%
\ \BBA {} Maas%
}{%
{\protect \APACyear {2010}}%
}]{%
Vosselman2010}
\APACinsertmetastar {%
Vosselman2010}%
\begin{APACrefauthors}%
Vosselman, G.%
\BCBT {}\ \BBA {} Maas, H.%
\end{APACrefauthors}%
\ (\BEDS).
\unskip\
\newblock
\APACrefYear{2010}.
\newblock
\APACrefbtitle {Airborne and terrestrial laser scanning} {Airborne and
  terrestrial laser scanning}.
\newblock
\APACaddressPublisher{}{CRC Press (Taylor \& Francis)}.
\PrintBackRefs{\CurrentBib}

\bibitem [\protect \citeauthoryear {%
Walter%
, Edwards%
, McDonald%
\BCBL {}\ \BBA {} Kuchel%
}{%
Walter%
\ \protect \BOthers {.}}{%
{\protect \APACyear {2019}}%
}]{%
Walter2019}
\APACinsertmetastar {%
Walter2019}%
\begin{APACrefauthors}%
Walter, J.%
, Edwards, J.%
, McDonald, G\BPBI K.%
\BCBL {}\ \BBA {} Kuchel, H.%
\end{APACrefauthors}%
\unskip\
\newblock
\APACrefYearMonthDay{2019}{}{}.
\newblock
{\BBOQ}\APACrefatitle {Estimating Biomass and Canopy Height With {LiDAR} for
  Field Crop Breeding} {Estimating biomass and canopy height with {LiDAR} for
  field crop breeding}.{\BBCQ}
\newblock
\APACjournalVolNumPages{Frontiers in Plant Science}{10}{}{}.
\newblock
\begin{APACrefDOI} \doi{https://doi.org/10.3389/fpls.2019.01145}
  \end{APACrefDOI}
\PrintBackRefs{\CurrentBib}

\bibitem [\protect \citeauthoryear {%
Wang%
, Singh%
, Marla%
, Morris%
\BCBL {}\ \BBA {} Poland%
}{%
Wang%
\ \protect \BOthers {.}}{%
{\protect \APACyear {2018}}%
}]{%
Wang2018}
\APACinsertmetastar {%
Wang2018}%
\begin{APACrefauthors}%
Wang, X.%
, Singh, D\BPBI S\BPBI K.%
, Marla, S\BPBI R.%
, Morris, G\BPBI P.%
\BCBL {}\ \BBA {} Poland, J\BPBI A.%
\end{APACrefauthors}%
\unskip\
\newblock
\APACrefYearMonthDay{2018}{}{}.
\newblock
{\BBOQ}\APACrefatitle {Field-based high-throughput phenotyping of plant height
  in sorghum using different sensing technologies} {Field-based high-throughput
  phenotyping of plant height in sorghum using different sensing
  technologies}.{\BBCQ}
\newblock
\APACjournalVolNumPages{Plant Methods}{14}{}{}.
\newblock
\begin{APACrefDOI} \doi{https://doi.org/10.1186/s13007-018-0324-5}
  \end{APACrefDOI}
\PrintBackRefs{\CurrentBib}

\bibitem [\protect \citeauthoryear {%
Watt%
\ \protect \BOthers {.}}{%
Watt%
\ \protect \BOthers {.}}{%
{\protect \APACyear {2020}}%
}]{%
Watt2020}
\APACinsertmetastar {%
Watt2020}%
\begin{APACrefauthors}%
Watt, M.%
, Fiorani, F.%
, Usadel, B.%
, Rascher, U.%
, Muller, O.%
\BCBL {}\ \BBA {} Schurr, U.%
\end{APACrefauthors}%
\unskip\
\newblock
\APACrefYearMonthDay{2020}{}{}.
\newblock
{\BBOQ}\APACrefatitle {Phenotyping: New windows into the plant for breeders}
  {Phenotyping: New windows into the plant for breeders}.{\BBCQ}
\newblock
\APACjournalVolNumPages{Annual review of plant biology}{71}{}{689--712}.
\newblock
\begin{APACrefDOI} \doi{https://doi.org/10.1146/annurev-arplant-042916-041124}
  \end{APACrefDOI}
\PrintBackRefs{\CurrentBib}

\bibitem [\protect \citeauthoryear {%
Williams%
, Barrus%
, Morley%
\BCBL {}\ \BBA {} Shirley%
}{%
Williams%
\ \protect \BOthers {.}}{%
{\protect \APACyear {2005}}%
}]{%
Williams2005}
\APACinsertmetastar {%
Williams2005}%
\begin{APACrefauthors}%
Williams, A.%
, Barrus, S.%
, Morley, R\BPBI K.%
\BCBL {}\ \BBA {} Shirley, P.%
\end{APACrefauthors}%
\unskip\
\newblock
\APACrefYearMonthDay{2005}{}{}.
\newblock
{\BBOQ}\APACrefatitle {An Efficient and Robust Ray-Box Intersection Algorithm}
  {An efficient and robust ray-box intersection algorithm}.{\BBCQ}
\newblock
\BIn{} \APACrefbtitle {ACM SIGGRAPH 2005 Courses} {Acm siggraph 2005 courses}\
  (\BPG~9–es).
\newblock
\APACaddressPublisher{New York, NY, USA}{Association for Computing Machinery}.
\newblock
\begin{APACrefDOI} \doi{https://doi.org/10.1145/1198555.1198748}
  \end{APACrefDOI}
\PrintBackRefs{\CurrentBib}

\bibitem [\protect \citeauthoryear {%
Xu%
\ \BBA {} Li%
}{%
Xu%
\ \BBA {} Li%
}{%
{\protect \APACyear {2022}}%
{\protect \APACexlab {{\protect \BCnt {1}}}}}]{%
Xu2022Amodular}
\APACinsertmetastar {%
Xu2022Amodular}%
\begin{APACrefauthors}%
Xu, R.%
\BCBT {}\ \BBA {} Li, C.%
\end{APACrefauthors}%
\unskip\
\newblock
\APACrefYearMonthDay{2022{\protect \BCnt {1}}}{}{}.
\newblock
{\BBOQ}\APACrefatitle {A modular agricultural robotic system {(MARS)} for
  precision farming: Concept and implementation} {A modular agricultural
  robotic system {(MARS)} for precision farming: Concept and
  implementation}.{\BBCQ}
\newblock
\APACjournalVolNumPages{Journal of Field Robotics}{39}{4}{387-409}.
\newblock
\begin{APACrefDOI} \doi{https://doi.org/10.1002/rob.22056} \end{APACrefDOI}
\PrintBackRefs{\CurrentBib}

\bibitem [\protect \citeauthoryear {%
Xu%
\ \BBA {} Li%
}{%
Xu%
\ \BBA {} Li%
}{%
{\protect \APACyear {2022}}%
{\protect \APACexlab {{\protect \BCnt {2}}}}}]{%
Xu2022Areview}
\APACinsertmetastar {%
Xu2022Areview}%
\begin{APACrefauthors}%
Xu, R.%
\BCBT {}\ \BBA {} Li, C.%
\end{APACrefauthors}%
\unskip\
\newblock
\APACrefYearMonthDay{2022{\protect \BCnt {2}}}{}{}.
\newblock
{\BBOQ}\APACrefatitle {A Review of High-Throughput Field Phenotyping Systems:
  Focusing on Ground Robots} {A review of high-throughput field phenotyping
  systems: Focusing on ground robots}.{\BBCQ}
\newblock
\APACjournalVolNumPages{Plant Phenomics}{}{}{}.
\newblock
\begin{APACrefDOI} \doi{https://doi.org/10.34133/2022/9760269} \end{APACrefDOI}
\PrintBackRefs{\CurrentBib}

\bibitem [\protect \citeauthoryear {%
Yuan%
, Bennett%
, Wang%
\BCBL {}\ \BBA {} Chamberlin%
}{%
Yuan%
\ \protect \BOthers {.}}{%
{\protect \APACyear {2019}}%
}]{%
Yuan2019}
\APACinsertmetastar {%
Yuan2019}%
\begin{APACrefauthors}%
Yuan, H.%
, Bennett, R\BPBI S.%
, Wang, N.%
\BCBL {}\ \BBA {} Chamberlin, K\BPBI D.%
\end{APACrefauthors}%
\unskip\
\newblock
\APACrefYearMonthDay{2019}{}{}.
\newblock
{\BBOQ}\APACrefatitle {Development of a peanut canopy measurement system using
  a ground-based lidar sensor} {Development of a peanut canopy measurement
  system using a ground-based lidar sensor}.{\BBCQ}
\newblock
\APACjournalVolNumPages{Frontiers in plant science}{10}{}{203}.
\newblock
\begin{APACrefDOI} \doi{https://doi.org/10.3389/fpls.2019.00203}
  \end{APACrefDOI}
\PrintBackRefs{\CurrentBib}

\bibitem [\protect \citeauthoryear {%
Zhu%
\ \protect \BOthers {.}}{%
Zhu%
\ \protect \BOthers {.}}{%
{\protect \APACyear {2021}}%
}]{%
Zhu2021}
\APACinsertmetastar {%
Zhu2021}%
\begin{APACrefauthors}%
Zhu, Y.%
, Sun, G.%
, Ding, G.%
, Zhou, J.%
, Wen, M.%
, Jin, S.%
\BDBL {}Zhou, J.%
\end{APACrefauthors}%
\unskip\
\newblock
\APACrefYearMonthDay{2021}{}{}.
\newblock
{\BBOQ}\APACrefatitle {Large-scale field phenotyping using backpack {LiDAR} and
  {GUI}-based {CropQuant-3D} to measure structural responses to different
  nitrogen treatments in wheat} {Large-scale field phenotyping using backpack
  {LiDAR} and {GUI}-based {CropQuant-3D} to measure structural responses to
  different nitrogen treatments in wheat}.{\BBCQ}
\newblock
\APACjournalVolNumPages{bioRxiv}{}{}{}.
\newblock
\begin{APACrefDOI} \doi{https://doi.org/10.1101/2021.05.19.444842}
  \end{APACrefDOI}
\PrintBackRefs{\CurrentBib}

\end{thebibliography}

\end{document}